\documentclass{article}
\pdfoutput=1

\usepackage[square,sort,comma,numbers]{natbib}

\usepackage[preprint]{arxiv}

\usepackage[utf8]{inputenc} 
\usepackage[T1]{fontenc}    
\usepackage{hyperref}       
\usepackage{url}            
\usepackage{booktabs}       
\usepackage{amsmath}        
\usepackage{amsfonts}       
\usepackage{nicefrac}       
\usepackage{microtype}      
\usepackage{lipsum}
\usepackage[size=tiny]{todonotes}
\usepackage{multicol}

\usepackage{color, colortbl}
\definecolor{Gray}{gray}{0.9}
\definecolor{LightCyan}{rgb}{0.88,1,1}
\definecolor{redpink}{rgb}{0.73,0.12,0.41}
\definecolor{bluepurp}{rgb}{0.34,0.26,0.89}
\definecolor{purp}{rgb}{0.86,0.81,1}
\definecolor{strongblue}{rgb}{0.11,0.20,0.71}
\usepackage{rotating}

\hypersetup{
    colorlinks=true,
    linkcolor=bluepurp,
    filecolor=magenta,
    urlcolor=strongblue,
    citecolor=redpink,
    }
\usepackage{graphicx}
\usepackage{caption}
\usepackage{subcaption}

\title{Simulation Intelligence: Towards a New Generation of Scientific Methods}

\author{
    \textbf{Alexander Lavin}\thanks{\url{lavin@simulation.science} (ISI \& Pasteur Labs)} \\
    Institute for Simulation Intelligence
\and
    \textbf{David Krakauer}\\ 
    Santa Fe Institute
\and
    \textbf{Hector Zenil}\\
    The Alan Turing Institute\\
\and
    \textbf{Justin Gottschlich} \\
    Intel Labs
\and
    \textbf{Tim Mattson} \\
    Intel
\and
    \textbf{Kamil Rocki} \\ 
    Neuralink
\and
    \textbf{Anima Anandkumar} \\
    Nvidia
\and
    \textbf{Sanjay Choudry} \\
    Nvidia
\\
\and
    \textbf{Johann Brehmer} \\ 
    Qualcomm AI
\\
\and
    \textbf{Atılım Güneş Baydin} \\
    University of Oxford
\and
    \textbf{Carina Prunkl}\\
    University of Oxford
\and
    \textbf{Brooks Paige}\\
    The Alan Turing Institute
\and
    \textbf{Olexandr Isayev}\\
    Carnegie Mellon University
\\
\and
    \textbf{Erik J. Peterson} \\
    Carnegie Mellon University
\and
    \textbf{Peter L. McMahon}\\
    Cornell University
\and
    \textbf{Jakob H. Macke} \\
    University of Tübingen
\and
    \textbf{Kyle Cranmer} \\
    New York University
\\
\and 
    \textbf{Jiaxin Zhang} \\
    Oak Ridge National Lab
\and
    \textbf{Haruko Wainwright} \\
    Lawrence Berkeley National Lab
\and
    \textbf{Adi Hanuka} \\ 
    SLAC National Accelerator Lab
\\
\and
    \textbf{Samuel Assefa}\\
    US Bank AI Innovation
\and
    \textbf{Stephan Zheng}\\
    Salesforce Research
\and
    \textbf{Manuela Veloso}\\
    JPM AI Research
\\
\and
    \textbf{Avi Pfeffer} \\ 
    Charles River Analytics
}

\begin{document}
\maketitle
\setcounter{footnote}{0}

\begin{abstract}
The original ``Seven Motifs'' set forth a roadmap of essential methods for the field of scientific computing, where a motif is an algorithmic method that captures a pattern of computation and data movement.\footnote{We eschew the original term ``dwarf'' for the more appropriate ``motif'' in this paper, and encourage the field to follow suit.} We present the \textit{Nine Motifs of Simulation Intelligence}, a roadmap for the development and integration of the essential algorithms necessary for a merger of scientific computing, scientific simulation, and artificial intelligence. We call this merger \emph{simulation intelligence} (SI), for short. We posit the motifs of simulation intelligence are interconnected and interdependent, much like the components within the layers of an operating system. Using this metaphor, we explore the nature of each layer of the simulation intelligence ``operating system'' stack (SI-stack) and the motifs therein: 
\begin{multicols}{2}
\begin{enumerate}
    \item Multi-physics and multi-scale modeling
    \item Surrogate modeling and emulation
    \item Simulation-based inference
    \item Causal modeling and inference
    \item Agent-based modeling
    \item Probabilistic programming
    \item Differentiable programming
    \item Open-ended optimization
    \item Machine programming
\end{enumerate}
\end{multicols}
We believe coordinated efforts between motifs offers immense opportunity to accelerate scientific discovery, from solving inverse problems in synthetic biology and climate science, to directing nuclear energy experiments and predicting emergent behavior in socioeconomic settings. We elaborate on each layer of the SI-stack, detailing the state-of-art methods, presenting examples to highlight challenges and opportunities, and advocating for specific ways to advance the motifs and the synergies from their combinations.
Advancing and integrating these technologies can enable a robust and efficient \emph{hypothesis--simulation--analysis} type of scientific method, which we introduce with several 
use-cases for human-machine teaming and automated science. 
\end{abstract}

\begin{keywords}
Simulation; Artificial Intelligence; Machine Learning; Scientific Computing; Physics-infused ML; Inverse Design; Human-Machine Teaming; Optimization; Causality; Complexity; Open-endedness; Machine Programming
\end{keywords}

\renewcommand{\thefootnote}{\roman{footnote}}

\hfill \break
\tableofcontents

\newpage
\section{Introduction}\label{sec_intro}

Simulation has become an indispensable tool for researchers across the sciences to explore the behavior of complex, dynamic systems under varying conditions \cite{Cranmer2020TheFO}, including hypothetical or extreme conditions, and increasingly tipping points in environments such as climate \cite{vautard2013simulation,dunne2013gdfl,roberts2018benefits}, biology \cite{dada2011multi,dror2012biomolecular}, sociopolitics \cite{elshafei2016sensitivity,hamilton2005climate}, and others with significant consequences. Yet there are challenges that limit the utility of simulators (and modeling tools broadly) in many settings. First, despite advances in hardware to enable simulations to model increasingly complex systems, computational costs severely limit the level of geometric details, complexity of physics, and the number of simulator runs. This can lead to simplifying assumptions, which often render the results unusable for hypothesis testing and practical decision-making. In addition, simulators are inherently biased as they simulate only what they are programmed to simulate; sensitivity and uncertainty analyses are often impractical for expensive simulators; simulation code is composed of low-level mechanistic components that are typically non-differentiable and lead to intractable likelihoods; and simulators can rarely integrate with real-world data streams, let alone run online with live data updates.

Recent progress with artificial intelligence (AI) and machine learning (ML) in the sciences has advanced methods towards several key objectives for AI/ML to be useful in sciences (beyond discovering patterns in high-dimensional data). These advances allow us to import priors or domain knowledge into ML models and export knowledge from learned models back to the scientific domain; leverage ML for numerically intractable simulation and optimization problems, as well as maximize the utility of real-world data; generate myriads of synthetic data; quantify and reason about uncertainties in models and data; and infer causal relationships in the data.

It is at the intersection of AI and simulation sciences where we can expect significant strides in scientific experimentation and discovery, in essentially all domains. For instance, the use of neural networks to accelerate simulation software for climate science \cite{Ramadhan2020CapturingMP}, or multi-agent reinforcement learning and game theory towards economic policy simulations \cite{Zheng2020TheAE}. Yet this area is relatively nascent and disparate, and a unifying holistic perspective is needed to advance the intersection of AI and simulation sciences.

This paper explores this perspective. We lay out the methodologies required to make significant strides in simulation and AI for science, and how they must be fruitfully combined. The field of scientific computing was at a similar inflection point when Phillip Colella in 2004 presented to DARPA the \textit{``Seven Dwarfs'' for Scientific Computing}, where each of the seven represents an algorithmic method that captures a pattern of computation and data movement \cite{7dwarfs,Asanovi2006TheLO,kaltofen2012seven}.\footnote{The 2004 ``Seven Motifs for Scientific Computing'': Dense Linear Algebra, Sparse Linear Algebra, Computations on Structured Grids, Computations on Unstructured Grids, Spectral Methods, Particle Methods, and Monte Carlo \cite{7dwarfs}.
}
For the remainder of this paper, we choose to replace a potentially insensitive term with ``motif'', a change we suggest for the field going forward.

The motifs nomenclature has proven useful for reasoning at a high level of abstraction about the behavior and requirements of these methods across a broad range of applications, while decoupling these from specific implementations. Even more, it is an understandable vocabulary for talking across disciplinary boundaries. Motifs also provide ``anti-benchmarks'': not tied to narrow performance or code artifacts, thus encouraging innovation in algorithms, programming languages, data structures, and hardware \cite{Asanovi2006TheLO}. Therefore the motifs of scientific computing provided an explicit roadmap for R\&D efforts in numerical methods (and eventually parallel computing) in sciences.

In this paper, we similarly define the \textit{Nine Motifs of Simulation Intelligence}, classes of complementary algorithmic methods that represent the foundation for synergistic simulation and AI technologies to advance sciences; \textit{simulation intelligence (SI)} describes a field that merges of scientific computing, scientific simulation, and artificial intelligence towards studying processes and systems \textit{in silico} to better understand and discover \textit{in situ} phenomena. 
Each of the SI motifs has momentum from the scientific computing and AI communities, yet must be pursued in concert and integrated in order to pass the existing bottlenecks of scientific simulators, overcome the shortcomings of engineering physics\footnote{\textit{Engineering physics} (or \textit{engineering science}) refers to the study of the combined disciplines of physics, mathematics, chemistry, biology, and engineering, particularly computer, nuclear, electrical, electronic, aerospace, materials or mechanical engineering. The discipline is meant for cross-functionality and bridging the gap between theoretical science and practical engineering, seeking ways to apply, design, and develop new solutions in engineering.}, and enable new scientific workflows.

\begin{figure}[!ht]
\centering
\includegraphics[width=1.0\linewidth]{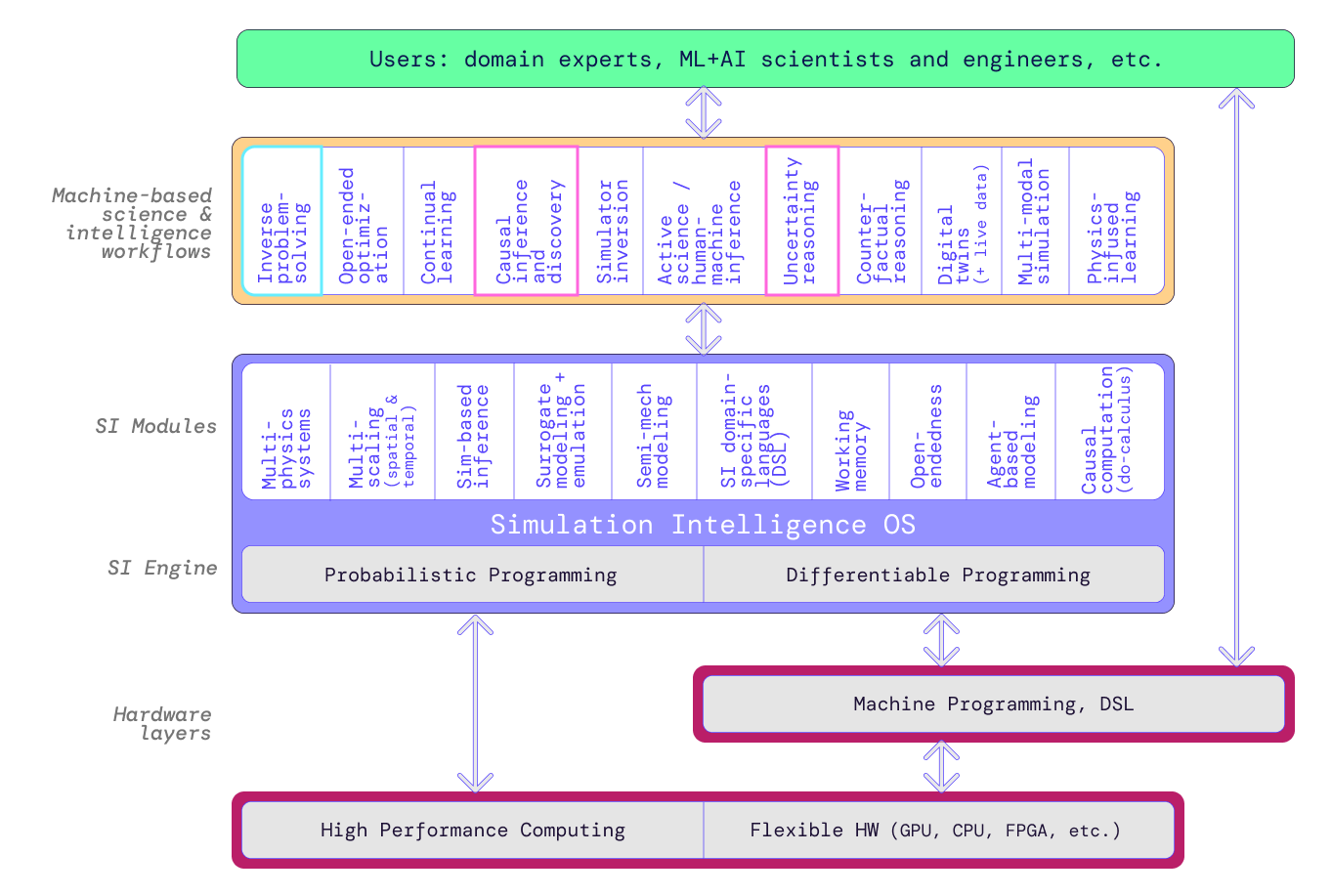}
\caption{An operating system (OS) diagram, elucidating the relationships of the nine Simulation Intelligence motifs, their close relatives, such as domain-specific languages (DSL) \cite{Matthee2011DomainSL} and working memory \cite{Gershman2019TheRA}, and subsequent SI-based science workflows. 
The red, purple, orange, and green plates represent the hardware, OS, application, and user layers, respectively (from bottom to top). In the applications layer (orange plate) are some main SI workflows -- notice some entries have a cyan outline, signifying machine-science classes that contain multiple workflows in themselves (see text for details), and entries with a pink outline denote statistical and ML methods that have been used for decades, but here in unique SI ways. In the text we generically refer to this composition as the ``SI stack'', although the mapping to an OS is a more precise representation of the SI layers and their interactions.
We could have shown a simpler diagram with only the nine SI motifs, but the context of the broader stack that is enabled by their integration better elucidates the improved and new methods that can arise. 
}
\label{fig:os}
\end{figure}

Unlike the older seven motifs of scientific computing, our SI motifs are not necessarily independent. Many of these are interconnected and interdependent, much like the components within the layers of an operating system. 
The individual modules can be combined and interact in multiple ways, gaining from this combination.  Using this metaphor, we explore the nature of each layer of the ``SI stack'', the motifs within each layer, and the combinatorial possibilities available when they are brought together -- the layers are illustrated in Fig.~\ref{fig:os}.

We begin by describing the core layers of the SI stack, detailing each motif within: the concepts, challenges, state-of-the-art methods, future directions, ethical considerations, and many motivating examples. As we traverse the SI stack, encountering the numerous modules and scientific workflows, we will ultimately be able to lay out how these advances will benefit the many users of simulation and scientific endeavors.
Our discussion continues to cover important SI themes such as inverse problem solving and human-machine teaming, and essential infrastructure areas such as data engineering and accelerated computing.

By pursuing research in each of these SI motifs, as well as ways of combining them together in generalizable software towards specific applications in science and intelligence, the recent trend of decelerating progress may be reversed \cite{Cowen2019IsTR,Bhattacharya2020StagnationAS}.
This paper aims to motivate the field and provide a roadmap for those who wish to work in AI and simulation in pursuit of new scientific methods and frontiers.

\hfill \break

\section{Simulation Intelligence Motifs}

Although we define nine concrete motifs, they are not necessarily independent -- many of the motifs are synergistic in function and utility, and some may be building blocks underlying others. We start by describing the ``module'' motifs, as in Fig. \ref{fig:os}, followed by the underlying ``engine'' motifs: probabilistic and differentiable programming. We then describe the motifs that aim to push the frontier of intelligent machines: open-endedness and machine programming.

\hfill \break
\section{\textit{The Modules}}

Above the ``engine'' in the proverbial SI stack (Fig. \ref{fig:os}) are ``modules'', each of which can be built in probabilistic and  differentiable programming frameworks, and make use of the accelerated computing building blocks in the hardware layer.
We start this section with several motifs that closely relate to the physics-informed learning topics most recently discussed above, and then proceed through the SI stack, describing how and why the module motifs compliment one another in myriad, synergistic ways.

\subsection{1. MULTI-PHYSICS \& MULTI-SCALE MODELING}

Simulations are pervasive in every domain of science and engineering, yet are often done in isolation: climate simulations of coastal erosion do not model human-driven effects such as urbanization and mining, and even so are only consistent within a constrained region and timescale. Natural systems involve various types of physical phenomena operating at different spatial and temporal scales. For simulations to be accurate and useful they must support multiple physics and multiple scales (spatial and temporal). The same goes for AI \& ML, which can be powerful for modeling multi-modality, multi-fidelity scientific data, but machine learning alone -- based on data-driven relationships -- ignores the fundamental laws of physics and can result in ill-posed problems or non-physical solutions. For ML (and AI-driven simulation) to be accurate and reliable in the sciences, methods must integrate multi-scale, multi-physics data and uncover mechanisms that explain the emergence of function.


\begin{figure}[!ht]
\centering
\includegraphics[width=0.6\linewidth]{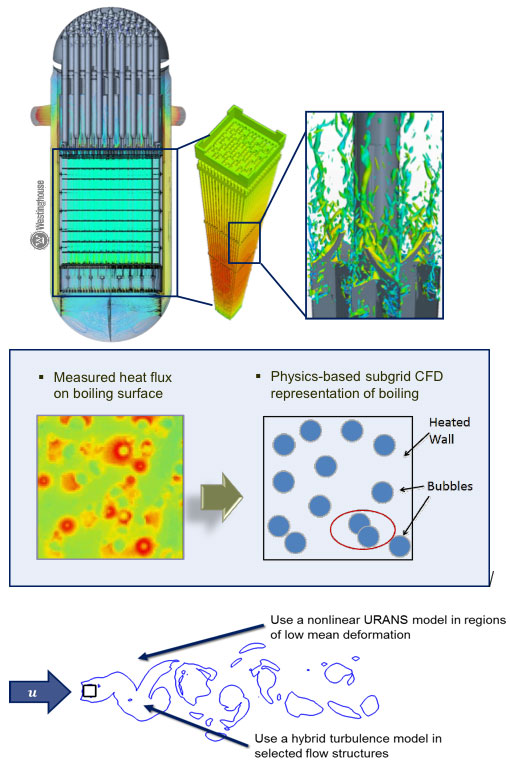}
\caption{
Diagram of the various physics and scales to simulate for turbulent flow within a nuclear reactor core -- top-to-bottom zooming in reveals finer and finer scales, each with different physics to model. AI-driven computational fluid dynamics (CFD) can radically improve the resolution and efficiency of such simulations \cite{Kochkov2021MachineLC,Hennigh2020NVIDIASA}.
}
\label{fig:nuclear}
\end{figure}

\paragraph{\textit{Multi-physics}}
Complex real-world problems require solutions that span a multitude of physical phenomena, which often can only be solved using simulation techniques that cross several engineering disciplines. Almost all practical problems in fluid dynamics involve the interaction between a gas and/or liquid with a solid object, and include a range of associated physics including heat transfer, particle transport, erosion, deposition, flow-induced-stress, combustion and chemical reaction. A multi-physics environment is defined by coupled processes or systems involving more than one simultaneously occurring physical fields or phenomena. 
Such an environment is typically described by multiple partial differential equations (PDEs), and tightly coupled such that solving them presents significant challenges with nonlinearities and time-stepping.
In general, the more interacting physics in a simulation, the more costly the computation.

\paragraph{\textit{Multi-scale}}
Ubiquitous in science and engineering, \textit{cascades-of-scales} involve more than two scales with long-range spatiotemporal interactions (that often lack self-similarity and proper closure relations).
In the context of biological and behavioral sciences, for instance, multi-scale modeling applications range from the molecular, cellular, tissue, and organ levels all the way to the population level; multi-scale modeling can enable researchers to probe biologically relevant phenomena at smaller scales and seamlessly embed the relevant mechanisms at larger scales to predict the emergent dynamics of the overall system \cite{Hunt2018TheSO}.
Domains such as energy and synthetic biology require engineering materials at the nanoscale, optimizing multi-scale processes and systems at the macroscale, and even the discovery of new governing physicochemical laws \textit{across} scales.
These scientific drivers call for a deeper, broader, and more integrated understanding of common multi-scale phenomena and scaling cascades.
In practice, multi-scale modeling is burdened by computational inefficiency, especially with increasing complexity and scales; it is not uncommon to encounter ``hidden'' or unknown physics of interfaces, inhomogeneities, symmetry-breaking and other singularities.

Methods for utilizing information across scales (and physics that vary across space and time) are often needed in real-world settings, and with data from various sources: \textit{Multi-fidelity modeling} aims to synergistically combine abundant, inexpensive, low-fidelity data and sparse, expensive, high-fidelity data from experiments and simulations. Multi-fidelity modeling is often useful in building efficient and robust surrogate models (which we detail in the surrogate motif section later) \cite{Alber2019IntegratingML} -- some examples include simulating the mixed convection flow past a cylinder~\cite{Perdikaris2016ModelIV} and cardiac electrophysiology~\cite{Costabal2019MultifidelityCU}.

In computational fluid dynamics (CFD), classical methods for multi-physics and multi-scale simulation (such as finite elements and pseudo-spectral methods) are only accurate if flow feature are all smooth, and thus meshes must resolve the smallest features. Consequently, direct numerical simulation for real-world systems such as climate and jet physics are impossible. It is common to use smoothed versions of the Navier Stokes equations to allow coarser meshes while sacrificing accuracy. Although successful in design of engines and turbo-machinery, there are severe limits to what can be accurately and reliably simulated -- the resolution-efficiency tradeoff imposes a significant bottleneck.
Methods for AI-driven acceleration and surrogate modeling could accelerate CFD and multi-physics multi-scale simulation by orders of magnitude. We discuss specific methods and examples below.

\paragraph{\textit{Physics-informed ML}}
The newer class of \textit{physics-informed machine learning} methods integrate mathematical physics models with data-driven learning. More specifically, making an ML method physics-informed amounts to introducing appropriate observational, inductive, or learning biases that can steer or constrain the learning process to physically consistent solutions \cite{Karniadakis2021PhysicsinformedML}:
\begin{enumerate}
    \item \textit{Observational biases} can be introduced via data that allows an ML system to learn functions, vector fields, and operators that reflect physical structure of the data.
    \item \textit{Inductive biases} are encoded as model-based structure that imposes prior assumptions or physical laws, making sure the physical constraints are strictly satisfied.
    \item \textit{Learning biases} force the training of an ML system to converge on solutions that adhere to the underlying physics, implemented by specific choices in loss functions, constraints, and inference algorithms.
\end{enumerate}

\begin{figure}[ht]
\centering
\includegraphics[width=0.9\linewidth]{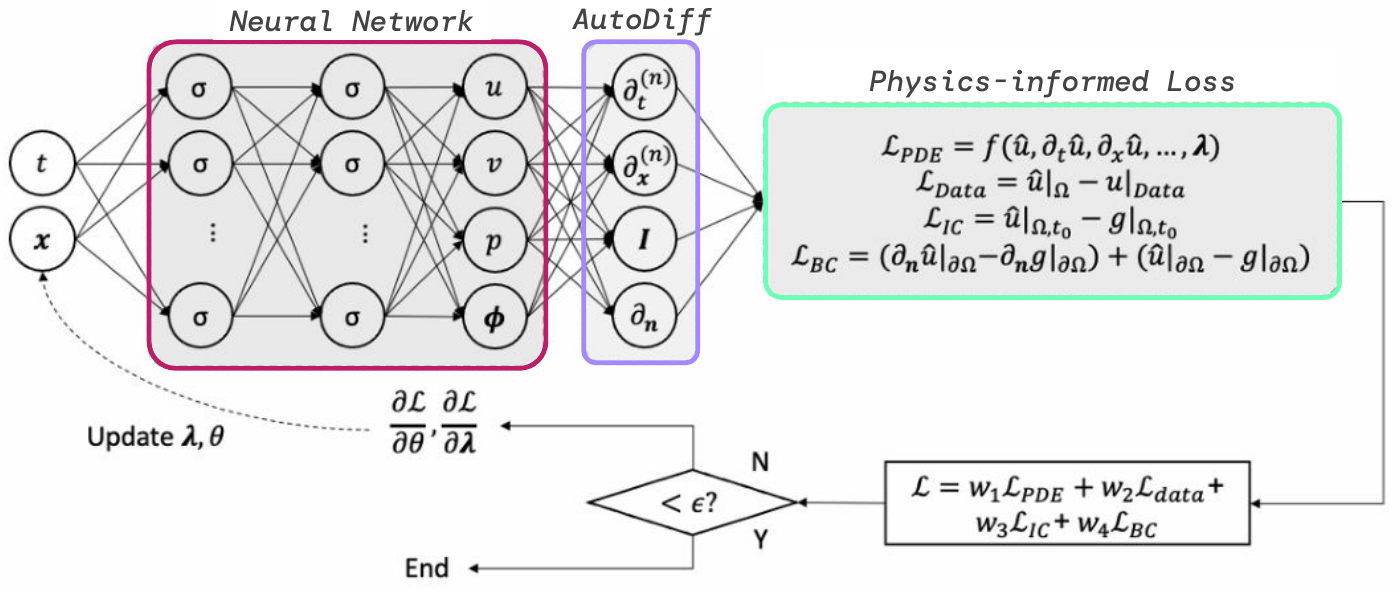}
\caption{
Diagram of a physics-informed neural network (PINN), where a fully-connected neural network (red), with time and space coordinates $(t,\textbf{x})$ as inputs, is used to approximate the multi-physics solutions $\hat{u} = [u,v,p,\phi]$. The derivatives of $\hat{u}$ with respect to the inputs are calculated using automatic differentiation (purple; autodiff is discussed later in the SI engine section) and then used to formulate the residuals of the governing equations in the loss function (green), that is generally composed of multiple terms weighted by different coefficients.
By minimizing the physics-informed loss function (i.e., the right-to-left process) we simultaneously learn the parameters of the neural network $\theta$ and the unknown PDE parameters $\lambda$. (Figure reproduced from Ref.~\cite{Cai2021PhysicsinformedNN})
}
\label{fig:pinn}
\end{figure}

A common problem template involves extrapolating from an initial condition obtained from noisy experimental data, where a governing equation is known for describing at least some of the physics. An ML method would aim to predict the latent solution $u(t, x)$ of a system at later times $t > 0$ and propagate the uncertainty due to noise in the initial data. A common use-case in scientific computing is reconstructing a flow field from scattered measurements (e.g., particle image velocimetry data), and using the governing Navier–Stokes equations to extrapolate this initial condition in time. 

For fluid flow problems and many others, Gaussian processes (GPs) present a useful approach for capturing the physics of dynamical systems.
A GP is a Bayesian nonparametric machine learning technique that provides a flexible prior distribution over functions, enjoys analytical tractability, defines kernels for encoding domain structure, and has a fully probabilistic workflow for principled uncertainty reasoning \cite{Rasmussen2006GaussianPF, Ghahramani2015ProbabilisticML}.
For these reasons GPs are used widely in scientific modeling, with several recent methods more directly encoding physics into GP models: \textit{Numerical GPs} have covariance functions resulting from temporal discretization of time-dependent partial differential equations (PDEs) which describe the physics \cite{Raissi2018NumericalGP, Raissi2018HiddenPM}, modified Mat\'{e}rn GPs can be defined to represent the solution to stochastic partial differential equations \cite{Lindgren2011AnEL} and extend to Riemannian manifolds to fit more complex geometries \cite{Borovitskiy2020MaternGP}, and the \textit{physics-informed basis-function GP} derives a GP kernel directly from the physical model \cite{Hanuka2020PhysicsinformedGP} -- the latter method we elucidate in an experiment optimization example in the surrogate modeling motif.

With the recent wave of deep learning progress, research has developed methods for building the three physics-informed ML biases into nonlinear regression-based physics-informed networks.
More specifically, there is increasing interest in \textit{physics-informed neural networks (PINNs)}: deep neural nets for building surrogates of physics-based models described by partial differential equations (PDEs) \cite{Raissi2019PhysicsinformedNN, Zhu2019PhysicsConstrainedDL}.
Despite recent successes (with some examples explored below) \cite{Raissi2018DeepLO, Zhang2019QuantifyingTU, Meng2020ACN, Berg2018AUD, Sirignano2018DGMAD, Sun2020SurrogateMF}, PINN approaches are currently limited to tasks that are characterized by relatively simple and well-defined physics, and often require domain-expert craftsmanship.
Even so, the characteristics of many physical systems are often poorly understood or hard to implicitly encode in a neural network architecture.

\begin{figure}[!ht]
\centering
\includegraphics[width=0.7\linewidth]{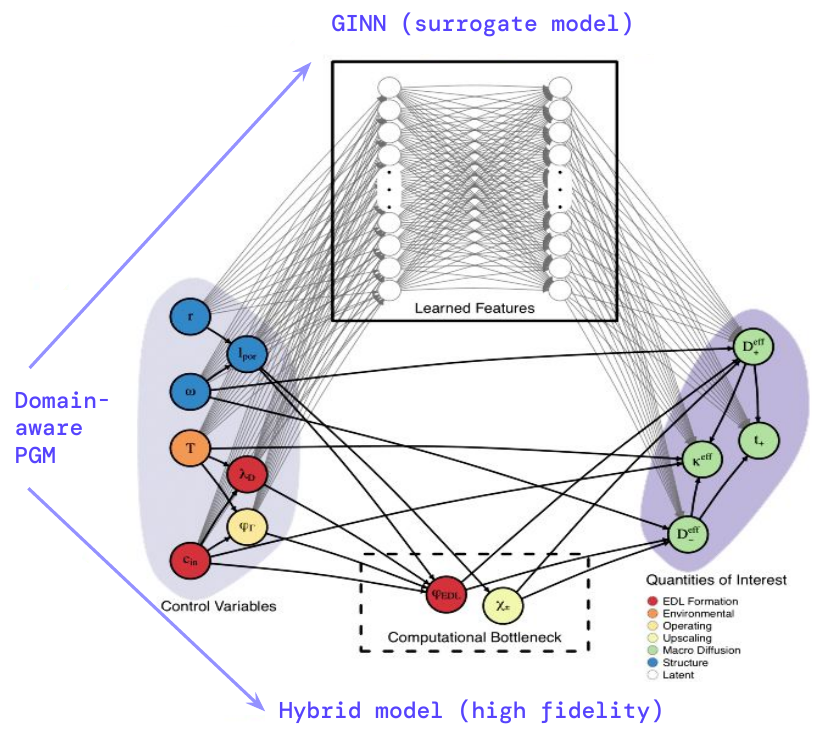}
\caption{
Graph-Informed Neural Networks (GINNs) \cite{Hall2021GINNsGN} provide a computational advantage while maintaining the advantages of structured modeling with PGMs, by replacing computational bottlenecks with NN surrogates. Here a PGM encoding structured priors serves as input to both a Bayesian Network PDE (lower route) and a GINN (upper) for a homogenized model of ion diffusion in supercapacitors. A simple fully-connected NN is pictured, but in principle any architecture can work, for instance physics-informed methods that further enforce physical constraints.
}
\label{fig:ginn}
\end{figure}

Probabilistic graphical models (PGMs) \cite{Koller2009ProbabilisticGM}, on the other hand, are useful for encoding \textit{a priori} structure, such as the dependencies among model variables in order to maintain physically sound distributions. This is the intuition behind the promising direction of \textit{graph-informed neural networks (GINN)} for multi-scale physics \cite{Hall2021GINNsGN}. There are two main components of this approach (shown in Fig. \ref{fig:ginn}): First, embedding a PGM into the physics-based representation to encode complex dependencies among model variables that arise from domain-specific information and to enable the generation of physically sound distributions. Second, from the embedded PGM identify computational bottlenecks intrinsic to the underlying physics-based model and replace them with an efficient NN surrogate. 
The hybrid model thus encodes a domain-aware physics-based model that synthesizes stochastic and multi-scale modeling, while  computational bottlenecks are replaced by a fast surrogate NN whose supervised learning and prediction are further informed by the PGM (e.g., through structured priors). With significant computational advantages from surrogate modeling within GINN, we further explore this approach in the surrogate modeling motif later.

Modeling and simulation of complex nonlinear multi-scale and multi-physics systems requires the inclusion and characterization of uncertainties and errors that enter at various stages of the computational workflow.
Typically we're concerned with two main classes of uncertainties in ML: \textit{aleatoric} and \textit{epistemic} uncertainties. 
The former quantifies system stochasticity such as observation and process noise, and the latter is model-based or subjective uncertainty due to limited data.
In environments of multiple scales and multiple physics, one should consider an additional type of uncertainty due the randomness of parameters of stochastic physical systems (often described by stochastic partial- or ordinary- differential equations (SPDEs, SODEs)).
One can view this type of uncertainty arising from the computation of a sufficiently well-posed deterministic problem, in contrast to the notion of epistemic or aleatoric uncertainty quantification.
The field of \textit{probabilistic numerics} \cite{Hennig2015ProbabilisticNA} makes a similar distinction, where the use of probabilistic modeling is to reason about uncertainties that arise strictly from the lack of information inherent in the solution of intractable problems such as quadrature methods and other integration procedures. 
In general the probabilistic numeric viewpoint provides a principled way to manage the parameters of numerical procedures.
We discuss more on probabilistic numerics and uncertainty reasoning later in the SI themes section.
The GINN can quantify uncertainties with a high degree of statistical confidence, while Bayesian analogs of PINNs  are a work in progress \cite{Yang2021BPINNsBP}--one cannot simply plug in MC-dropout or other deep learning uncertainty estimation methods.
The various uncertainties may be quantified and mitigated with methods that can utilize data-driven learning to inform the original systems of differential equations. We define this class of physics-\textit{infused} machine learning later in this section and in the next motif.


The synergies of mechanistic physics models and data-driven learning are brought to bear when physics-informed ML is built with differentiable programming (one of the engine motifs), which we explore in the first example below.

\subsection*{Examples}\label{sec:PhysScale_ex}

\paragraph{Accelerated CFD via physics-informed surrogates and differentiable programming}
Kochkov et al. \cite{Kochkov2021MachineLC} look to bring the advantages of semi-mechanistic modeling and differentiable programming to the challenge of complex computational fluid dynamics (CFD). The Navier Stokes (NS) equations describe fluid dynamics well, yet in cases of multiple physics and complex dynamics, solving the equations at scale is severely limited by the computational cost of resolving the smallest spatiotemporal features. Approximation methods can alleviate this burden, but at the cost of accuracy. In Kochkov et al, the components of traditional fluids solvers most affected by the loss of resolution are replaced with better performing machine-learned alternatives (as presented in the semi-mechanistic modeling section later). This AI-driven solver algorithm is represented as a differentiable program with the neural networks and the numerical methods written in the JAX framework \cite{jax2018github}. JAX is a leading framework for differentiable programming, with reverse-mode automatic differentiation that allows for end-to-end gradient based optimization of the entire programmed algorithm. In this CFD use-case, the result is an algorithm that maintains accuracy while using 10x coarser resolution in each dimension, yielding an ~80-fold improvement in computation time with respect to an advanced numerical method of similar accuracy.


Related approaches implement PINNs without the use of differentiable programming, such as TF-Net for modeling turbulent flows with several specially designed U-Net deep learning architectures \cite{Wang2020TowardsPD}.
A promising direction in this and other spatiotemporal use-cases is \textit{neural operator learning}: using NNs to learn mesh-independent, resolution-invariant solution operators for PDEs. To achieve this, Li et al. \cite{Li2020FourierNO} use a Fourier layer that implements a Fourier transform, then a linear transform, and an inverse Fourier transform for a convolution-like operation in a NN. In a Bayesian inverse experiment, the Fourier neural operator acting as a surrogate can draw MCMC samples from the posterior of initial NS vorticity given sparse, noisy observations in 2.5 minutes, compared to 18 hours for the traditional solver.

\paragraph{Multi-physics and HPC simulation of blood flow in an intracranial aneurysm}

\begin{figure}[!t]
\centering
\includegraphics[width=1.0\linewidth]{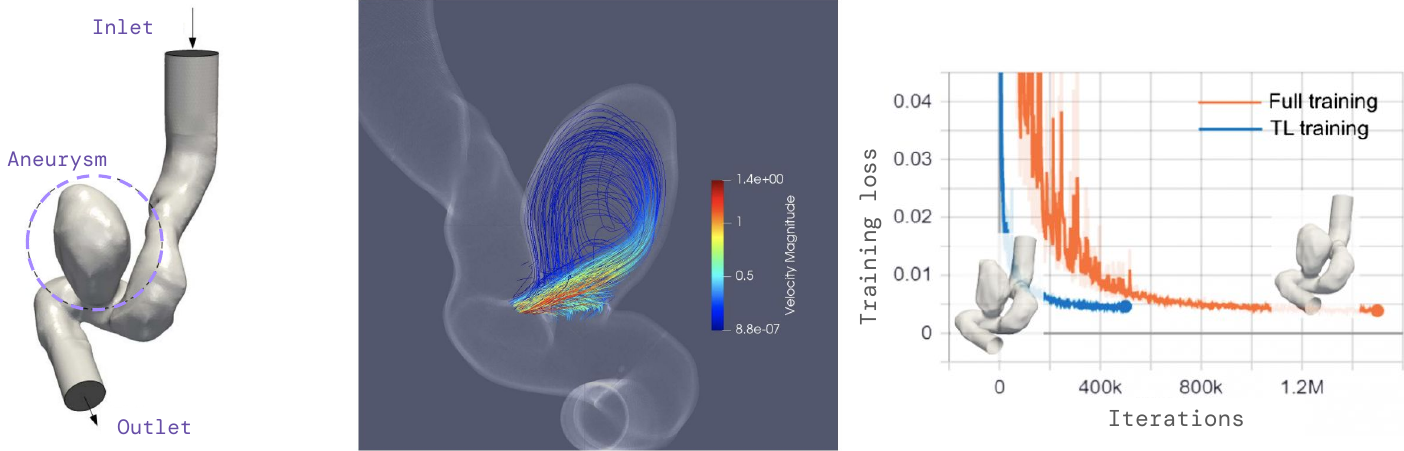}
\caption{
Modulus simulation results for the aneurysm problem \cite{Hennigh2020NVIDIASA}. Left: patient-specific geometry of an intracranial aneurysm. Center: Streamlines showing accurate flow field simulation inside the aneurysm sac. Right: Transfer learning within the NN-based simulator accelerates the computation of patient-specific geometries.
}
\label{fig:simnet}
\end{figure}

\textit{Modulus} is an AI-driven multi-physics simulation framework based on neural network solvers--more specifically it approximates the solution to a PDE by a neural network \cite{Hennigh2020NVIDIASA}. Modulus improves on previous NN-solvers to take on the challenge of gradients and discontinuities introduced by complex geometries or physics. The main novelties are the use of Signed Distance Functions for loss weighting, and integral continuity planes for flow simulation. An intriguing real-world use case is simulating the flow inside a patient-specific geometry of an aneurysm, as shown in Fig.\ref{fig:simnet}. It is particularly challenging to get the flow field to develop correctly, especially inside the aneurysm sac. Building Modulus to support multi-GPU and multi-node scaling provides the computational efficiency necessary for such complex geometries.
There's also optimization over repetitive trainings, such as training for surrogate-based design optimization or uncertainty quantification, where transfer learning reduces the time to convergence for neural network solvers. Once a model is trained for a single geometry, the trained model parameters are transferred to solve a different geometry, without having to train on the new geometry from scratch. As shown in Fig.\ref{fig:simnet}, transfer learning accelerates the patient-specific intracranial aneurysm simulations.  

Raissi et al.~\cite{Raissi2020HiddenFM} similarly approach the problem of 3D physiologic blood flow in a patient-specific intracranial aneurysm, but implementing a physics-informed NN technique: the Hidden Fluid Mechanics approach that uses autodiff (i.e. within differentiable programming) to simultaneously exploit information from the Navier Stokes equations of fluid dynamics and the information from flow visualization snapshots. Continuing this work has high potential for the robust and data-efficient simulation in physical and biomedical applications. Raissi et al. effectively solve this as an inverse problem using blood flow data, while Modulus approaches this as a forward problem without data. Nonetheless Modulus has potential for solving inverse problems as well (within the inverse design workflow of Fig. \ref{fig:inverse-batteries} for example).

\begin{figure}[!ht]
\centering
\includegraphics[width=0.8\linewidth]{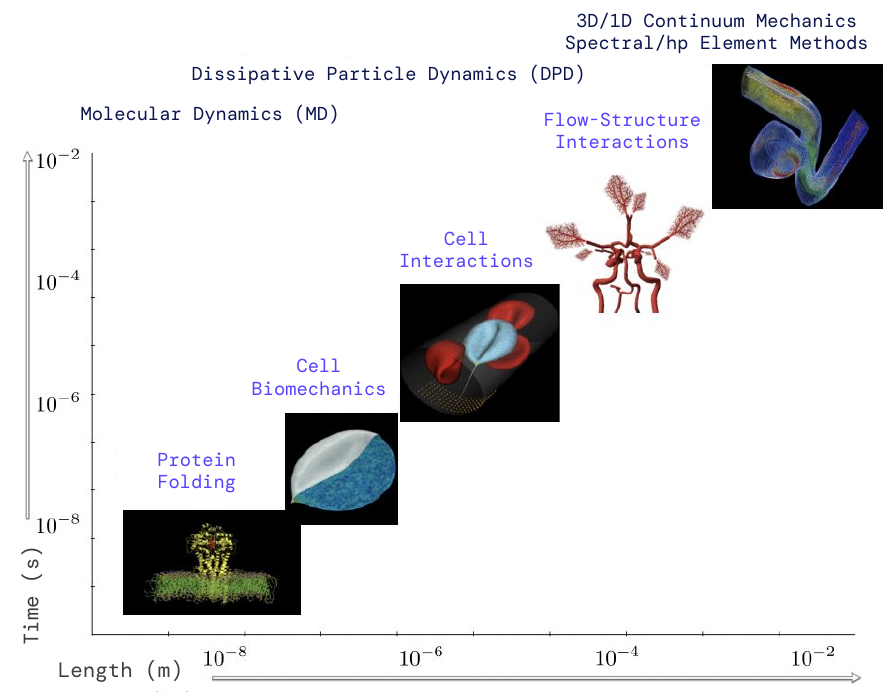}
\caption{
A cascade of spatial and temporal scales governing key biophysical
mechanisms in brain blood flow requires effective multi-scale modeling and simulation techniques (inspired by \cite{Perdikaris2016MultiscaleMA}). Listed are the standard modeling approaches for each spatial and temporal regime, where an increase in computational demands is inevitable as one pursues an integrated resolution of interactions in finer and finer scales. Modulus \cite{Hennigh2020NVIDIASA}, JAX MD \cite{Schoenholz2020JAXMA}, and related works with the motifs will help mitigate these computational demands and more seamlessly model dynamics across scales.
}
\label{fig:brainbloodflow}
\end{figure}


\hfill \break
\paragraph{Learning quantum chemistry}

Using modern algorithms and supercomputers, systems containing thousands of interacting ions and electrons can now be described using approximations to the physical laws that govern the world on the atomic scale, namely the Schrödinger equation \cite{Butler2018MachineLF}. 
Chemical-simulation can allow for the properties of a compound to be anticipated (with reasonable accuracy) before synthesizing it in the laboratory. 
These computational chemistry applications range from catalyst development for greenhouse gas conversion, materials discovery for energy harvesting and storage, and computer-assisted drug design \cite{Walsh2013ComputationalAT}. 
The potential energy surface is the central quantity of interest in the modeling of molecules and materials, computed by approximations to the time-independent Schrödinger equation. Yet there is a computational bottleneck: high-level wave function methods have high accuracy, but are often too slow for use in many areas of chemical research. The standard approach today is density functional theory (DFT)~\cite{Lejaeghere2016ReproducibilityID}, which can yield results close to chemical accuracy, often on the scale of minutes to hours of computational time. 
DFT has enabled the development of extensive databases that cover the calculated properties of known and hypothetical systems, including organic and inorganic crystals, single molecules, and metal alloys \cite{Hachmann2011TheHC,Jain2013CommentaryTM,Calderon2015TheAS}.
In high-throughput applications, often used methods include force-field (FF) and semi-empirical quantum mechanics (SEQM), with runtimes on the order of fractions of a second, but at the cost of reliability of the predictions \cite{Christensen2021OrbNetDA}. 
ML methods can potentially provide accelerated solvers without loss of accuracy, but to be reliable in the chemistry space the model must capture the underlying physics, and well-curated training data covering the relevant chemical problems must be available -- Butler et al. \cite{Butler2018MachineLF} provide a list of publicly accessible structure and property databases for molecules and solids.
Symmetries in geometries or physics, or \textit{equivariance}, defined as the property of being independent of the choice of reference frame, can be exploited to this end. 
For instance, a message passing neural network~\cite{Gilmer2017NeuralMP} called OrbNet encodes a molecular system in graphs based on features from a quantum calculation that are low cost by implementing symmetry-adapted atomic orbitals \cite{Qiao2020OrbNetDL,Christensen2021OrbNetDA}. The authors demonstrate effectiveness of their equivariant approach on several organic and biological chemistry benchmarks, with accuracies on par to that of modern DFT functionals, providing a far more efficient drop-in replacement for DFT energy predictions. 
Additional approaches that constraining NNs to be symmetric under these geometric operations have been successfully applied in molecular design \cite{Simm2021SymmetryAwareAF} and quantum chemistry \cite{Qiao2021UNiTEUN}.


\subsection*{Future directions}\label{sec:PhysScale_future}

In general, machine learned models are ignorant of fundamental laws of physics, and can result in ill-posed problems or non-physical solutions. This effect is exacerbated when modeling the interplay of multiple physics or cascades of scales. Despite recent successes mentioned above, there is much work to be done for multi-scale and multi-physics problems. For instance, PINNs can struggle with high-frequency domains and frequency bias \cite{Basri2019TheCR}, which is particularly problematic in multi-scale problems \cite{Wang2020WhenAW}. Improved methods for learning multiple physics simultaneously are also needed, as the training can be prohibitively expensive -- for example, a good approach with current tooling is training a model for each field separately and subsequently learning the coupled solutions through either a parallel or a serial architecture using supervised learning based on additional data for a specific multi-physics problem.

In order to approach these challenges in a collective and reproducible way, there is need to create \textit{open benchmarks for physics-informed ML}, much like other areas of ML community such as computer vision and natural language processing.
Yet producing quality benchmarks tailored for physics-informed ML can be more challenging: 
\begin{enumerate}
    \item To benchmark physics-informed ML methods, we additionally need the proper parameterized physical models to be explicitly included in the databases.
    \item Many applications in physics and chemistry require full-field data, which cannot be obtained experimentally and/or call for significant compute.
    \item Often different, problem-specific, physics-based evaluation methods are necessary, for example the metrics proposed in \cite{Ltjens2021PhysicallyConsistentGA} for scoring physical consistency and \cite{Vlachas2020BackpropagationAA} for scoring spatiotemporal predictions.
\end{enumerate}

An overarching benchmarking challenge but also advantageous constraint is the multidisciplinary nature of physics-informed ML: there must be multiple benchmarks in multiple domains, rather than one benchmark to rule them all, which is a development bias that ImageNet has put on the computer vision field the past decade.
And because we have domain knowledge and numerical methods for the underlying data generating mechanisms and processes in physical and life sciences, we have an opportunity to quantify robustly the characteristics of datasets to better ground the performances of various models and algorithms. This is contrast to the common practice of na\"ive data gathering to compile massive benchmark datasets for deep learning -- for example, scraping the internet for videos to compose a benchmark dataset for human action recognition (\href{https://deepmind.com/research/open-source/kinetics}{deepmind.com/research/open-source/kinetics}) -- where not only are the underlying statistics and causal factors \textit{a priori} unknown, the target variables and class labels are non-trivial to define and can lead to significant ethical issues such as dataset biases that lead to model biases, which in some cases can propagate harmful assumptions and stereotypes.

Further we propose to broaden the class of methods beyond \textit{physics-informed}, to \textbf{\textit{physics-infused machine learning}}. The former is unidirectional (physics providing constraints or other information to direct ML methods), whereas the latter is bidirectional, including approaches that can better synergize the two computational fields. For instance, for systems with partial information, methods in physics-infused ML can potentially compliment known physical models to learn missing or misunderstood components of the systems. We specifically highlight one approach named \textit{Universal Differential Equations} \cite{Rackauckas2020UniversalDE} in the surrogate modeling motif next.


Physics-infused ML can enable many new simulation tools and use-cases because of this ability to integrate physical models and data within a differentiable software paradigm.
JAX and Modulus are nice examples, each enabling physics-informed ML methods for many science problems and workflows. Consider, for instance, the JAX fluid dynamics example above: another use-case in the same DP framework is JAX MD \cite{Schoenholz2020JAXMA} for performing differentiable physics simulations with a focus on molecular dynamics.

Another exciting area of future multi-physics multi-scale development is inverse problem solving.
We already mentioned the immense acceleration in the CFD example above with Fourier Neural Operators \cite{Li2020FourierNO}. Beyond efficiency gains, physics-infused ML can take on applications with inverse and ill-posed problems which are either difficult or impossible to solve with conventional approaches, notably quantum chemistry: 
A recent approach called FermiNet \cite{Pfau2019AbInitioSO} takes a dual physics-informed learning approach to solving the many-electron Schrodinger equation, where both inductive bias and learning bias are employed.
The advantage of physics-infused ML here is eliminating extrapolation problems with the standard numerical approach, which is a common source of error in computational quantum chemistry.
In other domains such as biology, biomedicine, and behavioral sciences, focus is shifting from solving forward problems based on sparse data towards solving inverse problems to explain large datasets \cite{Alber2019IntegratingML}. The aim is to develop multi-scale simulations to infer the behavior of the system, provided access to massive amounts of observational data, while the governing equations and their parameters are not precisely known.
We further detail the methods and importance of inverse problem solving with SI later in the Discussion section.

\hfill \break
\subsection{2. SURROGATE MODELING \& EMULATION}

A \textit{surrogate model} is an approximation method that mimics the behavior of an expensive computation or process. For example, the design of an aircraft fuselage includes computationally intensive simulations with numerical optimizations that may take days to complete, making design space exploration, sensitivity analysis, and inverse modeling infeasible. In this case a computationally efficient surrogate model can be trained to represent the system, learning a mapping from simulator inputs to outputs.
And Earth systems models (ESMs), for example, are extremely computationally expensive to run due to the large range of spatial and temporal scales and large number of processes being modeled. ESM surrogates can be trained on a few selected samples of the full, expensive simulations using supervised machine learning tools.
In this simulation context, the aim of surrogate modeling (or \textit{statistical emulation}) is to replace simulator code with a machine learning model (i.e., \textit{emulator}) such that running the ML model to infer the simulator outputs is more efficient than running the full simulator itself. An emulator is thus a model of a model: a statistical model of the simulator, which is itself a mechanistic model of the world.

\begin{figure}[!ht]
\centering
\includegraphics[width=1.0\linewidth]{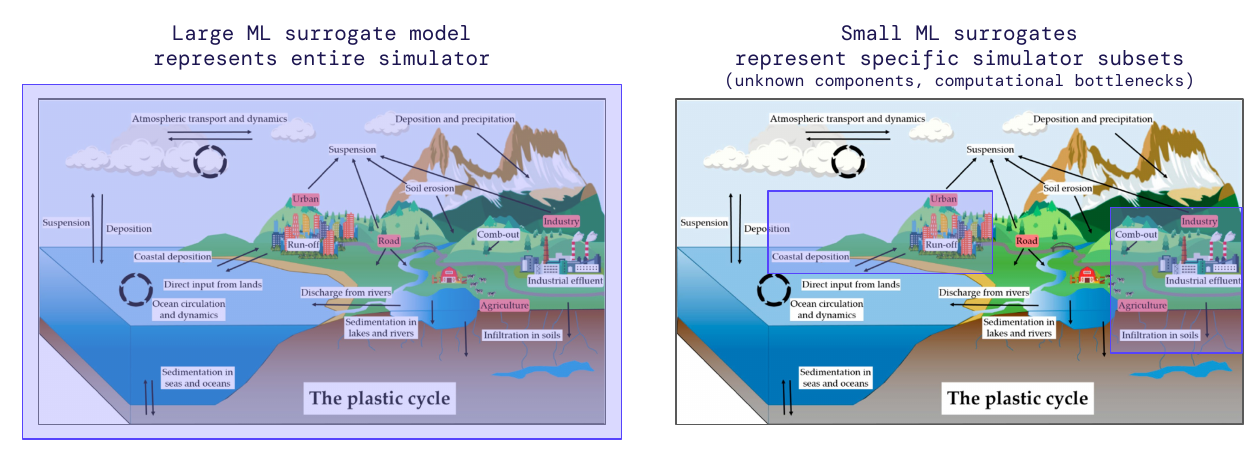}
\caption{
An example Earth system model (ESM) for the plastic cycle, with two variations of ML surrogates (purple screens): On the left, an ML surrogate learns the whole model. A variety of physics-infused ML methods can be applied -- training is relatively straightforward because the loss function only has neural networks, and the trained network can be used towards inverse problem solving. However, we now have a black-box simulator: there is no interpretability of the trained network, and we cannot utilize the mechanistic components (i.e. differential equations of the simulator) for scientific analyses. On the right, two unknown portions of the model are learned by NN surrogates, while the remaining portions are represented by known mechanistic equations -- possible with surrogate modeling approaches like UDE and GINN. This ``partial surrogate'' case has several advantages: the surrogate's number of parameters is reduced and thus the network is more stable (similar logic holds for nonparametric Gaussian process surrogate models), and the simulator retains all structural knowledge and the ability to run numerical analysis. The main challenge is that backpropagation of arbitrary scientific simulators is required, which we can address with the engine motif differentiable programming, producing learning gradients for arbitrary programs. The computational gain associated with the use of a hybrid surrogate-simulator cascades into a series of additional advantages including the possibility of simulating more scenarios towards counterfactual reasoning and epistemic uncertainty estimates, decreasing grid sizes, or exploring finer-scale parameterizations \cite{Rackauckas2020GeneralizedPL, Goldstein2015MachineLC}.
}
\label{fig:esm}
\end{figure}

For surrogate modeling in the sciences, non-linear, nonparametric Gaussian processes (GP) \cite{Rasmussen2006GaussianPF} are typically used because of their flexibility, interpretability, and accurate uncertainty estimates \cite{Kennedy2001BayesianCO}. Although traditionally limited to smaller datasets because of $O(N^3)$ computational cost of training (where $N$ is the number of training data points), much work on reliable GP sparsification and approximation methods make them viable for real-world use \cite{Titsias2009VariationalLO,Hensman2013GaussianPF,Wilson2015ThoughtsOM,Izmailov2018ScalableGP}.

Neural networks (NNs) can also be well-suited to the surrogate modeling task as function approximation machines: A feedforward network deﬁnes a mapping $y=f(x;\theta)$ and learns the value of the parameters $\theta$ that result in the best function approximation. The \textit{Universal Approximation Theorem} demonstrates that sufficiently large NNs can approximate any nonlinear function with a finite set of parameters \cite{Lin2018ResNetWO,Winkler2017PerformanceOD}. Although sufficient to represent any function, a NN layer may be unfeasibly large such that it may fail to learn and generalize correctly \cite{Goodfellow2015DeepL}. Recent work has shown that a NN with an infinitely wide hidden layer converges to a GP, representing the normal distribution over the space of functions.

In Fig.~\ref{fig:esm} we show an example of how a NN surrogate can be used as an emulator that encapsulates either the entire simulator or a specific part of the simulator. In the former, training is relatively straightforward because the loss function only has neural networks, and the trained network can be used towards inverse problem solving. However, we now have a black-box simulator: there is no interpretability of the trained network, and we cannot utilize the mechanistic components (i.e. differential equations of the simulator) for scientific analyses. In the case of the partial surrogate we have several advantages: the surrogate’s number of parameters is reduced and thus the network is more stable (similar logic holds for nonparametric GP surrogate models), and the simulator retains all structural knowledge and the ability to run numerical analysis. Yet the main challenge is that backpropagation of arbitrary scientific simulators is required -- thus the significance of the differentiable programming motif we discussed earlier. The computational gain associated with the use of a hybrid surrogate-simulator cascades into a series of additional advantages including the possibility of simulating more scenarios towards counterfactual reasoning and epistemic uncertainty estimates, decreasing grid sizes, or exploring finer-scale parameterizations \cite{Rackauckas2020GeneralizedPL,Goldstein2015MachineLC}.


\paragraph{\textit{Semi-mechanistic modeling}}
It follows from the Universal Approximation Theorem that an NN can learn to approximate any sufficiently regular differential equation. The approach of recent neural-ODE methods \cite{Chen2018NeuralOD} is to learn to approximate differential equations directly from data, but these can perform poorly when required to extrapolate \cite{Yan2020OnRO}. More encouraging for emulation and scientific modeling, however, is to directly utilize mechanistic modeling simultaneously with NNs (or more generally, universal approximator models) in order to allow for arbitrary data-driven model extensions. The result is a \textit{semi-mechanistic} approach, more specifically \textit{Universal Differential Equations (UDE)} \cite{Rackauckas2020UniversalDE} where part of the differential equation contains a universal approximator model -- we’ll generally assume an NN is used for UDE in this paper, but other options include GP, Chebyshev expansion, or random forest.

UDE augments scientific models with machine-learnable structures for scientifically-based learning. This is very similar in motivation to the physics-informed neural nets (PINN) we previously discussed, but the implementation in general has a key distinction: a PINN is a deep learning model that become physics-informed because of some added physics bias (observational, inductive, or learning), whereas a UDE is a differential equation with one or more mechanistic components replaced with a data-driven model -- this distinction is why we earlier defined physics-infused ML as the \textit{bidirectional} influence of physics and ML.


\paragraph{\textit{Bayesian optimal experiment design}}

A key aspect that makes emulators useful in scientific endeavors is that they allow us to reason probabilistically about \textit{outer-loop decisions} such as optimization \cite{Shahriari2016TakingTH}, data collection \cite{Krause2007NonmyopicAL}, and to use them to explain how uncertainty propagates in a system \cite{Bilionis2015BayesianUP}. 

Numerous challenges in science and engineering can be framed as optimization tasks, including
the maximization of reaction yields, the optimization of molecular and materials properties, and the
fine-tuning of automated hardware protocols \cite{Aldeghi2021GolemAA}.
When we seek to optimize the parameters of a system with an expensive cost function $f$, we look to employ \textit{Bayesian optimization (BO)} \cite{McIntire2016SparseGP,Frazier2018ATO,Snoek2012PracticalBO, Shahriari2016TakingTH} to efficiently explore the search space of solutions with a probabilistic surrogate model $\hat{f}$ rather than experimenting with the real system. Gaussian process models are the most common surrogates due to their flexible, nonparametric behavior. Various strategies to explore-exploit the search space can be implemented with acquisition functions to efficiently guide the BO search by estimating the utility of evaluating $f$ at a given point (or parameterization).
Often in science and engineering settings this provides the domain experts with a few highly promising candidate solutions to then try on the real system, rather than searching the intractably large space of possibilities themselves -- for example, generating novel molecules with optimized chemical properties \cite{Griffiths2019ConstrainedBO,GmezBombarelli2018AutomaticCD}, materials design with expensive physics-based simulations \cite{Zhang2020BayesianOF}, and design of aerospace engineering systems \cite{Poloczek2018AdvancesIB}.

Similarly, scientists can utilize BO for designing experiments such that the outcomes will be as informative as possible about the underlying process. 
\textit{Bayesian optimal experiment design (BOED)} is a powerful mathematical framework for tackling this problem \cite{Lindley1956OnAM, Chaloner1995BayesianED, Foster2021DeepAD, zhang2021scalable}, and can be implemented across disciplines, from bioinformatics \cite{Vanlier2012ABA} to pharmacology \cite{Lyu2018BayesianAD} to physics \cite{Dushenko2020SequentialBE} to psychology \cite{Myung2013ATO}.
In addition to \textit{design}, there are also \textit{control} methods in experiment optimization, which we detail in the context of a particle physics example below.


\subsection*{Examples}\label{sec:Surrogate_ex}

\paragraph{Simulation-based online optimization of physical experiments}

It is often necessary and challenging to design experiments such that outcomes will be as informative as possible about the underlying process, typically because experiments are costly or dangerous. 
Many applications such as nuclear fusion and particle acceleration call for online control and tuning of system parameters to deliver optimal performance levels -- i.e., the \textit{control} class of experiment design we introduced above.

In the case of particle accelerators, although physics models exist, there are often significant differences between the simulation and the real accelerator, so we must leverage real data for precise tuning. Yet we cannot rely on many runs with the real accelerator to tune the hundreds of machine parameters, and archived data does not suffice because there are often new machine configurations to try -- a control or tuning algorithm must robustly find the optimum in a complex parameter space with high efficiency.
With physics-infused ML, we can exploit well-verified mathematical models to learn approximate system dynamics from few data samples and thus optimize systems online and \textit{in silico}. It follows that we can additionally look to optimize new systems without prior data. 

For the online control of particle accelerators, Hanuka et al. \cite{Hanuka2020PhysicsinformedGP} develop the \textit{physics-informed basis-function GP}. To clarify what this model encompasses, we need to understand the several ways to build such a GP surrogate for BO of a physical system:
\begin{enumerate}
    \item Data-informed GP using real experimental data
    \item Physics-informed GP using simulated data
    \item Basis-function GP from deriving a GP kernel directly from the physical model
    \item Physics-informed basis-function GP as a combination of the above -- methods 2 and 3 were combined in Hanuka et al.
\end{enumerate}

The resulting physics-informed GP is more representative of the particle accelerator system, and performs faster in an online optimization task compared to routinely used optimizers (ML-based and otherwise).
Additionally, the method presents a relatively simple way to construct the GP kernel, including correlations between devices -- learning the kernel from simulated data instead of machine data is a form of kernel transfer learning, which can help with generalizability. Hanuka et al. interestingly point out that constructing the kernel from basis functions without using the likelihood function is a form of Gaussian process with \textit{likelihood-free inference}, which is the regime of problems that simulation-based inference is designed for (i.e. the motif we discuss next).

This and similar physics-informed methods are emerging as a powerful strategy for \textit{in silico} optimization of expensive scientific processes and machines, and further to enable scientific discovery by means of autonomous experimentation. For instance, the recent Gemini \cite{Hickman2021GeminiDB} and Golem \cite{Aldeghi2021GolemAA} molecular experiment optimization algorithms, which are purpose-built for automated science workflows with SI: using surrogate modeling techniques for proxying expensive chemistry experiments, the BO and uncertainty estimation methods are designed for robustness to common scientific measurement challenges such as input variability and noise, proxy measurements, and systematic biases. Similarly, Shirobokov et al. \cite{shirobokov2020blackbox} propose a method for gradient-based optimization of black-box simulators using local generative surrogates that are trained in successive local neighborhoods of the parameter space during optimization, and demonstrate this technique in the optimization of the experimental design of the SHiP (Search for Hidden Particles) experiment proposed at CERN. These and other works of Al\'{a}n Aspuru-Guzik et al. are good sources to follow in this area of automating science.
There are potentially significant cause-effect implications to consider in these workflows, as we introduce in the causality motif later.

\paragraph{Multi-physics multi-scale surrogates for Earth systems emulation}

The climate change situation is worsening in accelerating fashion: the most recent decade (2010 to 2019) has been the costliest on record with the climate-driven economic damage reaching \$2.98 trillion-US, nearly double the decade 2000–2009 \cite{Bauer2021TheDR}. 
The urgency for climate solutions motivates the need for modeling systems that are computationally efficient and reliable, lightweight for low-resource use-cases, informative towards policy- and decision-making, and cyber-physical with varieties of sensors and data modalities. Further, models need to be integrated with, and workflows extended to, climate-dependent domains such as energy generation and distribution, agriculture, water and disaster management, and socioeconomics.
To this end, we and many others have been working broadly on \textit{Digital Twin Earth (DTE)}, a catalogue of ML and simulation methods, datasets, pipelines, and tools for Earth systems researchers and decision-makers.
In general, a \textit{digital twin} is a computer representation of a real-world process or system -- from large aircraft to individual organs. We define digital twin in the more precise sense of simulating the real physics and data-generating processes of an environment or system, with sufficient fidelity such that one can reliably run queries and experiments \textit{in silico}.

Some of the main ML-related challenges for DTE include integrating simulations of multiple domains, geographies, and fidelities; not to mention the need to integrate real and synthetic data, as well as data from multiple modalities (such as fusing Earth observation imagery with on-the-ground sensor streams).
SI methods play important roles in the DTE catalogue, notably the power of machine learned surrogates to accelerate existing climate simulators. Here we highlight one example for enabling lightweight, real-time simulation of coastal environments: Existing simulators for coastal storm surge and flooding are physics-based numerical models that can be extremely computationally expensive. Thus the simulators cannot be used for real-time predictions with high resolution, are unable to quantify uncertainties, and require significant computational infrastructure overhead only available to top national labs. To this end, Jiang et al.~\cite{Jiang2021DigitalTE} developed physics-infused ML surrogates to emulate several of the main coastal simulators worldwide, NEMO~\cite{Madec2008NEMOOE} and CoSMoS~\cite{Barnard2014DevelopmentOT}. Variations of the Fourier Neural Operator (FNO) \cite{Li2020FourierNO} (introduced in the multi-physics motif) were implemented to produce upwards of 100x computational efficiency on comparable hardware. 


This use-case exemplified a particularly thorny data preprocessing challenge that is commonplace working with spatiotemporal simulators and Digital Twin Earth: One of the coastal simulators to be emulated uses a standard grid-spaced representation for geospatial topology, which is readily computable with DFT in the FNO model, but another coastal simulator uses highly irregular grids that differ largely in scale, and further stacks these grids at varying resolutions. A preprocessing pipeline to regrid and interpolate the data maps was developed, along with substitute Fourier transform methods.
It is our experience that many applications in DTE call for tailored solutions such as this -- as a community we are lacking shared standards and formats for scientific data and code.

\paragraph{Hybrid PGM and NN for efficient domain-aware scientific modeling}

One can in general characterize probabilistic graphical models (PGM) \cite{Koller2009ProbabilisticGM} as structured models for encoding domain knowledge and constraints, contrasted with deep neural networks as data-driven function-approximators. The advantages of PGM have been utilized widely in scientific ML \cite{Friedman2004InferringCN,Beerenwinkel2015CancerEM,Leifer2008QuantumGM,Boixo2017SimulationOL}, and of course recent NN methods as discussed throughout this paper. \textit{Graph-Informed Neural Networks (GINNs)} \cite{Hall2021GINNsGN} are a new approach to incorporating the best of both worlds: PGMs incorporate expert knowledge, available data, constraints, etc. with physics-based models such as systems of ODEs and PDEs, while computationally intensive nodes in this hybrid model are replaced by learned features as NN surrogates.
GINNs are particularly suited to enhance the computational workflow for complex systems featuring intrinsic computational bottlenecks and intricate physical relations variables. 
Hall et al. demonstrate GINN towards simulation-based decision-making in
a multiscale model of electrical double-layer (EDL) supercapacitor dynamics.
The ability for downstream decision-making is afforded by robust and reliable sensitivity analysis (due to the probabilistic ML approach), and orders of magnitude more computational efficiency means many hypotheses can be simulated and predicted posteriors quantified.

\paragraph{Auto-emulator design with neural architecture search}

Kasim et al.~\cite{Kasim2020UpTT} look to recent advances in \textit{neural architecture search (NAS)} to automatically design and train a NN as an efficient, high-fidelity emulator, as doing this manually can be time-consuming and require significant ML expertise.
NAS methods aim to learn a network topology that can achieve the best performance on a certain task by searching over the space of possible NN architectures given a set of NN primitives -- see Elsken et al. \cite{Elsken2019NeuralAS} for a thorough overview.

The NAS results are promising for automated emulator construction: running on ten distinct scientific simulation cases, from fusion energy science \cite{Anirudh2020ImprovedSI,GaldnQuiroga2018BeamIonAD} to aerosol-climate \cite{Tegen2019TheGA} and oceanic \cite{Khatiwala2007ACF} modeling, the results are reliably accurate output simulations with NN-based emulators that run thousands to billions times faster than the originals, while also outperforming other NAS based emulation approaches as well as manual emulator design. For example, a global climate model (GCM) simulation tested normally takes about 1150 CPU-hours to run \cite{Tegen2019TheGA}, yet the emulator speedup is a factor of 110 million in direct comparison, and over 2 billion with a GPU — providing scientists with simulations on the order of seconds rather than days enables faster iteration of hypotheses and experiments, and potentially new experiments never before thought possible.

Also in this approach is a modified MC dropout method for estimating the predictive uncertainty of emulator outputs. Alternatively, we suggest pursuing Bayesian optimization-based NAS methods \cite{White2021BANANASBO} for more principled uncertainty reasoning. For example, the former can flag when an emulator architecture is overconfident in its predictions, while the latter can do that \textit{and} use the uncertainty values to dynamically adjust training parameters and search strategies.

The motivations of Kasim et al. for emulator-based accelerated simulation are same as we've declared above: enable rapid screening and ideas testing, and real-time prediction-based experimental control and optimization.
The more we can optimize and automate the development \textit{and} verification of emulators, the more efficiently scientists without ML expertise can iterate over simulation experiments, leading to more robust conclusions and more hypotheses to explore.

\paragraph{Deriving physical laws from data-driven surrogates}

Simulating complex dynamical systems often relies on governing equations conventionally obtained from rigorous first principles such as conservation laws or knowledge-based phenomenological derivations.  Although non-trivial to derive, these symbolic or mechanistic equations are interpretable and understandable for scientists and engineers. NN-based simulations, including surrogates, are not interpretable in this way and can thus be challenging to use, especially in many cases where it is important the scientist or engineer understand the causal, data-generating mechanisms.

\begin{figure}[!ht]
\centering
\includegraphics[width=0.7\linewidth]{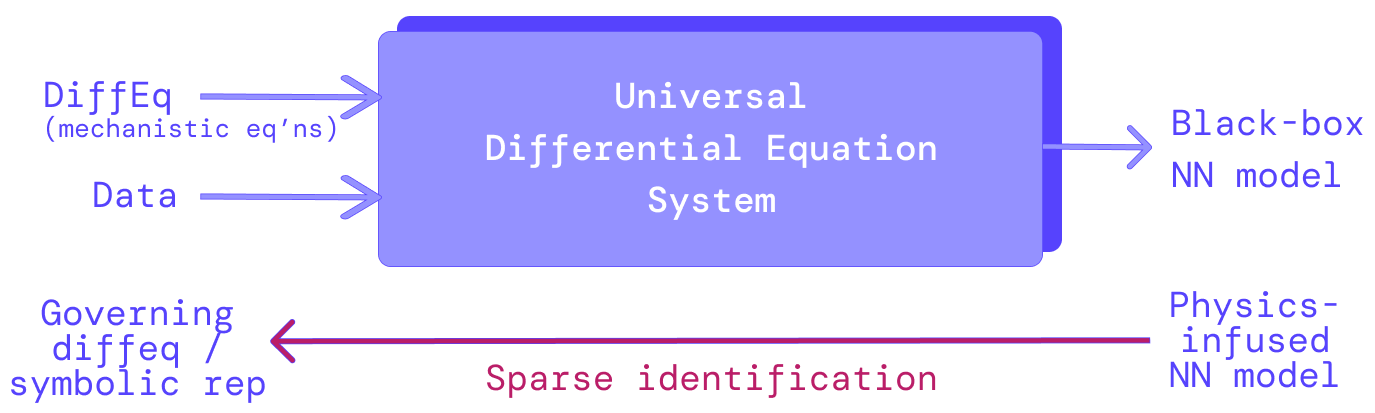}
\vspace{0.2cm}
\caption{Illustrating the UDE forward process (top), where mechanistic equations are used with real data to produce a trained neural network (NN) model, followed by the inverse problem of recovering the governing equations in symbolic form (bottom).
}
\label{fig:ude}
\end{figure}

Recent ML-driven advances have led approaches for \textit{sparse identification of nonlinear dynamics (SINDy)}~\cite{Brunton2016SparseIO} to learn ODEs or PDEs from observational data. SINDy essentially selects dominant candidate functions from a high-dimensional nonlinear function space based on sparse regression to uncover ODEs that match the given data; one can think of SINDy as providing an ODE surrogate model. This exciting development has led to scientific applications from biological systems \cite{Mangan2016InferringBN} to chemical processes \cite{Bhadriraju2019MachineLA} to active matter \cite{Cichos2020MachineLF}, as well as data-driven discovery of spatiotemporal systems governed by PDEs \cite{Rudy2017DatadrivenDO,Schaeffer2017LearningPD}.
Here we showcase two significant advances on SINDy utilizing methods described in the surrogate modeling motif and others:

\begin{enumerate}
    \item \textbf{Synergistic learning deep NN surrogate and governing PDEs from sparse and independent data} -- Chen et al. \cite{Chen2020DeepLO} present a novel physics-informed deep learning framework to discover governing PDEs of nonlinear spatiotemporal systems from scarce and noisy data accounting for different initial/boundary conditions (IBCs).
    Their approach integrates the strengths of deep NNs for learning rich features, automatic differentiation for accurate and efficient derivative calculation, and $l_{0}$ sparse regression to tackle the fundamental limitation of existing methods that scale poorly with data noise and scarcity. 
    The special network architecture design is able to account for multiple independent datasets sampled under different IBCs, shown with simple experiments that should still be validated on more complex datasets.
    An alternating direction optimization strategy simultaneously trains the NN on the spatiotemporal data \textit{and} determine the optimal sparse coefficients of selected candidate terms for reconstructing the PDE(s) -- the NN provides accurate modeling of the solution and its derivatives as a basis for constructing the governing equation(s), while the sparsely represented PDE(s) in turn informs and constraints the DNN which makes it generalizable and further enhances the discovery.
    The overall semi-mechanistic approach -- bottom-up (data-driven) and top-down (physics-informed) processes -- is promising for ML-driven for scientific discovery.

    \item \textbf{Sparse identification of missing model terms via Universal Differential Equations} -- We earlier described several scenarios where an ML surrogate is trained for only part of the full simulator system, perhaps for the computationally inefficient or the unknown parts. This is also a use-case of the UDE, replacing parts of a simulator described by mechanistic equations with a data-driven NN surrogate model. Now consider we're at the end of the process of building a UDE (we have learned and verified an approximation for part of the causal generative model (i.e. a simulator)). Do we lose interpretability and analysis capabilities? 
    With a knowledge-enhanced approach of the SINDy method we can sparse-identify the learned semi-mechanistic UDE back to mechanistic terms that are understandable and usable by domain scientists.
    Rackauckas et al. \cite{Rackauckas2020UniversalDE} modify the SINDy algorithm to apply to only subsets of the UDE equation in order to perform equation discovery specifically on the trained neural network components. In a sense this narrows the search space of potential governing equations by utilizing the prior mechanistic knowledge that wasn't replaced in training the UDE.
    Along with the UDE approach in general, this sparse identification method needs further development and validation with more complex datasets.
\end{enumerate}


\subsection*{Future directions}\label{sec:Surrogate_future}


The UDE approach has significant implications for use with simulators and physical modeling, where the underlying mechanistic models are commonly differential equations. By directly utilizing mechanistic modeling simultaneously with universal approximator models, UDE is a powerful semi-mechanistic approach allowing for arbitrary data-driven model extensions. In the context of simulators, this means a synergistic model of domain expertise and real-world data that more faithfully represents the true data-generating process of the system.

What we've described is a transformative approach for ML-augmented scientific modeling. That is, 
\begin{enumerate}
    \item Practitioner identifies known parts of a model and builds a UDE -- when using probabilistic programming (an SI engine motif), this step can be done in a high-level abstraction where the user does not need to write custom inference algorithms.
    \item Train an NN (or other surrogate model such as Gaussian process) to capture the missing mechanisms -- one may look to NAS and Bayesian optimization approaches to do this in an automated, uncertainty-aware way.
    \item The missing terms can be sparse-identified into mechanistic terms -- this is an active area of research and much verification of this concept is needed, as mentioned in the example above.
    \item Verify the recovered mechanisms are scientifically sane -- for future work, how can we better enable this with human-machine teaming?
    \item Verify quantitatively: extrapolate, do asymptotic analysis, run posterior predictive checks, predict bifurcations.
    \item Gather additional data to validate\footnote{Note we use ``verify'' and ``validate'' specifically, as there's important difference between \textit{verification and validation (V\&V)}: verification asks ``are we building the solution right?'' whereas validation asks ``are we building the right solution?''~\cite{lavin2021technology}.} the new terms.
\end{enumerate}

Providing the tools for this semi-mechanistic modeling workflow can be immense for enabling scientists to make the best use of domain knowledge \textit{and} data, and is precisely what the SI stack can deliver -- notably with the differentiable programming and probabilistic programming ``engine'' motifs. Even more, building in a unified framework that's purpose-built for SI provides extensibility, for instance to integrate recent graph neural network approaches from Cranmer et al. \cite{Cranmer2020DiscoveringSM} that can recover the governing equations in symbolic forms from learned physics-informed models, or the ``AI Feynmann''~\cite{Udrescu2020AIFA,Udrescu2020AIF2} approaches based on traditional fitting techniques in coordinate with neural networks that leverage physics properties such as symmetries and separability in the unknown dynamics function(s) -- key features such as NN-equivariance and normalizing flows (and how they may fit into the stack) are discussed later. 

\hfill \break
\subsection{3. SIMULATION-BASED INFERENCE}

\begin{figure}[!ht]
\centering
\includegraphics[width=0.95\linewidth]{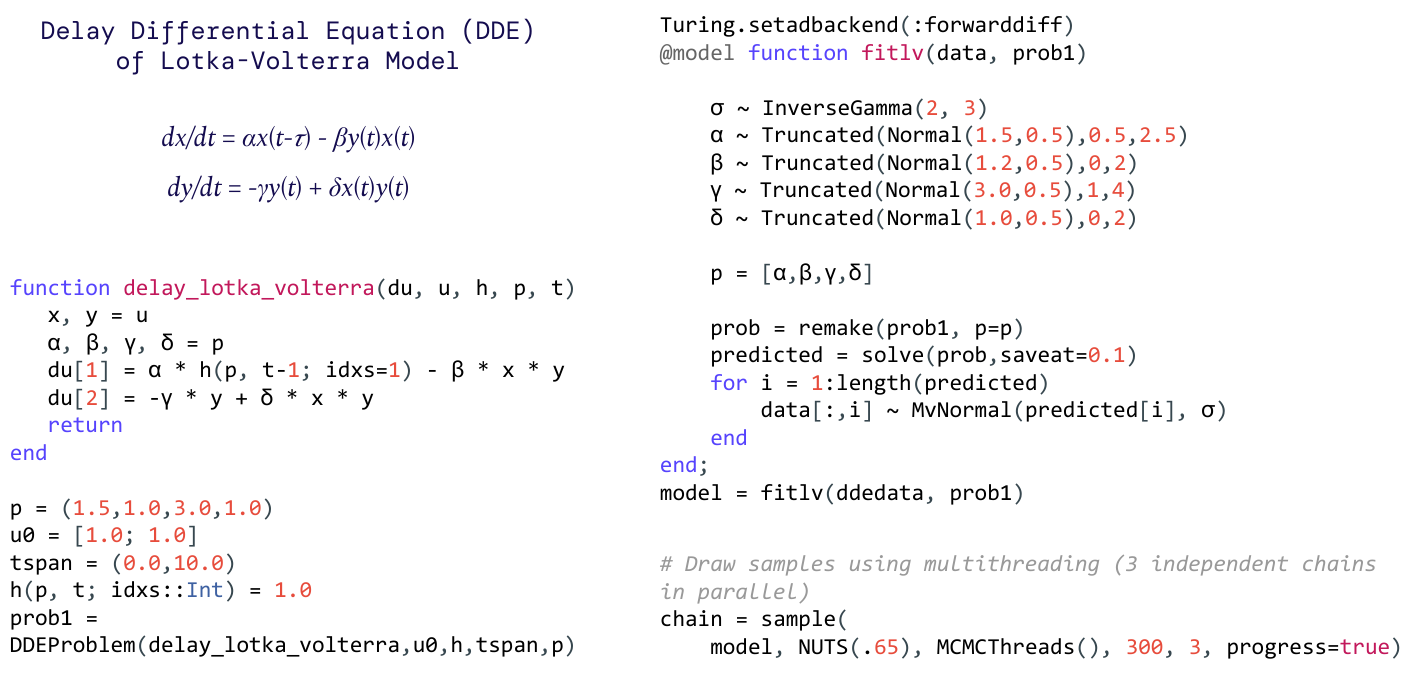}
\caption{Bayesian estimation of differential equations -- 
Here we show an example of a Turing.jl \cite{Ge2018TuringAL} probabilistic program and inference with a \textit{delay differential equation (DDE)}, a DE system where derivatives are functions of values at an earlier point in time, generally useful to model a delayed effect, such as incubation time of a virus.
Here we show the The Lotka–Volterra equations, a pair of first-order nonlinear differential equations that are frequently used to describe the dynamics of biological systems in which two species interact as predator and prey.
Differential equation models often have non-measurable parameters. The popular ``forward-problem'' of simulation consists of solving the differential equations for a given set of parameters, the ``inverse problem'' to simulation, known as parameter estimation, is the process of utilizing data to determine these model parameters. Bayesian inference provides a robust approach to parameter estimation with quantified uncertainty.
(Example code sourced from \href{https://turing.ml}{turing.ml})
}
\label{fig:ppl-code}
\end{figure}

Numerical simulators are used across many fields of science and engineering to build computational models of complex phenomena. These simulators are typically built by incorporating scientific knowledge about the mechanisms which are known (or assumed) to underlie the process under study. Such mechanistic models have often been extensively studied and validated in the respective scientific domains. In complexity, they can range from extremely simple models that have a conceptual or even pedagogical flavor (e.g. the Lotka-Volterra equations describing predator-prey interactions in ecological systems and also economic theories \cite{Gandolfo2008GiuseppePA} (expressed in Fig. \ref{fig:ppl-code})) to extremely detailed and expensive simulations implemented in supercomputers, e.g. whole-brain simulations \cite{markram2015reconstruction}.

A common challenge -- across scientific disciplines and complexity of models -- is the question of how to link such simulation-based models with empirical data. Numerical simulators typically have some parameters whose exact values are non \textit{a priori}, and have to be inferred from data. For reasons detailed below, classical statistical approaches can not readily be applied to models defined by numerical simulators. The field of \textit{simulation-based inference (SBI)} \cite{Cranmer2020TheFO} aims to address this challenge, by designing statistical inference procedures that can be applied to complex simulators. Building on foundational work from the statistics community (see \cite{sisson2018handbook} for an overview), SBI is starting to bring together work from multiple fields -- including, e.g.,  population genetics, neuroscience, particle physics, cosmology, and astrophysics -- that are facing the same challenges and using tools from machine learning to address them. SBI can provide a unifying language, with common tools \cite{tejero2020sbi,lintusaari2018elfi,klinger2018pyabc} and benchmarks \cite{lueckmann2021benchmarking} being developed and generalized across different fields and applications.

\begin{figure}
    \centering
    \includegraphics[width=\textwidth]{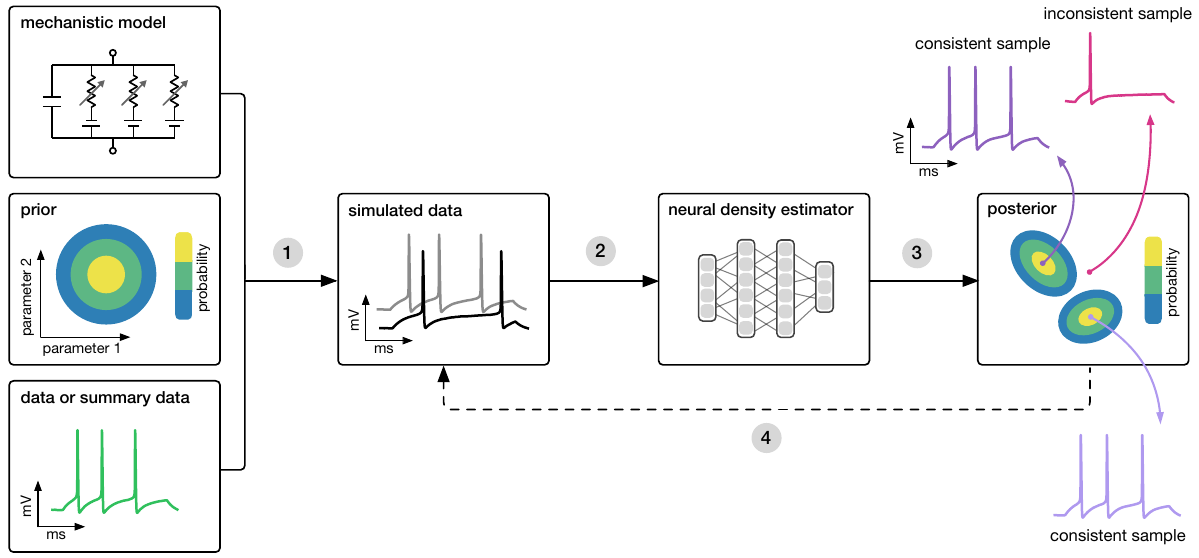}
    \caption{
    The goal of simulation-based inference (SBI) is to algorithmically identify parameters of simulation-based models which are compatible with observed data and prior assumptions. 
    SBI algorithms generally take three inputs (left): A candidate mechanistic model (e.g. a biophysical neuron model), prior knowledge or constraints on model parameters, and observational data (or summary statistics thereof). The general process shown is to (1) sample parameters from the prior followed by simulating synthetic data from these parameters; (2) learn the (probabilistic) association between data (or data features) and underlying parameters (i.e., to learn statistical inference from simulated data) for which different SBI methods (discussed in the text) such as neural density estimation \cite{gonccalves2020training} can be used; (3) apply the learned model to empirical data to derive the full space of parameters consistent with the data and the prior, i.e. the \textit{posterior distribution}. Posterior distributions may have complex shapes (such as multiple modes), and different parameter configurations may lead to data-consistent simulations.
    If needed, (4) an initial estimate of the posterior can be used to adaptively generate additional informative simulations.
   (Illustration from \cite{gonccalves2020training})
    }
    \label{fig:sbi-main}
\end{figure}


Why is it so challenging to constrain numerical simulations by data?  Many numerical simulators have stochastic components, which are included either to provide a verisimilar model of the system under study if it is believed to be stochastic itself, or often also pragmatically to reflect incomplete knowledge about some components of the system.  Linking such stochastic models with data falls within the domain of statistics, which aims to provide methods for constraining the parameters of a model by data, approaches for selecting between different model-candidates, and criteria for determining whether a hypothesis can be rejected on grounds of empirical evidence.  In particular, statistical inference aims to determine which parameters -- and combinations of parameters --  are compatible with empirical data and (possibly) \textit{a priori} assumptions. A key ingredient of most statistical procedures is the likelihood $p(x|\theta)$ of data $x$ given parameters $\theta$. For example, Bayesian inference characterizes parameters which are compatible both with data and prior by the posterior distribution $p(\theta|x)$, which is proportional to the product of likelihood and prior, $p(\theta|x) \propto p(x|\theta) p(\theta)$. Frequentist inference procedures typically construct confidence regions based on hypothesis tests, often using the likelihood ratio as test statistic.

However, for many simulation-based models, one can easily sample from the model (i.e., generate synthetic data $x \sim p(x|\theta)$) but \emph{evaluating} the associated likelihoods can be computationally prohibitive -- because, for instance, the same output $x$ could result from a very large number of internal paths through the simulator, and integrating over all of them is prohibitive. More pragmatically, it might also be the case that the simulator is implemented in a ``black-box'' manner which does not provide access to its internal workings or states. If likelihoods can not be evaluated, most conventional inference approaches can not be used. The goal of simulation-based inference is to make statistical inference possible for so-called \emph{implicit} models which allow generating simulated data, but not evaluation of likelihoods. 

SBI is not a new idea. Simulation-based inference approaches have been studied extensively in statistics, typically under the heading of \textit{likelihood-free inference}. An influential approach has been that of \textit{Approximate Bayesian Computation}~\cite{rubin1984, beaumont2002approximate, sisson2018handbook}. In its simplest form it consists of drawing parameter values from a proposal distribution, running the simulator for these parameters to generate synthetic outputs $x~p(x|\theta)$, comparing these outputs against the observed data, and accepting the parameter values only if they are close to the observed data under some distance metric, $\lVert x - x_{\mathrm{observed}}\rVert < \epsilon$. After following this procedure repeatedly, the accepted samples approximately follow the posterior. A second class of methods approximates the likelihood by sampling from the simulator and estimating the density in the sample space with kernel density estimation or histograms. This approximate density can then be used in lieu of the exact likelihood in frequentist or Bayesian inference techniques~\cite{Diggle1984MonteCM}. 

Both of these methods enable approximate inference in the likelihood-free setting, but they suffer from certain shortcomings: In the limit of a strict ABC acceptance criterion ($\epsilon \to 0$) or small kernel size, the inference results become exact, but the sample efficiency is reduced (the simulation has to be run many times). Relaxing the acceptance criterion or increasing the kernel size improves the sample efficiency, but reduces the quality of the inference results. The main challenge, however, is that these methods do not scale well to high-dimensional data, as the number of required simulations grows approximately exponentially with the dimension of the data $x$. In both approaches, the raw data is therefore usually first reduced to low-dimensional summary statistics. These are typically designed by domain experts with the goal of retaining as much information on the parameters $\theta$ as possible. In many cases, the summary statistics are not sufficient and this dimensionality reduction limits the quality of inference or model selection \cite{robert2011lack}. Recently, new methods for \emph{learning} summary statistics in a fully \cite{jiang2017learning,chen2021neural}  or semi-automatic manner \cite{fearnhead2012constructing} are emerging, which might alleviate some of these limitations.

The advent of deep learning has powered a number of new simulation-based inference techniques. Many of these methods rely on the key principle of training a neural surrogate for the simulation. Such models are closely related to the emulator models discussed in the previous section, but not geared towards efficient sampling. Instead, we need to be able to access its likelihood~\cite{2018arXiv180507226P, 2018arXiv180509294L} (or the related likelihood ratio~\cite{Neal:2007zz, 2012arXiv1212.1479F, Cranmer:2015bka, 2016arXiv161003483M, 2016arXiv161110242D,miller2021truncated}) or the posterior~\cite{le2017using,NIPS2016_6084, 2017arXiv171101861L,greenberg2019automatic}. After the surrogate has been trained, it can be used during frequentist or Bayesian inference instead of the simulation. On a high level, this approach is similar to the traditional method based on histograms or kernel density estimators~\cite{Diggle1984MonteCM}, but modern ML models and algorithms allow it to scale to higher-dimensional and potentially structured data. 

The impressive recent progress in SBI methods does not stem from deep learning alone. Another important theme is active learning: running the simulator and inference procedure iteratively and using past results to improve the proposal distribution of parameter values for the next runs~\cite{ranjan2008sequential, Bect2012, 2014arXiv1401.2838M, 2015arXiv150103291G, 2015arXiv150603693M, 2018arXiv180507226P, 2018arXiv180509294L, NIPS2016_6084, 2017arXiv171101861L,jarvenpaa2019efficient}. This can substantially improve the sample efficiency. Finally, in some cases simulators are not just black boxes, but we have access to (part of) their latent variables and mechanisms or probabilistic characteristics of their stack trace. In practice, such information can be made available through domain-specific knowledge or by implementing the simulation in a framework that supports differential or probabilistic programming-- i.e., the SI engine. If it is accessible, such data can substantially improve the sample efficiency with which neural surrogate models can be trained, reducing the required compute~\cite{graham2017asymptotically, Brehmer:2018hga, Stoye:2018ovl,baydin2019efficient,kersting2020differentiable}. On a high level, this represents a tighter integration of the inference engine with the simulation \cite{Baydin2019EtalumisBP}.

\begin{figure}[!t]
\centering
\includegraphics[width=1.0\linewidth]{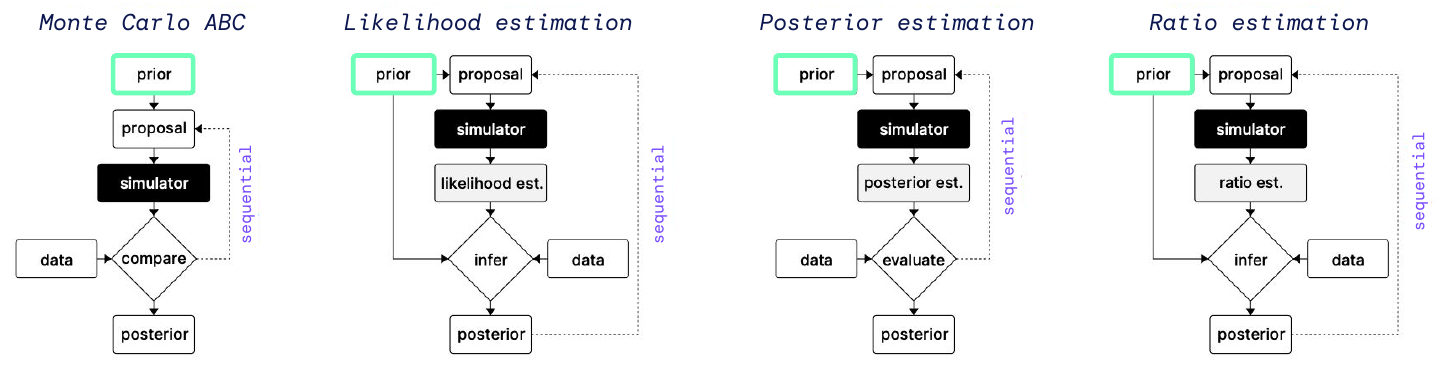}
\caption{Various simulation-based inference workflows (or prototypes) are presented in Cranmer et al.~\cite{Cranmer2020TheFO}. Here we show four main workflows (or templates) of simulation-based inference: the left represents Approximate Bayesian Computation (ABC) approaches, and then to the right are three model-based approaches for approximating likelihoods, posteriors, and density ratios, respectively. Notice that all include algorithms that use the prior distribution to propose parameters (green), as well as algorithms for sequentially adapting the proposal (purple)---i.e., steps (1) and (4) shown in Fig. \ref{fig:sbi-main}.
(Figure reproduced from Ref.~\cite{lueckmann2021benchmarking})
}
\label{fig:sbi}
\end{figure}

These components -- neural surrogates for the simulator, active learning, the integration of simulation and inference -- can be combined in different ways to define workflows for simulation-based inference, both in the Bayesian and frequentist setting. We show some example inference workflows in Fig.~\ref{fig:sbi}. The optimal choice of the workflow depends on the characteristics of the problem, in particular on the dimensionality and structure of the observed data and the parameters, whether a single data point or multiple i.i.d.\ draws are observed, the computational complexity of the simulator, and whether the simulator admits accessing its latent process.

\subsection*{Examples}\label{sec:SBI_ex}


Simulation-based inference techniques have been used to link models to data in a wide range of problems, reflecting the ubiquity of simulators across the sciences. In physics, SBI is useful from the smallest scales in particle colliders~\cite{Brehmer:2018kdj, Brehmer:2018eca, Brehmer:2019xox}, where it allows us to measure the properties of the Higgs boson with a higher precision and less data, to the largest scales in the modeling of gravitational waves~\cite{Delaunoy:2020zcu, Dax:2021tsq}, stellar streams~\cite{Hermans:2020skz}, gravitational lensing~\cite{Brehmer:2019jyt}, and the evolution of the universe~\cite{Alsing:2018eau}. These methods have also been applied to study the evolutionary dynamics of protein networks~\cite{ratmann2007using} and in yeast strains~\cite{avecilla2021simulation}. In neuroscience, they have, e.g.,  been used to estimate  the properties of ion channels from high-throughput voltage-clamp, and properties of neural circuits from observed rhythmic behaviour in the stomatogastric ganglion~\cite{gonccalves2020training}, to identify network models which can capture the dynamics of neuronal cultures \cite{sukenik2021neuronal}, to study how the connectivity between different cell-types shapes dynamics in cortical circuits \cite{bittner2021interrogating}, and to identify biophysically realistic models of neurons in the retina \cite{oesterle2020bayesian,yoshimatsu2021ancestral}.

 SBI is not restricted to the natural sciences. In robotics, for example, simulators are commonly used to generate training data on which the parameters of control- and policy-search algorithms are subsequently trained. A common challenge is the question of how policies trained on simulated data can subsequently be transferred to the real world, called ``sim-to-real'' transfer. SBI has been used to infer posterior-distributions over the parameters of simulators \cite{ramos2019bayessim,muratore2021neural} -- the idea is that simulations from the posterior would yield more realistic data than `hand-tuned' simulators, and that by simulating from the posterior, one would also cover a range of different domains (or distributions, environment variations, etc.) rather than just over-fit on a single parameter regime. For example, \cite{marlier2021simulation} used this approach to identify posteriors over hand-configuration, on which a  multi-fingered robotic grasping algorithm was trained. 



Here we will not be able to do the wide variety of example applications of SBI justice, and instead briefly zoom in on two particular use cases from particle physics and neuroscience.

\paragraph{Measuring the properties of the Higgs boson}

The experiments at the Large Hadron Collider probe the laws of physics at the smallest length scales accessible to humans. A key goal is to precisely measure how the Higgs boson interacts with other elementary particles: patterns in these interactions may
provide evidence for physics beyond the ``Standard Model'', the established set of elementary particles and laws that govern them, and
ultimately help us answer some of the open fundamental questions of elementary physics.

\begin{figure}[!ht]
    \centering
    \includegraphics[width=\textwidth]{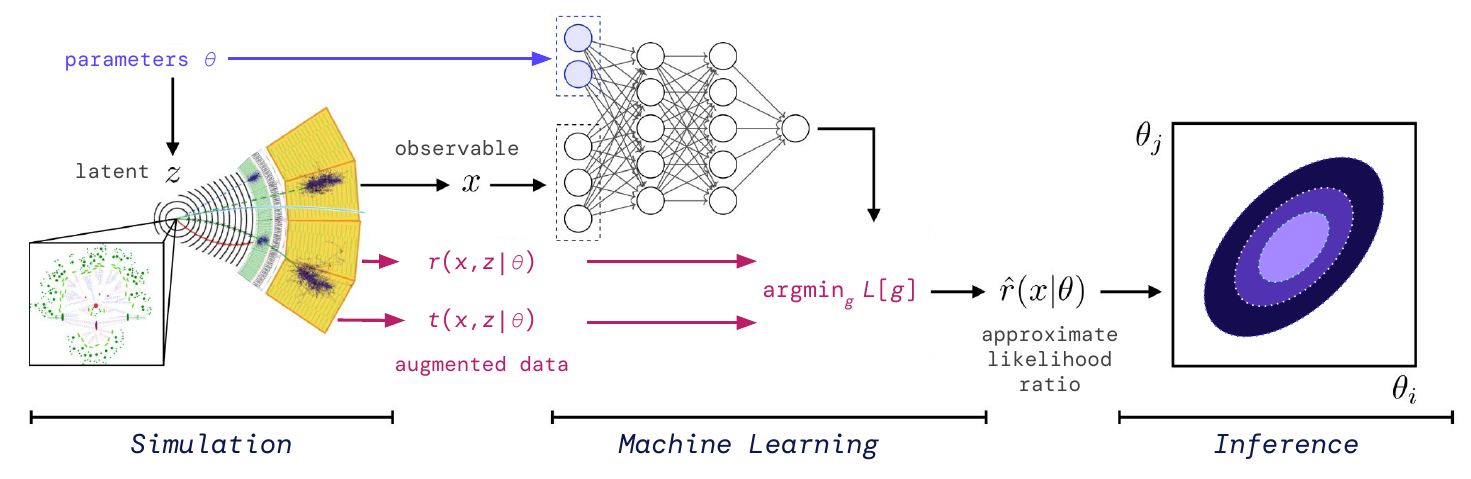}
    \caption{Schematic of our simulation-based inference workflow used in particle physics. (Reproduced from Ref.~\cite{Brehmer:2018kdj})}
    \label{fig:sbi_lhc_sketch}
\end{figure}

In this endeavour physicists rely on a chain of high-fidelity simulators for the interactions between elementary particles, their radiation and decay patterns, and the interactions with the detector. Owing to the large number of stochastic latent variables (state-of-the-art simulations involve multiple millions of random variables), the likelihood function of these simulators is intractable. 
Inference is traditionally made possible by first reducing the high-dimensional data to handcrafted summary statistics and estimating the likelihood with histograms, as discussed above, but popular summary statistics are often lossy, leading to less precise measurements.



Recently, particle physicists have begun developing and applying simulation-based inference methods driven by machine learning to this problem~\cite{Brehmer:2020cvb, Brehmer:2018kdj, Brehmer:2019xox}. These methods on the one hand draw from the growing interdisciplinary toolbox of simulation-based inference and on the rapid progress in deep learning. On the other hand, they are tailored to the characteristics of this scientific measurement problem: they make use of the latent structure in particle physics simulators, are geared towards a large number of i.\,i.\,d.\ samples, and support inference in a frequentist framework, the established paradigm in particle physics. Multiple different algorithms have been proposed, but the basic strategy commonly consists of three steps (as we sketch in Fig.~\ref{fig:sbi_lhc_sketch}):
\begin{enumerate}
    \item The chain of simulators is run for various values of the input parameters $\theta$, for instance characterizing the interaction patterns between the Higgs boson and other elementary particles. These parameters are sampled from a proposal distribution, which needs to cover the region of interest in the parameter space (but does not have to be related to a Bayesian prior). For each simulated sample, the simulator inputs $\theta$ and outputs $x \sim p(x|\theta)$ are stored. The outputs $x$ can consist of low-level data like energy deposits in the various detector components, but often it is more practical to parameterize them in form of preprocessed high-level variables like the kinematic properties of reconstructed particles. In addition, certain additional latent variables can be stored that characterize the simulated process.
    \item A machine learning model, often a multilayer perceptron \cite{Tang2016ExtremeLM}, is trained on the simulated dataset, minimizing one of several proposed loss functions. They all have in common that (assuming sufficient capacity, infinite data, and perfect optimization) the network will converge to the likelihood ratio function $r(x|\theta) = p(x|\theta) / p_{\mathrm{ref}}(x)$, where $p_{\mathrm{ref}}(x)$ is a reference distribution (for instance the marginal likelihood or the likelihood for some value of the parameters). Some loss functions leverage the latent variables from the simulator process to improve the sample efficiency of the training.
    \item The observed dataset consists of multiple i.\,i.\,d.~events $x_{\mathrm{obs}}$. The trained network is evaluated for each of these events. This provides a tractable and differentiable approximate likelihood function (up to an irrelevant overall constant), which is then used to define the maximum likelihood estimator $\hat{\theta}$ as well as confidence regions based on likelihood ratio tests, often relying on convenient asymptotic properties---in this frequentist approach, nuisance parameters through profiling~\cite{Cowan:2010js}.  
\end{enumerate}

First phenomenological studies on synthetic data have shown that these methods have the potential to substantially improve the precision of Higgs measurements at the LHC~\cite{Brehmer:2018kdj, Brehmer:2018eca, Brehmer:2019xox, Brehmer:2019gmn, Chen:2020mev, Barman:2021yfh, Bahl:2021dnc}. It remains a challenge to scale these analyses to real data and in particular to a realistic modeling of systematic uncertainties, which are usually modeled with thousands of nuisance parameters.




\begin{figure}[!h]
\centering
\includegraphics[width=1.0\linewidth]{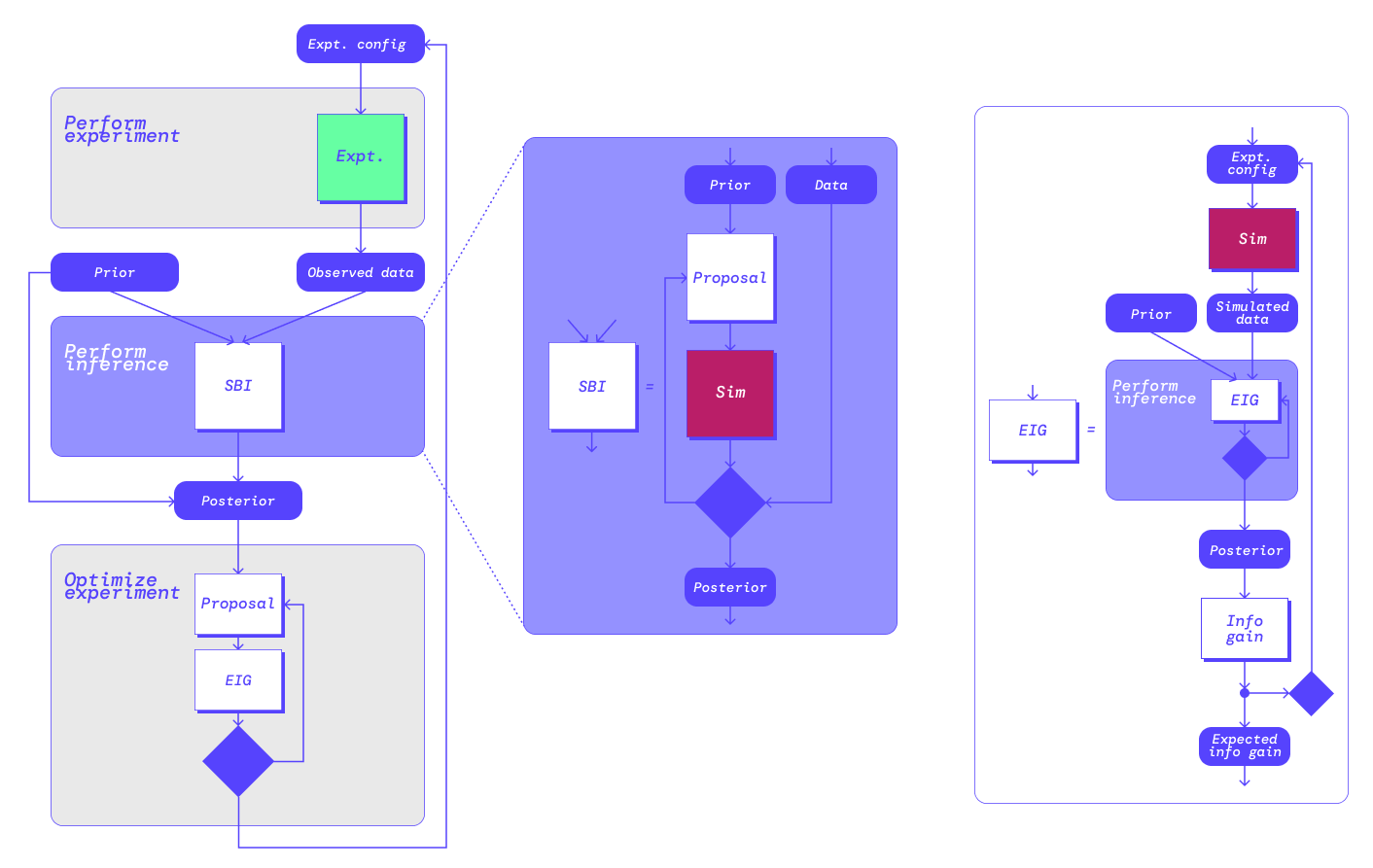}
\caption{
Workflows for ``active science'' with simulation-based inference methods \cite{Cranmer2017Github}, which we call \textit{human-machine inference}. The scientist's inputs to the system are an external workflow that implements some experimental protocol (green Expt), and an external workflow that implements a simulator for those experiments, which depends on some theoretical parameters that we would like to infer (red Sim component). The center diagram double-clicks on the simulation-based inference (SBI) part, representing generic likelihood-free inference engines that enable statistical inference on the parameters of a theory that are implicitly defined by a simulator. The right diagram shows how the experiment is optimized with active learning and sequential design algorithms (namely Bayesian optimization, to balance exploration and exploitation to efficiently optimize an expensive black box objective function). Here the objective to optimize is expected information gain (EIG) over the hypothesis parameters.
}
\label{fig:activesci}
\end{figure}

\subsection*{Future directions}\label{sec:SBI_future}

An intriguing direction to further develop SBI methods is towards \textit{human-machine inference}: an efficient and automated loop of the scientific method (at least, for sufficiently well posed problems). Described in Fig. \ref{fig:activesci}, the overall process is (1) perform experiment, (2) perform inference, (3) optimize experiment, and (4) repeat from (1). This approach provides an efficient means of physical experiments, particularly for expensive experiments in large search and hypothesis spaces, for instance the Large Hadron Collider at CERN \cite{Louppe2017TeachingMT}. 
This is one means by which simulation intelligence can broaden the scientific method. It will be fascinating to advance this SI-based science workflow by replacing the information gain objectives with open-ended optimization (discussed later in the SI engine motifs).

Simulation-based inference is a rapidly advancing class of methods, with progress fueled by newer programming paradigms such as probabilistic programming and differentiable programming. Cranmmer et al.~\cite{Cranmer2020TheFO} predict that ``several domains of science should expect either a significant improvement in inference quality or the transition from heuristic approaches to those grounded in statistical terms tied to the underlying mechanistic model. It is not unreasonable to expect that this transition may have a profound impact on science.''

\hfill \break
\subsection{4. CAUSAL REASONING}

Standard approaches in statistical analysis and pattern recognition draw conclusions based on associations among variables and learned correlations in the data. Yet these approaches do not suffice when scientific queries are of \textit{causal} nature rather than associative -- there is marked distinction between correlations and \textit{cause-effect} relationships. Such causal questions require some knowledge of the underlying data-generating process, and cannot be computed from the data alone, nor from the distributions that govern the data \cite{Pearl2000CausalityMR, Pearl2009CausalII}.
Understanding causality is both the basis and the ultimate goal of scientific research in many domains -- from genomics \cite{Gruber2010AnAO} and healthcare \cite{Prosperi2020CausalIA}, to economics \cite{Athey2018TheIO} and game theory \cite{Toulis2016LongtermCE}, to Earth systems sciences \cite{Runge2019InferringCF} and more. 

Abstractly we can consider a causal model to be qualitatively between a mechanistic model (described by differential equations, for example) and a statistical model (learned from data). That is, a mechanistic model is a description of the dynamics governing a system but is manually encoded by human experts, while a statistical model such as a neural network learns a mapping of variables from observational data but that only holds when the experimental conditions are fixed. Neither allows for causal queries. Causal learning lies between these two extremes, seeking to model the effects of interventions and distribution changes with a combination of data-driven learning and assumptions not already included in the statistical description of a system \cite{Scholkopf2021TowardCR}. 

\begin{figure}[!ht]
\centering
\includegraphics[width=.95\linewidth]{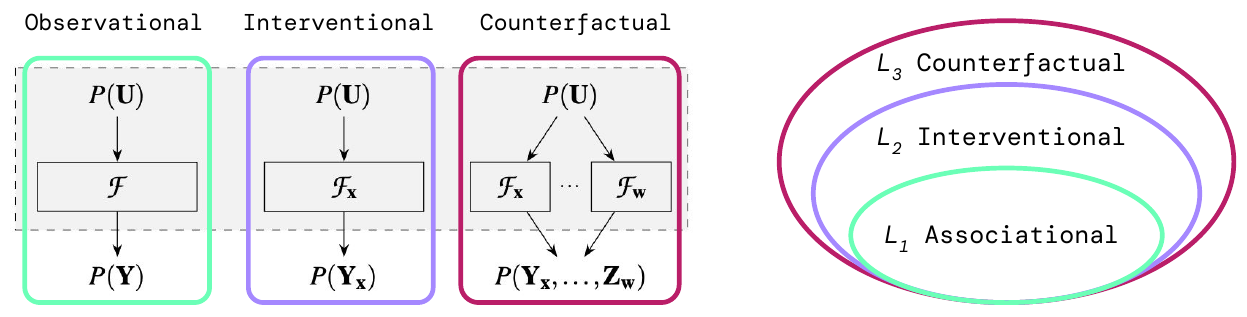}
\caption{
The three levels of the causal hierarchy: (1) associational, (2) interventional, and (3) counterfactual.
Left: $P(U)$ describes the natural state of a system in domain $U$, and $F$ represents the transformation to an observational world (green), an interventional world with modifying mechanism $F_{X}$ (purple), and multiple counterfactual worlds with multiple modifying mechanisms $\{X,...,W\}$ (red).
Right: The three levels of the causal hierarchy separate in a measure-theoretic sense: data at one level almost always underdetermines data at higher levels.
(Figure reproduced from Ref.~\cite{Bareinboim20211OP})
}
\label{fig:causallevels}
\end{figure}

Reasoning about cause and effect with a causal model corresponds to formulating causal queries, which have been, organized by Pearl \& Mackenzie~\cite{Pearl2018TheBO} in a three-level hierarchy termed the \textit{ladder of causation} \cite{Peyrard2020ALO} (Fig. \ref{fig:causallevels}):
\begin{enumerate}
    \item \textit{Observational} queries: seeing and observing. What can we tell about $Y$ if we observe $X = x$?
    \item \textit{Interventional} queries: acting and intervening. What can we tell about $Y$ if we do $X := x$?
    \item \textit{Counterfactual} queries: imagining, reasoning, and understanding. Given that $E = e$ actually happened, what would have happened to $Y$ had we done $X := x$?
\end{enumerate}

where $X$ and $Y$ are variables in the system of interest with events $E$.
The difference between observation (conditional probability) and action (interventional calculus) is what motivated the development of causality \cite{Hardt2021PatternsPA}.
More precisely the disagreement between interventional statements and conditional statements is defined as \textit{confounding}: $X$ and $Y$ are confounded when the causal effect of action $X := x$ on $Y$ does not coincide with the corresponding conditional probability. 

It is common to approach the analysis of causal relationships with \textit{counterfactual reasoning}: the difference between the outcomes if an action had been taken and if it had not been taken is defined as the causal effect of the action \cite{Pearl2009CausalII, Morgan2014CounterfactualsAT}. Counterfactual problems involve reasoning about why things happened, imagining the consequences of different actions for which you never observe, and determining which actions would have achieved a desired outcome. A counterfactual question may be ``would the virus have spread outside the city if the majority of adults were vaccinated one week sooner?'' 
In practice this can be done with \textit{structural causal models (SCMs)}, which give us a way to formalize the effect of hypothetical actions or interventions on the population within the assumptions of our model. An SCM, represented as a directed acyclic graph (DAG), is a collection of formal assumptions about how certain variables interact and compose the data generating process.


Answers to counterfactual questions strongly depend on the specifics of the SCM; it is possible to construct two models that have identical graph structures, and behave identically under interventions, yet give different answers or potential outcomes \cite{Peters2017ElementsOC}.
Without an SCM, we can still obtain causal estimates based on counterfactuals by using the \textit{potential outcomes framework}, or the Rubin-Neymann causal model \cite{Rubin1974EstimatingCE,Rubin2005CausalIU}, which computes cause-effect estimates by considering the set of potential events or actions leading to an outcome $Y$.

A step below on the causal ladder are \textit{interventional} questions, for instance ``how does the probability of herd immunity change if the majority of adults are vaccinated?''
Interventions in an SCM are made by modifying a subset of variable assignments in the model. 
Several types of interventions exist \cite{Eaton2007ExactBS}: hard/perfect involves directly fixing a variables value, soft/imperfect corresponds to changing the conditional distribution, uncertain intervention implies the learner is unsure which mechanism/variable is affected by the intervention, and of course the no intervention base case, where only observational data is obtained from the SCM. 

From the machine learning perspective, counterfactual reasoning is strictly harder than interventional as it requires imagining action rather than doing the action. And associative reasoning -- that is, with a statistical model such as a neural network -- is the simplest of all as it purely observes data (i.e., level 1 of the causal ladder). We illustrate these three levels of causality in Fig. \ref{fig:causallevels}. 
For full treatment please refer to Pearl's many works \cite{Pearl2000CausalityMR, Pearl2009CausalII}, the causal inference survey by Yao et al.~\cite{Yao2020ASO}, Sch{\"o}lkopf et al. \cite{Scholkopf2019CausalityFM, Scholkopf2021TowardCR} and the references therein for more details on causality in machine learning, and Guo et al.~\cite{Guo2020ASO} for an overview of causal data science broadly. Here we focus on causality in scientific simulators.


Simulators provide virtual testbeds for the exploration of complex real-world scenarios, and, if built with the tooling for causal machine learning (specifically counterfactual reasoning), can provide ``what-if'' analyses in an efficient and cost-effective way. 
That is, each run of a simulator is a realization of a particular possible world, and counterfactual what-if claims can be studied by varying independent parameters of the model and re-running the simulation.

\paragraph{\textit{Simulation trace-based causality}}
Given simulator execution traces in general consist of states, events, and actions, we can define causal relationships based on these entities -- without this grounding, one cannot make reliable causal claims, which we discuss in the final example of this section.
Herd \& Miles \cite{Herd2019DetectingCR} provide a practical formulation: event $C$ causes event $E$ in an actual trace $\pi$ if and only if there exists a counterfactual trace $\pi'$ that is similar enough to $\pi$ which requires that neither $C$ nor $E$ happens in the same state in which they happened in $\pi$. This suggests counterfactual analysis can be performed by searching all traces produced by the simulator for counterfactual $\pi'$, but this is clearly intractable for any non-trivial simulator. However, if we \textit{intervene} on the trace during the simulation process, a variety of counterfactual traces can be generated with slight differences in events such that we may reason about the causal relationships between events. It follows that this approach is called \textit{intervention-based counterfactual analysis}.

Simulation traces can be complex, amounting to several forms of causation based on the occurrence and omission of events in trace $\pi$ \cite{Herd2019DetectingCR}:
\begin{enumerate}
    \item \textit{Basic causation --} In order for event $A$ to cause event $B$ ($A \rightarrow B$), $A$’s occurrence has to be both sufficient and necessary for $B$ in $\pi$. 
    That is, for the mathematically inclined reader,
    $A \rightarrow B \equiv (A \Rightarrow B) \wedge (B \Rightarrow A) \equiv A \Leftrightarrow B$.
    \item \textit{Prevention --} In order for event $A$ to prevent event $B$, $A$’s occurrence has to be both necessary and sufficient for $B$’s omission. That is, $A \rightarrow B \equiv (A \Rightarrow \neg B) \wedge (\neg B \Rightarrow A) \equiv A \Leftrightarrow \neg B$.
    \item \textit{Transitivity --} If $A$ causes $B$ and $B$ causes $C$ then $A$ causes $C$.
    \item \textit{Causation by omission--} If $A$’s omission is both necessary and sufficient for $B$’s occurrence and we accept that omissions are allowed to be causes, then $A$’s omission can be considered to cause $B$’s occurrence.
    \item \textit{Prevention by caused prevention--} It follows from the prior causal claims that if $A$ causes $B$ and $B$ prevents $C$, then $A$ prevents $C$.
\end{enumerate}

The execution of a simulator results in one or more simulation traces, which represent various paths of traversing the state space given the underlying data-generating model, which is not unlike the execution traces resulting from probabilistic programming inference. By formalizing the sequence of simulator states as trace-based definition of causal dependencies, we have the computational and mathematical tooling to make causal inquiries, which we introduce in examples below. 
Further, with a simulator that produces a full posterior over the space of traces, as in probabilistic programming, we can attempt to draw causal conclusions with causal inference and discovery algorithms. This we introduce later in the motifs integrations section.

\begin{figure}[!ht]
\centering
\includegraphics[width=0.7\linewidth]{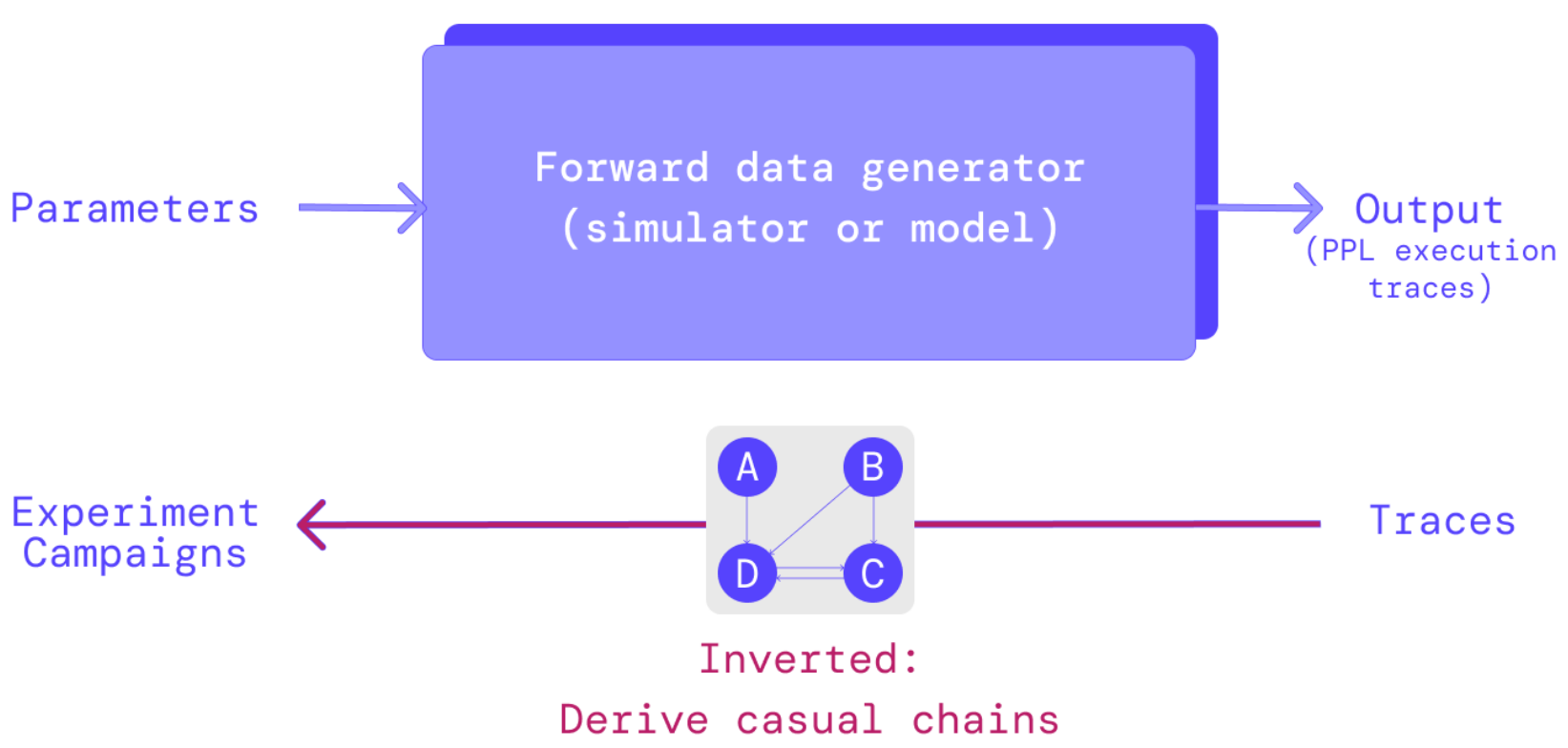}
\caption{
Inverting traditional forward data generation models -- recall that a simulator aims to encode the cause-to-effect data generation process -- to learn what causes could have produced the observed effects, and then to efficiently generate experimental campaigns to test these hypotheses. In practice, simulators that enable this cause-effect inversion learning can represent a new workflow that would broaden the scientific method \cite{Stevens2020AIFS}.
}
\label{fig:causal-invert}
\end{figure}

\subsection*{Examples}\label{sec:Causality_ex}

\begin{figure}[ht]
\centering
\includegraphics[width=0.85\linewidth]{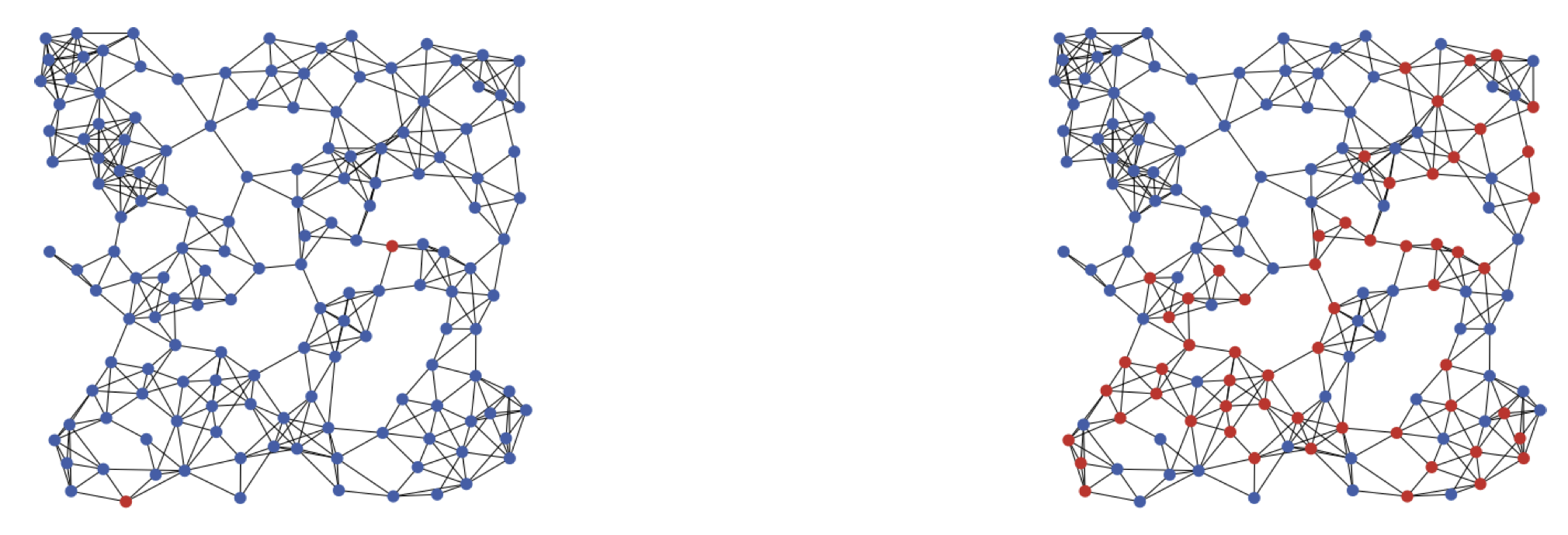}
\caption{Simulation runs from a virus-propagation agent-based model (ABM), where red nodes are infected agents. With root-cause analysis based on the defined simulation-based claims of causality, the agents were partitioned into two distinct groups of independent causal chains (of infection). (Original figure from Ref.~\cite{Herd2019DetectingCR})
}
\label{fig:virus}
\end{figure}

\paragraph{Tracing causal epidemiology chains with \textit{in silico} interventions}
Systems of networked entities or agents that evolve over time can be difficult to understand in a practical way towards decision-making. 
For instance, the propagation of a virus in an environment, where it can be useful to identify ``patient zero'' or the agent that initiated the chain of infection.
Methods for intervention-based counterfactual analysis can be used in agent-based modeling simulation to determine the root cause for each eventually infected agent.

First the researchers \cite{Herd2019DetectingCR} begin with a simple counterfactual simulation experiment without intervention -- that is, only manipulating the initial infection states before starting the simulation, not actively intervening in the simulation execution. This exemplifies the standard use of simulators for counterfactual reasoning: modify the input variables and environment parameters, observe the change in outputs per run, and assume cause-effect relationships. This inactive approach is susceptible to \textit{preemption confounding}, where only manipulating the initial state of the disease transmission model may open the door for spurious background effects to creep in, and, from an observer's perspective, mask the original causal relationship. Thus, except for very simplistic simulations, inactive counterfactual reasoning is not necessarily reliable for determining root causes. To actively intervene in the simulation, one needs to implement a series of experiments with both negated infections and altered times of infection. In a small testbed of 10 agents, the root cause analysis produces a correct partitioning of the agents into two groups, where each group shared a common root cause infection, shown in Fig. \ref{fig:virus}.
This demonstration is simplistic in the representation and dynamics of agent states. A more interesting use-case for future work would be simulations with multi-agent policies, such as the economic simulations we discuss next in the agent-based modeling motif, where one could actively intervene to discover why agents make certain decisions.
Additional challenges to this approach can arise when considering multiple competing and interacting causal chains, which is to be expected in simulations of complex scenarios. 


\paragraph{Active causal discovery and optimizing experiment design}

Asking the causal questions discussed above requires a causal model to begin with. The problem of inferring such a model from observational, interventional, or mixed data is called \textit{causal discovery}. 
Dedicated causal discovery algorithms exist and can be separated into two subtypes: constraint-based and score-based. The constraint-based algorithms construct the causal structure based on conditional independence constraints, while the score-based algorithms generate a number of candidate causal graphs, assign a score to each, and select a final graph based on the scores \cite{Shen2020ChallengesAO}. For an overview of the causal discovery subfield see Glymour et al. \cite{Glymour2019ReviewOC}.
Here we're interested in \textit{active} causal discovery, illustrating a method that can be a novel ``human-machine teaming'' workflow.

\begin{figure}[ht]
\centering
\includegraphics[width=0.7\linewidth]{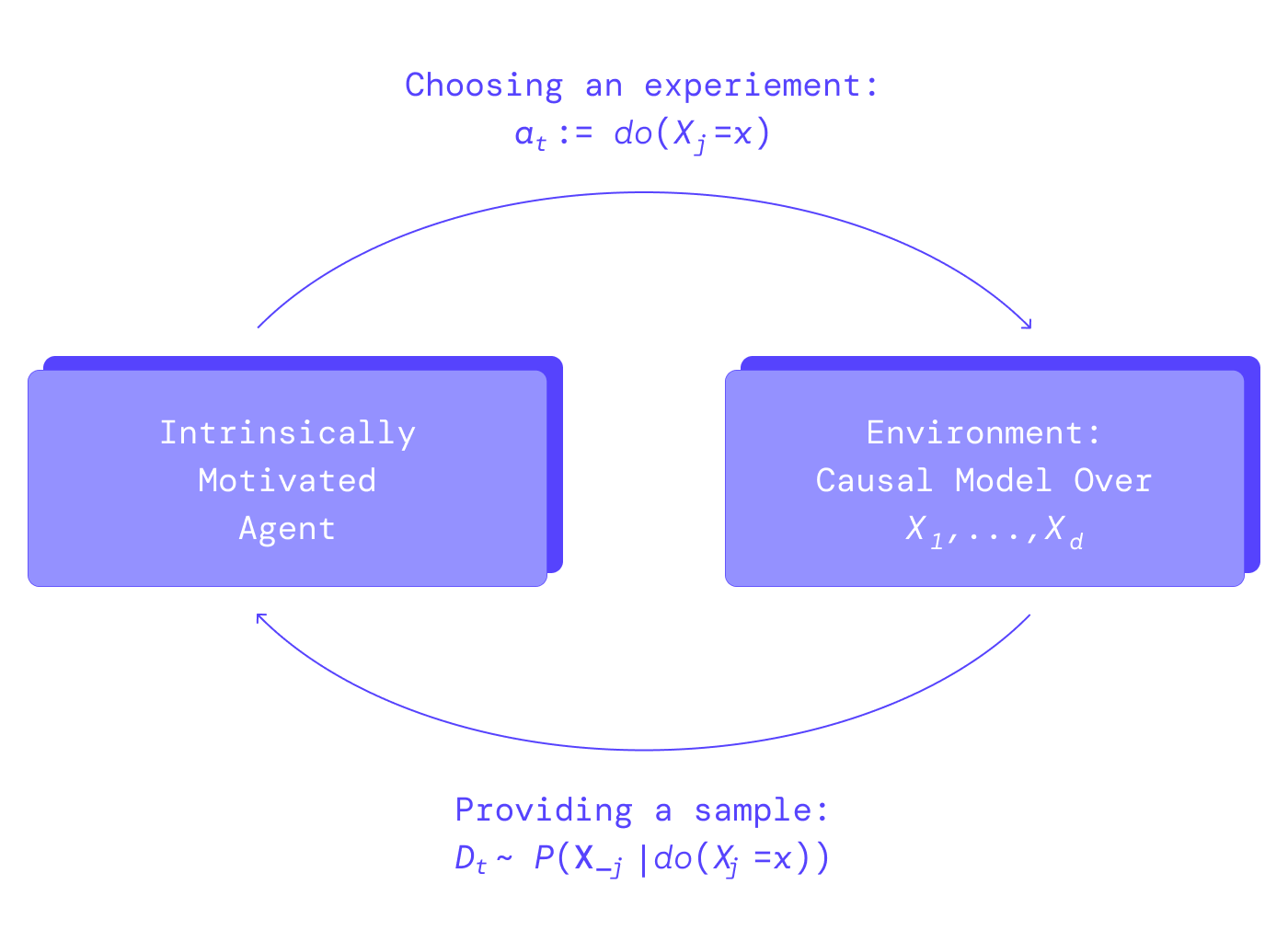}
\caption{``Active'' Bayesian causal discovery, where an \textit{in silico} agent iteratively experiments with its simulation environment in order to learn a structured causal model. This simplified diagram shows a causal model learned over observed variables, where the agent actively intervenes on some of the variables to derive cause-effect relationships.
}
\label{fig:causalBO}
\end{figure}

The setting for active causal discovery involves an agent iteratively selecting experiments (or targeted interventions) that are maximally informative about the underlying causal structure of the system under study. The experiment process is shown in Fig.~\ref{fig:causalBO}. At each iteration the agent,
\begin{enumerate}
    \item Performs an experiment in the form of intervening on one of the variables and fixing its value, $do(X_j = x)$; 
    \item Observes the outcome by sampling from the interventional distribution $P(X_{-j}|do(X_j = x))$; 
    \item Updates its beliefs with the observed outcome; 
    \item Plans the next experiment to maximize information gain (or similar criteria).
\end{enumerate}
The process repeats until the agent converges on a causal world model (with sufficient confidence) or an experiment threshold is reached. We use the syntax $do()$, which refers to Pearl's ``\textit{do}-calculus'' \cite{Pearl1995CausalDF, Tucci2013IntroductionTJ}. In short, the \textit{do}-operator forces a variable to take a certain value or distribution, which is distinct from conditioning that variable on a value or distribution.
This distinct operation allows us to estimate interventional distributions and causal effects from observational distributions (under certain conditions). 
Notice in the SI-stack Fig. \ref{fig:os} that \textit{do}-calculus is stated explicitly with the causality module.

The active learning approach with targeted experiments allows us to uniquely recover the causal graph, whereas score- or constraint-based causal discovery from observational data can struggle going beyond the Markov equivalence class \cite{Kgelgen2019OptimalED}.
In the simulation setting we have virtual agents, while real scientists can follow the same procedures in the lab, potentially to minimize the experiments needed on costly equipment -- notice the relationship to the sequential design-of-experiment methods we previously discussed, but now with the mathematics of causality. One can also imagine several variations involving human-machine teams to iterate over causal models and interventions (both \textit{in silico} and \textit{in situ}).

Von K{\"u}gelgen et al.~\cite{Kgelgen2019OptimalED} introduce a probabilistic approach with Bayesian nonparametrics, and there's been one application we know of, in the domain of neurodgenerative diseases \cite{Lavin2020NeurosymbolicND}.
The main developments are twofold: 

\setlength{\leftskip}{1cm}

1. The causal models contain continuous random variables with non-linear functional relationships, which are modeled with \textit{Gaussian process (GP)} priors.
A GP is a stochastic process which is fully specified by its mean function and covariance function such that any finite set of random variables have a joint Gaussian distribution \cite{GPML}. GPs provide a robust method for modeling non-linear functions in a Bayesian nonparametric framework; ordinarily one considers a GP prior over the function and combines it with a suitable likelihood to derive a posterior estimate for the function given data. GPs are flexible in that, unlike parametric counterparts, they adapt to the complexity of the data. GPs are largely advantageous in practice because they provide a principled way to reason about uncertainties \cite{Ghahramani2015ProbabilisticML}.

2. The space of possible interventions is intractable, so \textit{Bayesian optimization (BO)} is used to efficiently select the sequence of experiments that balances exploration and exploitation relative to gained information and optimizing the objective. 
BO is a probabilistic ML approach for derivative-free global optimization of a function $f$, where $f$ is typically costly to evaluate (in compute, time, or interactions with people and environment).
BO consists of two main components: a Bayesian statistical model for modeling the objective function (typically a GP, thus with uncertainty quantification), and an acquisition function for deciding where to sample next~\cite{Shahriari2016TakingTH, Frazier2018ATO}. 
The general idea is to trade off computation due to evaluating $f$ many times with computation invested in selecting more promising candidate solutions $x$, where the space $X$ can represent parameters of a real-world system, a machine learning model, an \textit{in situ} experiment setup, and so on.

\setlength{\leftskip}{0pt}

The use of probabilistic ML methods here enables us to maintain uncertainty estimates over both graph structures and their associated parameters during the causal discovery process, and potentially to propagate uncertainty values to downstream tasks. We are further investigating how to reliably quantify uncertainties with human-machine causal discovery. Another promising direction for work is to consider using physics-infused GP methods (discussed in the surrogate modeling motif) as the BO surrogate model -- we explore this and related combinations of motifs in the Integrations section later.

\paragraph{Building a driving simulator for causal reasoning}

Robust and reliable simulation tools can help address the overarching challenge in counterfactual (and causal) reasoning: one never observes the true counterfactual outcome.
In theory, the parameters of simulation environments can be systematically controlled, thereby enabling causal relationships to be established and confounders to be introduced by the researcher.
In practice, simulators for complex environments are typically built on game engines \cite{Chance2021OnDO} that do not necessarily suffice for causal reasoning, including the simulation testbeds for AI agents such as Unity AI \cite{Juliani2018UnityAG} and OpenAI Gym \cite{Brockman2016OpenAIG}.
We suggest several main limitations to overcome:
\begin{enumerate}
    \item Counterfactual reasoning in general is not as simple as changing a variable and observing the change in outcomes. Rather, principled counterfactual reasoning is more precise: quantifying the potential outcomes under the Rubin-Neymann causal model \cite{Rubin2005CausalIU} and within the constraints of potential confounders to derive true cause-effect relationships. For example, the implementation of counterfactual Gaussian Processes \cite{Schulam2017WhatIfRW} to simulate continuous-time potential outcomes of various clinical actions on patients.
    \item \textit{Confounders} -- factors that impact both the intervention and the outcomes -- can be known or measurable in some cases and hidden in others. The challenge here is to enable the researcher to systematically introduce, remove, and examine the impact of many different types of confounders (both known and hidden) while experimenting with simulations.
    As mentioned in the ABM example, only manipulating simulation inputs is subject to preemption confounding.
    \item Prior work with causality-specific simulators has remained in relatively small environments with simplistic entities and actions, for example balls or objects moving on 2D and 3D surfaces \cite{Yi2020CLEVRERCE,Li2020CausalDI,Gerstenberg2021ACS}.
\end{enumerate}

The recent ``CausalCity'' \cite{McDuff2021CausalCityCS} simulation environment aims to provide explicit mechanisms for causal discovery and reasoning -- i.e., enabling researchers to create complex, counterfactual scenarios including different types of confounders with relatively little effort. 
The main novelty over similar simulation environments is the notion of agency: each entity in the simulation has the capability to ``decide'' their low-level behaviors, which enables scenarios to be designed with simple high-level configurations rather than specifying every single low-level action. They suggest this is crucial to creating simulation environments that reflect the nature and complexity of these types of temporal real-world reasoning tasks.
CausalCity addresses problem (3) above with a city driving simulation of configurable, temporal driving scenarios and large parameter spaces. However, the main aims of a causal-first simulator are left unfulfilled, with only parts of problems (1) and (2) addressed: 
The notion of counterfactual reasoning in CausalCity is to forward simulate scenarios with specific conditions or parameters changed, and observe the difference in outcomes, meaning the counterfactual claims are limited to the full sequence of events 
rather than specific cause-effect relationships in the space between simulation inputs and outputs.
For analyzing confounders, CausalCity provides tools to control \textit{extraneous variables} -- all variables which are not the independent variable but could affect the results of the experiment -- which do not necessarily imply confounding, let alone show specific confounding relationships (for instance, as you would in an SCM).

Thus the CausalCity simulator does not include the means for deriving the precise causal mechanisms through direct interventions, nor control of causal inference bounds due to confounder assumptions.
There needs to be tooling for causal modeling directly on the simulator execution trace, as we suggest later in probabilistic programming -based causal simulations.
Nonetheless CausalCity represents a useful step in the direction of simulators for causal reasoning.

\subsection*{Future directions}\label{sec:Causality_future}

Causal reasoning facilitates the simulation of possible futures and prediction of intervention outcomes, leading to more principled scientific discovery and decision-making.
In several of the preceding examples we highlighted ``active'' processes with causal machine learning and simulators. Our belief is further development and validation can bring these active causal reasoning workflows into scientific practice to accelerate discovery and experimentation in diverse domains. 
Further, one can make the case that human-machine inference methods (Fig.~\ref{fig:activesci}) embodied with causal reasoning are necessary to make progress towards the \textit{Nobel Turing Challenge} \cite{Kitano2021NobelTC} to develop AI-scientists capable of autonomously carrying out research to make major scientific discoveries -- it'd difficult to imagine such an AI-scientist without, for example, the ability to intelligently explore-exploit a system to derive its causal structure as in Fig.~\ref{fig:causalBO}.

The specific directions of work we described are,
\begin{itemize}
    \item Causality definitions and inference constraints based on simulation trace relationships
    \item Potential outcomes methods purpose-built for simulation counterfactual reasoning
    \item Tooling for active interventions on running simulations
    \item Tooling for autonomous and human-machine active causal discovery
    \item Game engines and testbeds with infrastructure and programmable interfaces for principled counterfactual (and interventional) reasoning
\end{itemize}

Another direction we propose in the Integrations section later is probabilistic programming as a formalism for causal reasoning, as a natural representation framework and also enabling probabilistic causal reasoning.
Under the umbrella of advancing probabilistic causal reasoning, and continuing the thread from our causal optimization example above, work on causal inference with Bayesian optimization can offer new, improved approaches for decision making under uncertainty. 
Aglietti et al. \cite{Aglietti2020CausalBO} propose an intriguing approach for global causal optimization, and continuing this line work is an important direction -- for example, combining their proposed framework with causal discovery algorithms, to remove dependency on manual encoding graphs while also accounting for uncertainty in the graph structure.

Perhaps the most important contribution of causal reasoning in simulation intelligence systems comes from the fact that correlations may well indicate causal relationships but can never strictly \textit{prove} them. A model, no matter how robust, that reproduces the statistical properties of data, is a rather weak form of validation since there are many possible causal structures for a given set of statistical dependencies. Building simulators with proper representations and tooling for causal inference is critical for validating that simulators faithfully represent the true data generating processes, and ultimately for the reliability of simulation outcomes. 
One must consider the ethical implications of downstream actions and decision-making based on such outcomes, from epidemiology and medicine, to economics and government policy \cite{lavin2021technology}. 
Simulation-based approaches can better deal with cause and effect than purely statistical methods from widely used in AI, ML, and data science, yet similar ethical concerns like model trust and interpretability \cite{Lipton2018TheMO}, and algorithmic fairness \cite{Loftus2018CausalRF,Oneto2020FairnessIM,Fazelpour2020AlgorithmicFF} are crucial.

\hfill \break
\subsection{5. AGENT-BASED MODELING}

In \textit{agent-based modeling (ABM)} \cite{Holland1991ArtificialAA,Bonabeau2002AgentbasedMM}, systems are modeled as collections of autonomous interacting entities (or agents) with encapsulated functionality that operate within a computational world. ABM agents can represent both individual or collective entities (such as organizations or groups). For example, in an artificial economy simulation for quantitatively exploring consequences of policy decisions, agents could be skilled and unskilled workers at varying class-levels in a city environment \cite{Farmer2009TheEN}. Or in biology, individual bacteria can be modeled as agents from population statistics obtained with flow cytometry \cite{Garca2018StochasticIM}. 

ABM combines elements of game theory, complex systems, emergence, computational sociology, multi-agent systems, evolutionary programming, Monte Carlo methods, and reinforcement learning -- more recently we can add physics to this group, with advances in physics-infused (reinforcement) learning (e.g. learning computational fluid dynamics~\cite{Novati2021AutomatingTM}). Complex systems theory, for example, can be used to investigate how the interactions between parts can create collective behaviour within a dynamic system \cite{Mainzer2007ThinkingIC}. And the dynamic interaction between multiple agents or decision-makers, which simultaneously affects the state of the environment, can cause the emergence of specific and novel behavioral patterns, which can be studied with reinforcement learning techniques such as shaping reward structures and development of language between agents.
The decision-making process with autonomous individuals in a bounded rational environment is often heterogeneous and lends itself well to ABM as a tool for analysis \cite{Batty2008GenerativeSS}; individuals represented by agents are dynamically interacting with other agents based on simple rules that will give rise to complex behaviours and patterns of displacement.

It is standard that agents can only acquire new data about their world constructively (through interactions), and agents can have \textit{uncomputable beliefs} about their world that influence their interactions. Uncomputable beliefs can arise from inborn (initially configured) attributes, from communications received from other agents, and/or from the use of non-constructive methods (e.g., proof by contradiction) to interpret acquired data. These uncomputable beliefs enable agents to make creative leaps, i.e. to come up with new ideas about their world not currently supportable by measurements, observations, or logical extrapolations from existing information \cite{Borrill2011AgentbasedMT}.

Formally a standard framework to use is \textit{partial-observable multi-agent Markov Games (MGs)} \cite{Sutton2018ReinforcementLA} with a set of agents interacting in a state and action space that evolves over time—at each time step the environment evolves, and each agent receives an observation, executes an action, and receives a reward. Using machine learning to optimize agent behavior, each agent aims to learn a policy that maximizes its expected reward, which depends on the behavioral policies of the other agents and the environment transition dynamics. In a multi-agent economy setting, for example, the reward is modeled as a utility function, and rational economic agents optimize their total discounted utility over time. This describes the \textit{individual perspective} in ABM, or aiming to achieve one’s own goals. An additional \textit{planner perspective} instead seeks to achieve some notion of social welfare or equilibrium for an interacting population, which usually involves intervening on population-level dynamics through policies, institutions, and centralized mechanisms \cite{Dafoe2020OpenPI}. We describe this in the context of a machine learned social planner to improve societal outcomes with the first example later.

The utility of ABM has proved challenging due to the complexity of realistically modeling human behavior and large environments (despite the ability of ABM to adopt behavioral rules that are deduced from human experiments \cite{Arthur1991DesigningEA}). Recent advances in deep reinforcement learning (RL) provide opportunity to  overcome these challenges by framing ABM simulations as \textit{multi-agent reinforcement learning (MARL)} problems. In MARL, agents need to learn together with other learning agents, creating a non-stationary environment \cite{HernandezLeal2017ASO}. The past few years have seen tremendous progress on games such as go \cite{Silver2016MasteringTG,Schrittwieser2020MasteringAG}, StarCraft II \cite{Vinyals2019GrandmasterLI}, Dota2 \cite{Berner2019Dota2W}, and two-player poker \cite{Brown2019DeepCR}. These represent \textit{two-player zero-sum} games, which are convenient testbeds for multi-agent research because of their tractability: the predefined solutions coincide with the Nash equilibrium, the solutions are interchangeable, and they have worst-case guarantees \cite{Neumann1945TheoryOG}. Yet these testbeds limit research to domains that are inherently adversarial, while non-warfare real-world scenarios such as simulated economies imply cooperative objectives and dynamics, where agents learn to communicate, collaborate, and interact with one another. The MARL field is starting to trend towards games of pure common interest, such as Hanabi \cite{Bard2020TheHC} and Overcooked \cite{Carroll2019OnTU}, as well as team games \cite{Jaderberg2019HumanlevelPI,Baker2020EmergentTU}. \textit{Mixed-motives} settings are more aptly connected to real-world settings, for example in alliance politics \cite{Dafoe2020OpenPI} and tax policy \cite{Zheng2020TheAE}, which are important to research since some strategic dynamics only arise in these settings, such as issues of trust, deception, and commitment problems. The field of \textit{cooperative AI} investigates a spectrum of these common-vs-conflicting interests dynamics, as well as myriad types of cooperative agents—machine-machine, human-machine, human-human, and more complex constellations. While much AI research to date has focused on improving the individual intelligence of agents and algorithms, cooperative AI aims to improve \textit{social intelligence}: the ability of groups to effectively cooperate to solve the problems they face. Please see Dafoe et al.~\cite{Dafoe2020OpenPI} for a thorough overview.

Reinforcement learning (including MARL and cooperative AI) provides a useful toolbox for studying and generating the variety of behaviors agents develop. The primary factor that influences the emergence of agent behavior is the reward structure, which can be categorized as cooperative, competitive, or intrinsic \cite{Gronauer2021MultiagentDR}. Intrinsic motivation implies the agent is maximizing an internal reinforcement signal (intrinsic reward) by actively discovering novel or surprising patterns, a concept which resembles \textit{developmental learning} \cite{Piaget1971TheOO,Gopnik1999TheSI} where organisms have to spontaneously explore their environment and acquire new skills \cite{Barto2013IntrinsicMA,Aubret2019ASO}. In addition to the various reward strategies, the development of language corpora and communication skills of autonomous agents is significant in emerging behaviors and patterns. Gronauer et al.~\cite{Gronauer2021MultiagentDR} categorize this as \textit{referential} and \textit{dialogue}. The former describes cooperative games where the speaking agent communicates an objective via messages to another listening agent, and the latter includes negotiations, question-answering, and other back-and-forth dialogues between agents. Multi-agent communication can in general improve task performance and cumulative reward, but quantifying communication remains an open question -- numerical, task-specific performance measurements provide evidence but do not give insights about the communication abilities learned by the agents. 

It is natural for ABM testbeds to be implemented in game environments; while the AI Economist uses a simple 2-dimensional grid environment, the Dota2 and Starcraft II deep RL examples utilize those full video games. More generally, game engines such as Unity \cite{Juliani2018UnityAG} provide powerful simulation environments for MARL and ABM problems broadly. 
Minecraft, for instance, has been the testbed for RL and human-AI interaction \cite{Shah2021TheMB}, and the ``CausalCity'' \cite{McDuff2021CausalCityCS} simulation environment we discussed earlier is built with the Unreal game engine (leveraging the AirSim \cite{Shah2017AirSimHV} package for simulations for autonomous vehicles). 
Game- and physics-engines are increasingly useful components of the AI toolbox and the SI ecosystem.

\subsection*{Examples}\label{sec:ABM_ex}

The ABM toolbox is applicable for study of complex interactive processes in all scientific disciplines. 
ABM is particularly interesting for simulating and studying systems in social sciences and economics, given the relative ease of mapping agents to recognizable social entities, the natural hierarchical self-organization in social systems, and the interesting results in emergent behavior -- applications range from urban traffic simulation \cite{Balmer2008AgentbasedSO} to humanitarian assistance \cite{Crooks2013GISAA} to emergency evacuations \cite{Schoenharl2006WIPERAM}. 
Complex emergent economic behavior has been previously studied in economics through agent-based modeling \cite{Bonabeau2002AgentbasedMM,Garrido2013AnAB}, but this has largely proceeded without the benefit of recent advances in AI. 
We illustrate AI-infused ABM towards societal intelligence with a couple of examples below, followed by additional domains in life sciences. 

As with the other motifs, we advocate for probabilistic approaches to ABM, to properly calibrate models, assimilate observational data into models, and quantify uncertainties in results. For instance, we may make an agent’s behavior stochastic in order to express our uncertainty in the agent’s behavior, and we may implement Monte Carlo methods to compute a posterior of behaviors rather than single point predictions. In some cases we may be able to build a surrogate model of the ABM simulation (i.e. statistical emulation) for efficient execution and uncertainty quantification \cite{Angione2020UsingML,zhang2020modern}.

\paragraph{Optimizing economic policies}

\begin{figure}[ht]
\centering
\includegraphics[width=1.0\linewidth]
{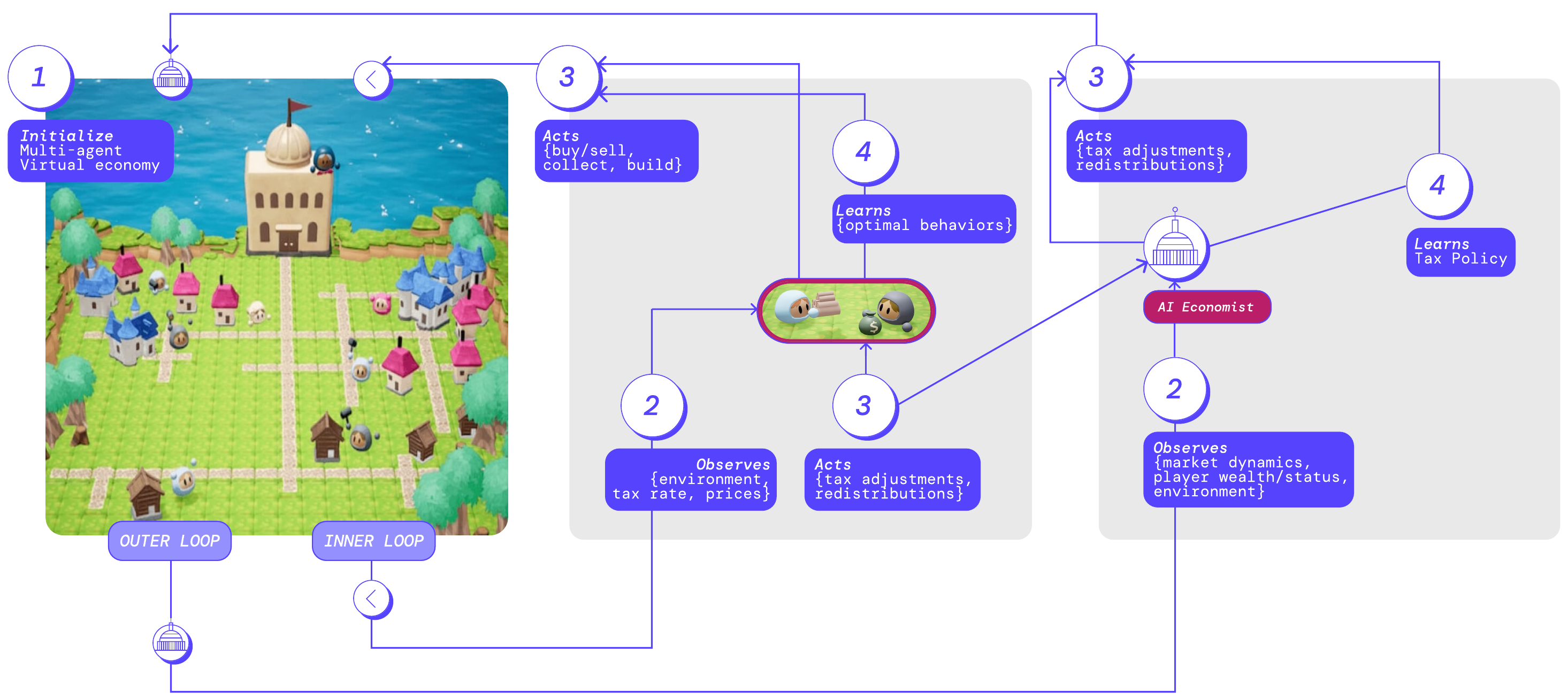}
\caption{
The AI Economist's two-level RL framework \cite{Zheng2020TheAE}. In the inner loop, various types of RL agents gain experience by performing labor, receiving income, and paying taxes, and learn through balancing exploration and exploitation how to adapt their behavior to maximize their utility. In the outer loop, the social planner adapts tax policies to optimize its social objective. The initial framework ran in a simplistic 2D environment with limited agency, whereas we show an example of work-in-progress extending AI Economist methods to more elaborate, complex multi-agent worlds \cite{poke}.
}
\label{fig:aieconomist}
\end{figure}

The ``AI Economist'' \cite{Zheng2020TheAE} aims to promote social welfare through the design of optimal tax policies in dynamic economies using ABM and deep RL. It uses a principled economic simulation with both workers and a policy maker (i.e. individual agents and a planner, respectively), all of whom are collectively learning in a dynamic inner-outer reinforcement learning loop (see Fig. \ref{fig:aieconomist}). In the inner loop, RL agents gain experience by performing labor, receiving income, and paying taxes, and learn through trial-and-error how to adapt their behavior to maximize their utility. In the outer loop, the social planner adapts its tax policy to optimize the social objective--a social welfare function from economic theory that considers the trade-off between income equality and productivity. The AI Economist framework is designed in a way that allows the social planner to optimize taxes for any social objective, which can be user-defined, or, in theory, learned. Another key benefit is the planner in this ABM setup does not need prior world knowledge, prior economic knowledge, or assumptions on the behavior of economic agents.

The data-driven AI Economist simulations are validated by matching the learned behavior to results known from economic theory, for example \textit{agent specialization}. Specialization emerges because differently skilled workers learn to balance their income and effort. For example, in some scenarios agents learn tax-avoidance behaviors, modulating their incomes across tax periods. The dynamic tax schedule generated by the planner performs well despite this kind of strategic behavior. This exemplifies the rich, realistic dynamics and emergent behavior of a relatively simplified economic simulation. The validations also build confidence that agents respond to economic drivers, and demonstrate the value of a data-driven dynamic tax policy versus prior practice.

The AI Economist provides a powerful framework for integrating economics theory with data-driven simulation, producing transparent and objective insights into the economic consequences of different tax policies. Yet there are limitations: the simulations do not yet model human-behavioral factors and interactions between people, including social considerations, and they consider a relatively small economy. It would also be interesting to improve the framework with counterfactual reasoning for \textit{what-if} experimentation -- this is a key value-add of AI-driven simulators in general, which we detail in the causality motif.

\paragraph{Simulating networked acts of violence}

Particularly intriguing use-cases for new ABM approaches are in seemingly random and chaotic human events, such as insurrections and other acts of violence. Political revolutions, for example, are not understood nearly well enough to predict: similar patterns lead to heterogeneous scenarios, the cascading effects of actions (and their timing) have high uncertainty with expensive consequences, origins of emergent behaviors are often unknown and thus unidentifiable in the future, and it is impossible to know which episodes lead to local versus mass mobilization \cite{Moro2016UnderstandingTD}. Other ABM concerns are especially relevant in these applications: data limitations (particularly challenges with multi-modal, sparse, noisy, and biased data), as well as challenges validating simulations (and further, trust).

Recent work applying ABM methods to violence events, such as aiming to model the insurrection events of the Arab Spring \cite{Moro2016UnderstandingTD}, can yield interesting outcomes, but nonetheless suffer from critical roadblocks: existing approaches use simplifying assumptions without real-world validation, cannot incorporate real data (historical nor new), and are not necessarily interpretable nor probe-able. Another is the agent-based insurgency model of Overbey et al.~\cite{Overbey2013ALS}, where they simulate hundreds of thousands of agents that can exhibit a variety of cognitive and behavioral states and actions, in an environment loosely based on the demographic and geophysical characteristics of Ramadi, Iraq in the early 2000s. Large scale parallel compute with GPUs was critical to produce simulations of realistic insurgent and counter-insurgent activities. Yet more acceleration and scaling would be needed for validation, including generating a posterior of potential outcomes for quantifying uncertainty and reasoning about counterfactual scenarios. These two use-cases exemplify the bottleneck ABM faces in geopolitical settings without newer AI-driven simulation techniques.


One important development would be multi-scale formalisms that are amenable to reinforcement learning and machine learning in general. A nice example in this direction is the hierarchical multi-agent, multi-scale setup of Bae \& Moon~\cite{Bae2016LDEFFF}, which was effective in simulating evacuation through a city’s road networks due to bombing attacks, and simulating ambulance effectiveness transporting victims. For overcoming compute bottlenecks, existing HPC systems can be leveraged, but NN-based surrogates would be needed, for acceleration and additional emulation uses (as described in the emulation motif earlier).

Existing works in geopolitics ABM rely solely on mechanistic, theoretical descriptions of human behavior rooted in theory \cite{Epstein2002ModelingCV}. And recent work with deep RL uses simplistic board-game environments \cite{Paquette2019NoPD}. For reliable simulations it’s important to utilize real-world data in a way that complements and validates the theory. In the applications here, we would implement a semi-mechanistic modeling approach that coordinates both mechanistic equations (such as information theoretic models of disinformation \cite{Kopp2018InformationtheoreticMO}) and data-driven function approximations. Further, building this ABM framework in a generative modeling (and more specifically probabilistic programming) way would be essential for simulating the multitudes of outcomes and their full posteriors. The probabilistic programming approach would also behoove counterfactual reasoning, or quantifying the potential outcomes due to various courses of action, as discussed in both the probabilistic programming and causality motifs.

\paragraph{Multi-scale systems biology}

Currently a main challenge in molecular modeling is the complex relationship between spatial resolution and time scale, where increasing one limits the other. There are macroscopic methods that make simplifying assumptions to facilitate modeling of the target molecular system (such as ODEs and PDEs operating at the population-level), and microscopic methods that are much finer in spatial resolution but highly computationally expensive at temporal scales of interest (such as Brownian and molecular dynamics methods). The reader may recall a similar spatial-temporal cascade of simulation methods in the motif on multi-scale modeling (Fig. \ref{fig:brainbloodflow}).

\begin{figure}[ht]
\centering
\includegraphics[width=0.85\linewidth]{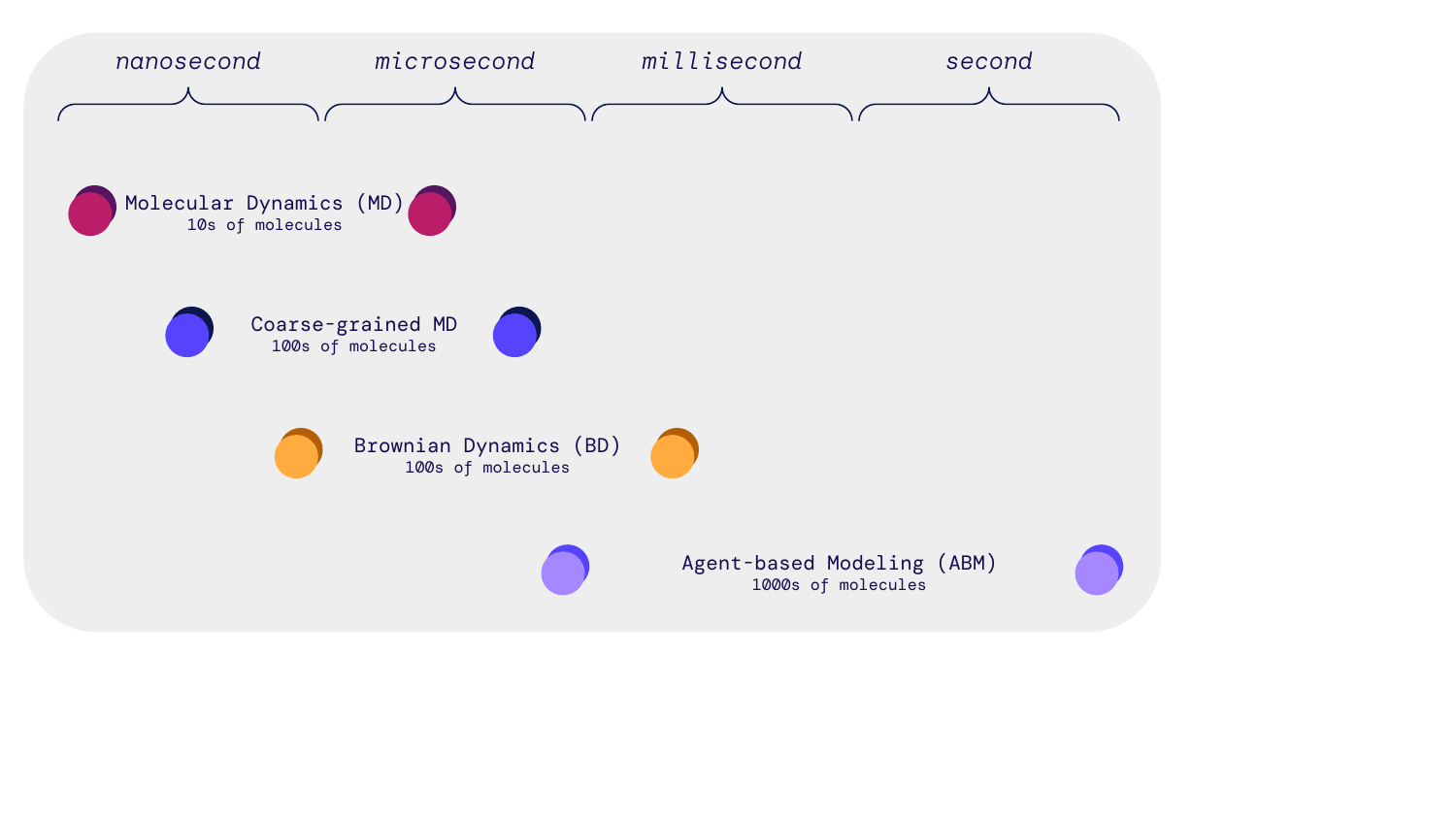}
\caption{
Comparing the size of the molecular system of interest and the timescales achievable by computational modeling techniques (motivated by \cite{Soheilypour2018AgentBasedMI}).
}
\label{fig:abm-scales}
\end{figure}

\begin{figure}[ht]
\centering
\includegraphics[width=0.85\linewidth]{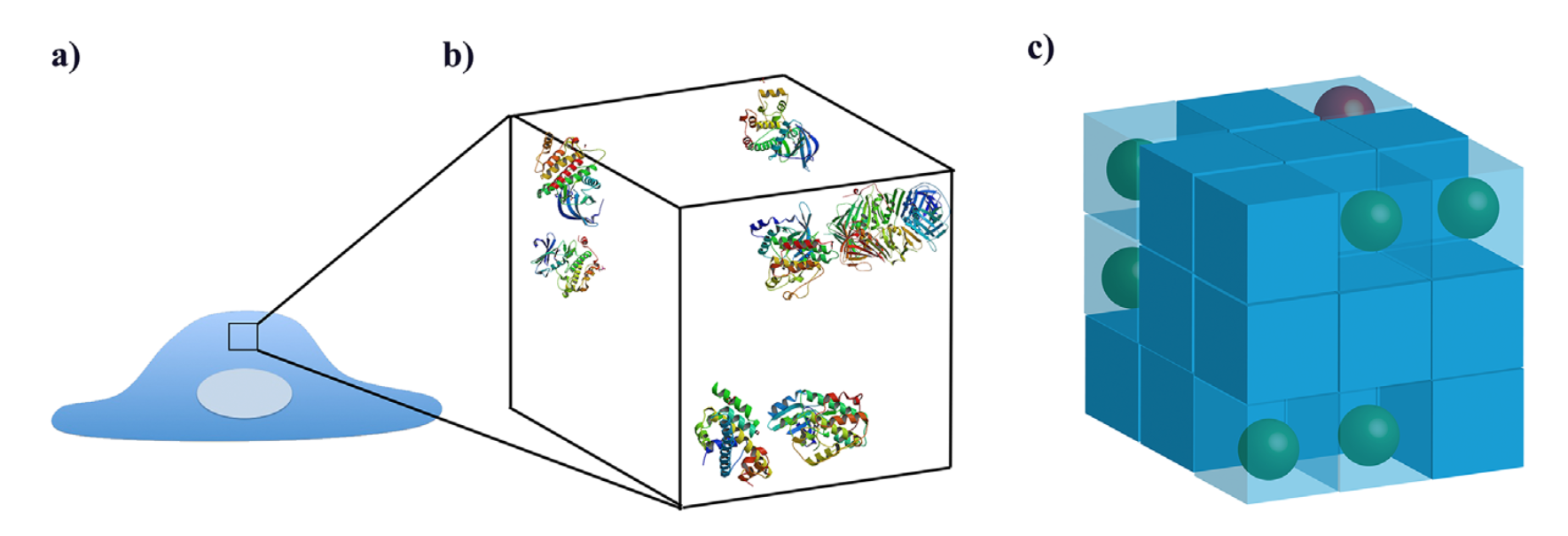}
\caption{
Simplification of representing a cell and its constituent objects (or agents) in a multi-scale agent-based model.
(a) The cell system is characterized by various complex molecular pathways inside the nucleus (grey) and cytoplasm (blue), where each pathway takes place in particular environments and involves different types of molecules. The interactions of these molecules within and between cellular environments lead to the complex behaviors of the overall system. 
(b) A zoomed-in view of seven proteins interacting in an imaginary molecular environment in the cytoplasm. 
(c) ABM representation of the imaginary pathway shown in (b). Information about the environment is projected onto discrete ABM cells. Agents representing biological factors move and interact with other agents and the environment based on rules and mechanisms that are predefined (or potentially learned from data). Agents interact with other agents only when they are in proximity (i.e., green agents as opposed to the red agent). (Reproduced from \cite{Soheilypour2018AgentBasedMI}) 
}
\label{fig:bio-abm}
\end{figure}

To highlight this cascade and how ABM can help overcome existing spatial-temporal systems biology limitations, consider several of the potentially many use-cases for modeling mRNA \cite{Soheilypour2018AgentBasedMI}:
\begin{itemize}
    \item The nuclear export of mRNA transcripts requires a millisecond time scale, which has been explored via ABM \cite{Azimi2014AnAM}, while high-resolution techniques such as Brownian or molecular dynamics are only tractable at nano- and micro-second scales, respectively. Further, these methods can compute 10s to 100s of molecules, whereas ABM can do 1000s.
    \item Cellular processes are often defined by spatial details and constrained environments --  e.g., RNA-binding proteins can travel between the nucleus and the cytoplasm (depending on their binding partners), while others like the nucleo- and cyto-skeleton linker is within the nuclear envelope. These spatial characteristics can be easily incorporated in ABM as prior knowledge that improves modeling (Fig. \ref{fig:bio-abm}).
    \item Tracking of individual particles over time is valuable in study of molecular systems, such as mRNA export. ABM is well-suited for particle tracking -- for instance, recall the virus-propagation example in Fig. \ref{fig:virus} of the causality motif -- and can complement \textit{in vivo} particle tracking advances \cite{Katz2016MappingT} towards cyber-physical capabilities.
    \item The concept of emergence in ABM is valuable here as well -- for instance, Soheileypour \& Mofrad \cite{Soheilypour2016RegulationOR} use ABM to show that several low-level characteristics of mRNA (such as quality control mechanisms) directly influence the overall retention of molecules inside the nucleus. ABM, as a complex systems approach, has the ability to predict how a molecular system behaves given the rules that govern the behavior of individual molecules.
\end{itemize}

Another illuminating example is multi-scale ABM for individual cell signalling, cell population behavior, and extracellular environment. ``PhysiBoSS'' is an example framework to explore the effect of environmental and genetic alterations of individual cells at the population level, bridging the critical gap from single-cell genotype to single-cell phenotype and emergent multicellular behavior \cite{Letort2019PhysiBoSSAM}. 
The ability of ABM to encompass multiple scales of biological processes as well as spatial relationships, expressible in an intuitive modeling paradigm, suggests that ABM is well suited for molecular modeling and translational systems biology broadly.



\paragraph{Generating agent-based model programs from scratch}

In recent work, a combination of agent-based modeling and program synthesis via genetic programming was used to develop a codified rule-set for the behavior of individual agents, with the loss function dependent on the behaviors of agents within the system as a whole \cite{Greig2021GeneratingAM}. 
\textit{Genetic programming} is the process of evolving computer programs from scratch or initial conditions: a set of primitives and operators, often in a domain-specific language. 
In \textit{program synthesis}, the  desired  program  behavior is  specified but the (complete) implementation is not: the synthesis tool determines how to produce an executable implementation. Similarly, the behavior or any  partial implementation can be ambiguously specified, and iteration  with  the  programmer,  data-driven techniques, or both are used to construct a valid solution.

Together, program synthesis via genetic programming allow non-trivial programs to be generated from example data. ABM is an intriguing area to apply this methodology, where emergence and individual-to-macro complexity are main characteristics to study and exploit.
Focusing on two use cases -- flocking behavior and opinion dynamics (or how opinions on a topic spread amongst a population of agents) -- the researchers were able to generate rules for each agent within the overall system and then evolve them to fit the overall desired behavior of the system. This kind of approach is early but has implications for large systems, particularly in sociological or ecological modeling. 
This and related approaches are emblematic of the artificial life field (ALife), which we discuss more in the open-ended optimization motif next.

\subsection*{Future directions}\label{sec:ABM_future}

The previous examples elucidate a central feature of ABM which should help motivate the application of ABM in diverse domains: they are able to generate system-level behaviors that could not have been reasonably inferred from, and often may be counterintuitive to, the underlying rules of the agents and environments alone. 
The implication is there can be simulation-based discovery of emergent phenomena in many systems at all scales, from particle physics and quantum chemistry, to energy and transportation systems, to economics and healthcare, to climate and cosmology. 

There are of course contexts in which the use of ABM warrants thorough ethical considerations, for instance ABM of social systems. In some applications, decision-making and social norms are modeled and inferred (e.g. of functional groups such as households, religious congregations, co-ops, and local governments \cite{BealCohen2021IntragroupDI}), which may unintentionally have elements of bias or ignorance when bridging individual-level assumptions and population-level dynamics \cite{Hammond2015ConsiderationsAB}. There should be awareness and explicit systems in place to monitor these concerns in the use of ABM \textit{and} for downstream tasks because the resulting predictions can inform the design or evaluation of interventions including policy choices.

Broadly in ABM, domain-specific approaches and narrow solutions are the norm. The lack of generalizable methods and unifying frameworks for comparing methods poses a significant bottleneck in research progress. Even more, this limits the utility of ABM for the many researchers interested in interdisciplinary simulations. A basic infrastructure of ABM is needed, with formalisms built on ABM logic and deep RL techniques (to integrate with state-of-art machine learning), and supporting the generic design of simulation engines working with multi-modal data. Building atop differentiable and probabilistic programming (the engine motifs discussed next) would behoove such an ABM framework. This is also a hardware problem: in line with the layer-by-layer integrated nature of simulation and AI computing that we describe throughout this paper, existing ABM simulation platforms for the most part can't take advantage of modern hardware, nor are they scalable to high-performance computing (HPC) deployments (which we discuss later in this work).

The AI Economist and similar works are moving towards an integration of traditional ABM methodology with newer deep RL approaches. In addition to powerful results, this can be a fruitful paradigm for studying language and communication between agents in many environments and settings -- as mentioned earlier, the development of metrics and quantitative tools for studying such communication (and its emergent behaviors) is essential. It will likely be necessary to utilize the machinery of causal inference and counterfactual reasoning, for studying various communication types as well as motivations (e.g. \cite{Jaques2019SocialIA}). 
It will also be critical to build methods in a hierarchical way for multi-scale ABM, to help overcome the burden of simplifying assumptions and approximations about individuals and groups of agents -- notice the close similarity in challenges and methods described in the multi-scale multi-physics motif.

\hfill \break

\section{\textit{The Engine}}


\subsection{6. PROBABILISTIC PROGRAMMING}

\begin{figure}[ht]
\centering
\includegraphics[width=0.7\linewidth]{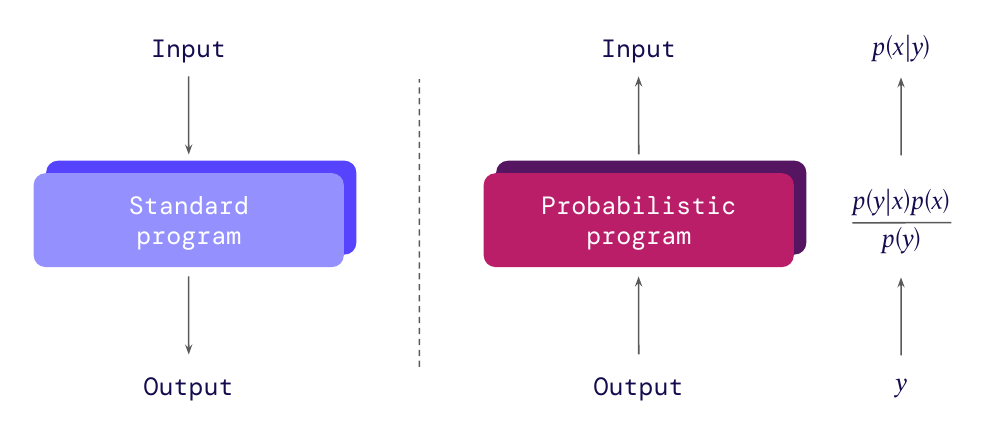}
\caption{
With a probabilistic program we define a joint distribution $p(x,y)$ of latent (unobserved) variables $x$ and observable variables $y$. PPL inference engines produce posterior distributions over unobserved variables given the observed variables or data, $p(x|y) = p(y|x)p(x)/p(y)$. As observed variables typically correspond to the output of a program, probabilistic programming provides us a way to ``invert'' a given program, meaning that we infer program's inputs given an instance of data that corresponds to the program's output.
Example PPL code is shown in Fig. \ref{fig:ppl-code}.
(Original figure credit: Frank Wood)
}
\label{fig:ppl}
\end{figure}

The \textit{probabilistic programming (PP)} paradigm equates probabilistic generative models with executable programs \cite{van2018introduction}. Probabilistic programming languages (PPL) enable practitioners to leverage the power of programming languages to create rich and complex models, while relying on a built-in inference backend to operate on any model written in the language. This decoupling is a powerful abstraction because it allows practitioners to rapidly iterate over models and experiments, in prototyping through deployment. The programs are generative models: they relate unobservable causes to observable data, to simulate how we believe data is created in the real world (Fig.~\ref{fig:ppl}).

PPL pack additional advantages for scientific ML, mainly by design allowing experts to incorporate domain knowledge into models and to export knowledge from learned models. First, PPL are often high-level programming languages where scientists can readily encode domain knowledge as modeling structure -- there is a direct correspondence between probability theory and coding machinery that is obfuscated in standard programming paradigms. Second, since models are encoded as programs, program structure gives us a structured framework within which to reason about the properties and relationships involved in a domain before inference is even attempted \cite{Pfeffer2018StructuredFI}. Crucially, in PPL we get uncertainty reasoning as a standard feature allowing us to express and work with \textit{aleatoric} (irreducible uncertainty due to stochasticity such as observation and process noise) and \textit{epistemic} (subjective uncertainty due to limited knowledge) uncertainty measures.

The main drawback of probabilistic programming is computational cost, which is more pronounced especially in the case of unrestricted (universal, or Turing-complete) PPL \cite{Gordon2014ProbabilisticP} that can operate on arbitrary programs \cite{Goodman2008ChurchAL, Pfeffer2009FigaroA, Mansinghka2014VentureAH, Wood-AISTATS-2014, GoodmanStuhlm14}. Another class of PPL is more restrictive and thus computationally efficient, where constraining the set of expressible models ensures that particular inference algorithms can be efficiently applied \cite{Lunn2009TheBP, winn2009probabilistic, Milch2005BLOGPM, Tran2016EdwardAL}. Note that extending an existing Turing complete programming language with operations for sampling and conditioning results in a universal PPL \cite{Gordon2014ProbabilisticP, Munk2019DeepPS}.

Although a relatively nascent field at the intersection of AI, statistics, and programming languages, probabilistic programming has delivered successful applications across biological and physical sciences. For example, inferring physics from high-energy particle collisions \cite{Baydin2019EtalumisBP}, pharmacometric modeling and drug development \cite{Weber2018BayesianAO}, inferring cell signalling pathways \cite{Merrell2020InferringSP}, predicting neurodegeneration \cite{Lavin2020NeurosymbolicND}, optimal design of scientific experiments \cite{Ouyang2016PracticalOE}, modeling epidemiological disease spread \cite{Wood2020PlanningAI, Witt2020SimulationBasedIF}, untangling dynamics of microbial communities \cite{Ranjeva2019UntanglingTD}, wildfire spread prediction \cite{Joseph2019SpatiotemporalPO}, spacecraft collision avoidance \cite{Acciarini2020SpacecraftCR}, and more. We next discuss several key mechanisms for connecting probabilistic programming and scientific simulation across these and other domains.

\paragraph{\textit{Probabilistic programs as simulators}}
First, \textit{a probabilistic program is itself a simulator}, as it expresses a stochastic generative model of data; that is, given a system or process sufficiently described in a PPL, the forward execution of the program produces simulated data.
Typically one runs a simulator forward to predict the future, yet with a probabilistic program we can additionally infer the parameters given the outcomes that are observed. For example, the neurodegeneration programs in \citet{Lavin2020NeurosymbolicND} define a statistical model with random samplings, from which we can generate probabilistic disease trajectories, and infer parameters for individual biomarkers.
Another advantage of simulating with probabilistic programs is for semi-mechanistic modeling, where Turing-complete PPL provide flexible modeling for integrating physical laws or mechanistic equations describing a system with data-driven components that are conditioned or trained on data observations -- the PPL Turing.jl \cite{Ge2018TuringAL}, for example, illustrates this with ordinary differential equations.
Similarly, some PPL enable semi-parametric modeling \cite{Dalibard2017BOATBA,Lavin2018DoublyBO}, where the inference engine can handle programs that combine parametric and nonparametric models: The parametric component learns a fixed number of parameters and allows the user to specify domain knowledge, while a nonparametric model (e.g. a Gaussian process) learns an unbounded number of parameters that grows with the training data.
These two programming features -- semi-mechanistic and semi-parametric modeling -- along with the overall expressiveness of PPL make them ideal for non-ML specialists to readily build and experiment with probabilistic reasoning algorithms.
Utilizing probabilistic programs as simulators plays an important role in the simulation-based inference motif.

\paragraph{\textit{Compiling simulators to neural networks}}
Second, we have \textit{inference compilation} \cite{Le2017InferenceCA, Harvey2019AttentionFI}, a method for using deep neural networks to amortize the cost of inference in universal PPL models. The transformation of a PPL inference problem into an amortized inference problem using neural networks is referred to as ``compilation'', based on the analogy of compiling computer code between two programming languages. In the context of simulation, one use for inference compilation is as a preprocessing step for probabilistic programming algorithms, where initial runs of the PPL-based simulator are used to train a neural network architecture, which constructs proposal distributions for all the latent random variables in the program based on the observed variables -- this represents one of the simulation-based inference workflows in Fig.~\ref{fig:sbi}.
Another main use for inference compilation is to construct probabilistic surrogate networks (PSN): any existing stochastic simulator, written in any programming language, can be turned into a probabilistic program by adding a small amount of annotations to the source code (for recording and controlling simulator execution traces at the points of stochasticity). Munk et al.~\cite{Munk2019DeepPS} demonstrate this on a non-trivial use-case with composite materials simulation, yet more testing with additional simulators is needed to validate the generalizability claims and robustness.
PSN and other extensions to inference compilation \cite{Casado2017ImprovementsTI} help unify probabilistic programming with existing, non-differentiable scientific simulators in the third and perhaps most powerful mechanism, simulator inversion.

\begin{figure}[ht]
\centering
\includegraphics[width=1.0\linewidth]{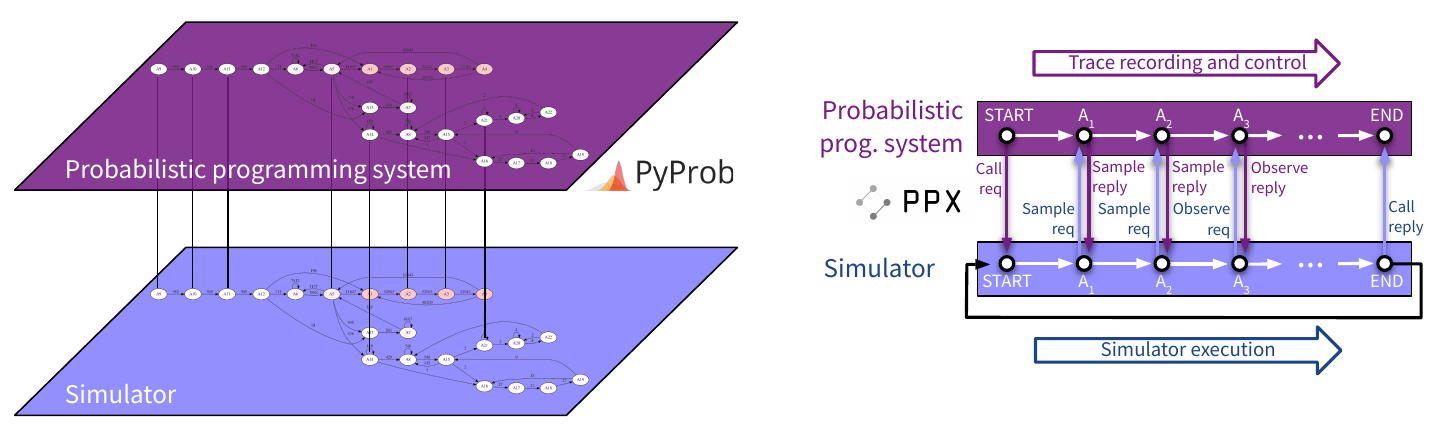}
\caption{
Illustration of the Etalumis probabilistic execution protocol (PPX) that interfaces the PPS (dark top-layer) with an existing simulator (light bottom-layer). As the simulator executes, the PPS inference engine records and controls the execution, as shown by the traces of nodes (representing unique sampled or observed variable labels at runtime). The two traces interface at random number draws and for conditioning in the simulator, corresponding to ``sample'' and ``observe'' statements in a typical PPL, respectively.
}
\label{fig:etalumis}
\end{figure}

\paragraph{\textit{Simulator inversion}}
The \textit{Etalumis} project (``simulate'' spelled backwards) uses probabilistic programming methods to invert existing, large-scale simulators via Bayesian inference \cite{Baydin2019EtalumisBP}, requiring minimal modification of a given simulator's source code. Many simulators model the forward evolution of a system (coinciding with the arrow of time), such as the interaction of elementary particles, diffusion of gasses, folding of proteins, or the evolution of the universe on the largest scale. The task of inference refers to finding initial conditions or global parameters of such systems that can lead to some observed data representing the final outcome of a simulation. In probabilistic programming, this inference task is performed by defining prior distributions over any latent quantities of interest, and obtaining posterior distributions over these latent quantities conditioned on observed outcomes (for example, experimental data) using Bayes rule. This process, in effect, corresponds to \textit{inverting the simulator} such that we go from the outcomes towards the inputs that caused the outcomes. A special protocol, called the probabilistic programming execution protocol (PPX) (Fig.~\ref{fig:etalumis}), is used to interface PPL inference engines with existing stochastic simulator code bases in two main ways:
\begin{enumerate}
    \item Record execution traces of the simulator as a sequence of sampling and conditioning operations in the PPL. The execution traces can be used for inspecting the probabilistic model implemented by the simulator, computing likelihoods, learning surrogate models, and generating training data for inference compilation NNs -- see Fig. \ref{fig:etalumis}.
    \item Control the simulator execution with the PPL by ``hijacking'' the simulator's random number generator. More precisely, general-purpose PPL inference guides the simulator by making random number draws not from the prior $p(x)$ but from proposal distributions $q(x|y)$ that depend on observed data $y$ \cite{GramHansen2019HijackingMS} -- see Fig. \ref{fig:hijack}.
\end{enumerate}

These methods have been demonstrated on massive, legacy simulator code bases, and scaled to high-performance computing (HPC) platforms for handling multi-TB data and supercomputer-scale distributed training and inference. The ability to scale probabilistic inference to large-scale simulators is of fundamental importance to the field of probabilistic programming and the wider scientific community.

\subsection*{Examples}\label{sec:PP_ex}

\paragraph{Compilation and inversion of a particle physics simulator} The Standard Model of particle physics has a number of parameters (e.g., particle masses) describing the way particles and fundamental forces act in the universe. In experiments such as those at the Large Hadron Collider (described in the simulation-based inference motif earlier), scientists generate and observe events resulting in cascades of particles interacting with very large instruments such as the ATLAS and CMS detectors. The Standard Model is generative in nature, and can be seen as describing the probability space of all events given the initial conditions and model parameters. \citet{Baydin2019EtalumisBP,baydin2019efficient} developed the Etalumis approach originally to work with the state-of-the-art Sherpa simulator \cite{osti_1562545}, an implementation of the Standard Model in C++. Based on the PPX protocol, they worked with this very large scale simulator involving approximately one million lines of C++ code and more than 25k latent variables and performed the first instance of Bayesian inference in this setting, demonstrating the approach on the decay of the $\tau$ (tau) lepton. Due to the scale of this simulator, the work made use of the Cori supercomputer at Lawrence Berkeley National Lab, mainly for distributed data generation and training of inference compilation networks for inference in Sherpa.

\begin{figure}[ht]
\centering
\includegraphics[width=0.9\linewidth]{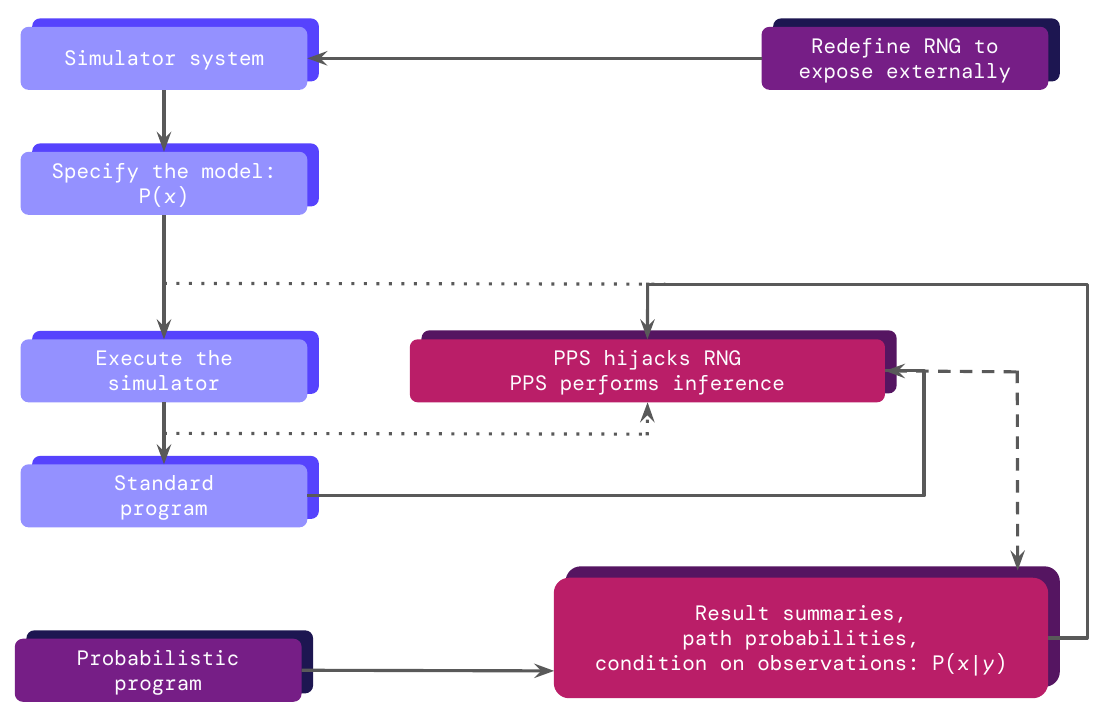}
\caption{The process for ``hijacking'' the random number generator (RNG) of a generic population-based epidemiology simulator (such as OpenMalaria); the outline colors blue, red, and purple respectively represent processes linked to the existing simulator codebase, the hijacking probabilistic programming system, and external dataflows. (Reproduced from \citet{GramHansen2019HijackingMS})
}
\label{fig:hijack}
\end{figure}

\paragraph{Probabilistic, efficient, interpretable simulation of epidemics} 
Epidemiology simulations have become a fundamental tool in the fight against the epidemics and pandemics of various infectious diseases like AIDS, malaria, and Covid \cite{schroederdewitt2020simulation}. However, these simulators have been mostly limited to forward sampling of epidemiology models with different sets of hand-picked inputs, without much attention given to the problem of inference. Even more, inefficient runtimes limit the usability for iterating over large spaces of scenarios and policy decisions. Motivated by the Covid-19 pandemic, \citet{wood2020planning} demonstrated how parts of the infectious disease-control and policy-making processes can be automated by formulating these as inference problems in epidemiological models. Using the Etalumis approach, \citet{GramHansen2019HijackingMS} developed a probabilistic programming system to turn existing population-based epidemiology simulators into probabilistic generative models (Fig~\ref{fig:hijack}). Applied to two state-of-the-art malaria simulators (namely EMOD \cite{Bershteyn2018ImplementationAA} and OpenMalaria \cite{Smith2008TowardsAC}) the system provides interpretable introspection, full Bayesian inference, uncertainty reasoning, and efficient, amortized inference.
This is accomplished by ``hijacking'' the legacy simulator codebases by overriding their internal random number generators. Specifically, by replacing the existing low-level random number generator in a simulator with a call to a purpose-built universal PPL ``controller'', the PPL can then track and manipulate the stochasticity of the simulator as it runs. The PPL controller can then run a handful of novel tasks with the hijacked codebase: running inference by conditioning the values of certain data samples and manipulating others, uncovering stochastic structure in the underlying simulation code, and using the execution traces from the PPL backend to automatically produce result summaries. The resulting execution traces can potentially be used for causal chain discovery (for instance, deriving the infection spread from host to host) as described in the causality motif.

\begin{figure}[!ht]
\centering
\includegraphics[width=0.85\linewidth]{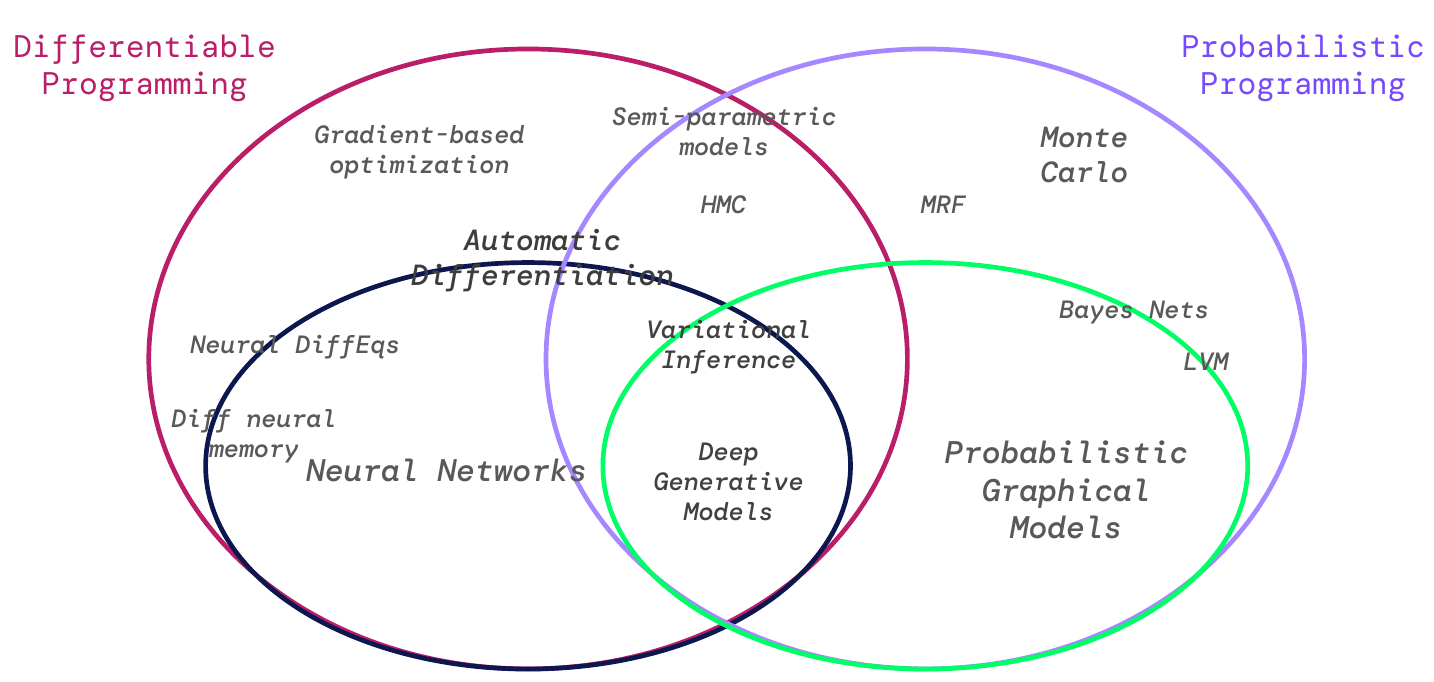}
\caption{
Probabilistic programming expands the scope of probabilistic graphical models (PGM) by allowing for inference over arbitrary models as programs, including programming constructs such as recursion and stochastic branching.
Analogously, differentiable programming is a generalization of neural networks and deep learning, allowing for gradient-based optimization of arbitrarily parameterized program modules.
Just as MCMC-based inference engines drive PPL, autodiff is the engine behind DP.
(Note that ``differentiable neural memory'' refers to methods such as Neural Turing Machines \cite{Graves2014NeuralTM} and Differentiable Neural Computers \cite{Graves2016HybridCU}; ``neural differential equations'' refers to Neural ODE \cite{Chen2018NeuralOD} and Universal DiffEQ (UDE) \cite{Rackauckas2020UniversalDE}; HMC is Hamiltonian Monte Carlo \cite{Betancourt2017ACI,Hoffman2014TheNS}, MRF is Markov Random Fields, and LVM is Latent Variable Models \cite{Louizos2017CausalEI,John1986LatentVM}.)
}
\label{fig:dppp}
\end{figure}

\paragraph{The probabilistic programming advantage for generating molecules}
As we've covered in several motifs and integrations in this paper, designing and discovering new molecules is key for addressing some of humanity's most pressing problems, from developing new medicines to providing agrochemicals in the face of shifting populations and environments, and of course discovering new materials to tackle climate change. The molecular discovery process usually proceeds in design-make-test-analyze cycles (shown in Fig. \ref{fig:inverse-batteries}).
ML can help accelerate this process (ultimately reducing the number of expensive cycles) by both broadening the search and by searching more intelligently. That respectively amounts to learning strong generative models of molecules that can be used to sample novel molecules for downstream screening and scoring tasks, and finding molecules that optimize properties of interest (e.g. binding affinity, solubility, non-toxicity, etc.) \cite{Bradshaw2020BarkingUT}. 
We've suggested gaps in this process from lack of information sharing across phases, particularly the necessity for communication between the molecule design / lead generation phase and the synthesis phase -- see for instance Fig. \ref{fig:pipeline}. One possibility to better link up the design and make steps in ML driven molecular generation is to explicitly include synthesis instructions in the design step. To this end, Bradshaw et al. \cite{Bradshaw2020BarkingUT} present a framework for multi-step molecular synthesis routes: directed acyclic graphs (DAG) represent is the synthetic routes, which are arranged in a hierarchical manner, where a novel neural message passing procedure exchanges information among multiple levels, all towards gradient-based optimization of the system that is enabled by differentiable programming.
Intuitively this works as follows: The model aims to generate ``recipes'' (i.e., synthesis trees or synthesis graphs as DAGs) which start from a small number of building blocks (e.g., easily purchasable compounds) that then can be combined (possibly iteratively) to produce progressively more complex molecules. We assume that a ``reaction outcome predictor'' exists, which can be anything that takes in a set of reactants and outputs a distribution over products. Given a dataset of potential products and their synthetic routes, this can then be fit as structured deep generative model, which generates a distribution over molecules by first selecting a handful of easily obtainable compounds as starting points, then iteratively combining them. This generative model can then be used for Bayesian optimization of molecules (or perhaps open-ended optimization in SI future work), in a way which regularizes the ``suggested'' molecules to only be those that have a high probability of being synthesized reliably in a lab. We discuss similar pipelines and methods in the Integrations section later.
The overall structure of the probabilistic model is rather complex as it depends on a series of branching conditions. It's thus advantageous to frame the entire generative procedure as a probabilistic program -- notice our location in reference to Fig. \ref{fig:dppp} at the intersection of DP and PP. Such a probabilistic program defines a distribution over DAG serializations, and running the program forward will sample from the generative process. The program can just as easily be used to evaluate the probability of a DAG of interest by instead accumulating the log probability of the sequence at each distribution encountered during the execution of the program. Inspecting the PPL execution traces may provide deeper levels of interpretability for the molecule generation process.
This exemplifies the utility of probabilistic programs for ``partially specified'' models, which are the rich area between data-driven (black-box, deep learning) and fully-specified simulators. We give another example of this in the Integrations section later, also in the context of molecular synthesis.

\paragraph{Simulation-based common sense cognition}

The pursuit of common sense reasoning has largely been missing from mainstream AI research, notably characteristics such as generalizability and causal reasoning \cite{Lake2016BuildingMT,George2017AGV}.
To achieve common sense, and ultimately general machine intelligence,
Marcus \& Davis \cite{Marcus2021InsightsFA} suggest starting by developing systems that can represent the core frameworks of human knowledge: time, space, causality, basic knowledge of physical objects and their interactions, basic knowledge of humans and their interactions.
Practical pursuit of this objective has largely focused on reverse engineering the ``common sense core'' from cognitive- and neuro-sciences: \textit{intuitive physics} \cite{Tgls2011PureRI} and \textit{intuitive psychology}, or basic representations of objects in the world and the physics by which they interact, as well as agents with their motives and actions \cite{Liu2017TenmontholdII}.
In particular, studying the hypothesis that many intuitive physical inferences are based on a \textit{mental physics engine} that is analogous in many ways to the \textit{machine physics engines} used in building interactive video games \cite{Ullman2017MindGG}, and by modeling the intuitive physics and intuitive psychology of children as mental simulation engines based on probabilistic programs \cite{Lake2016BuildingMT, Ullman2020BayesianMO}. 

To capture these kinds of ideas in engineering terms probabilistic programming is needed; PPL integrate our best ideas on machine and biological intelligence across three different kinds of mathematics: symbolic manipulation (i.e., algebra and logic), Bayesian inference (probability), and neural networks (calculus).
By wrapping physics engine programs and physics simulators inside frameworks for probabilistic modeling and inference, we gain the powerful tools necessary to capture common sense intuitive physics. In the works of Tenenbaum et al. this has been called ``the simulator in your head'',
and also ``theory of mind'' \cite{Gershman2016PlansHA},
which has a purpose distinct from the typical use of game engines in most of recent AI research: to capture the mental models that are used for online inference and planning, rather than using games for training a NN or other model over a massive amount of simulated data.

These common sense cognition concepts have been demonstrated in a series of simulation-based cognitive science experiments to test and analyze the PPL-based intuitive physics engine \cite{Battaglia2013SimulationAA,Hamrick2016InferringMI}, and the intuitive psych engine \cite{Baker2017RationalQA} where the same kind of model can be used to infer jointly what somebody wants and also what somebody believes, (because people sometimes take inefficient actions \cite{Gershman2016PlansHA}). 
Uncertainty reasoning is once again key here: the PPL-based approach leads to a more robust machine intelligence because any real-world AI agent will never have perfect perception nor perfect physics models, so, like the human brain, making probabilistic inferences based on the available data and models is a robust, generalizable approach. 
Follow-up neuroscience experiments with Tenenbaum et al. aim to show where and how your intuitive physics engine works, and how the neural mechanisms connect synergistically to other cognitive functions based on their neural locus.
An implication is that these kinds of probabilistic program models are operative even in the minds and brains of young babies, which may bear significance in developing methods for \textit{cooperative AI}, which we discuss in the agent-based modeling and uncertainty reasoning sections.

As we noted earlier, PPL sampling-based inference is predominant but in general slow, and certainly restrictive with games/graphics engines. We can alternatively use NNs to learn fast bottom-up approximations or inverses to these graphics engines, also known as ``vision as inverse graphics'' \cite{Mansinghka2013ApproximateBI,Kulkarni2015PictureAP}.
Looking to lean on NNs, PhysNet \cite{Wang20183DPhysNetLT} attempts to treat intuitive physics as a pattern recognition task. Although able to reproduce some of the stability-prediction results, typical of deep learning the PhysNet approach required massive amounts of training and still cannot generalize beyond the very specific physical object shapes and numbers.

Now a physics engine defined as a probabilistic program has to be a program that inputs and outputs other programs. This ``hard problem of learning'' 
represents a highly challenging search problem to find good programs, yet humans can solve that problem so machine intelligence should be able to as well.
Bayesian Program Learning \cite{Lake2015HumanlevelCL} is an early example of this but there is still considerable progress to be made.
Additional notable works towards learning as programming include ``the child as hacker'' \cite{Rule2020TheCA}, and ``Dream Coder''
which combines neural networks with symbolic program search to learn to solve programs in many different domains, essentially learning to construct a deep library of programs \cite{Ellis2020DreamCoderGG}.

\begin{figure}[!h]
\centering
\includegraphics[width=1.0\linewidth]{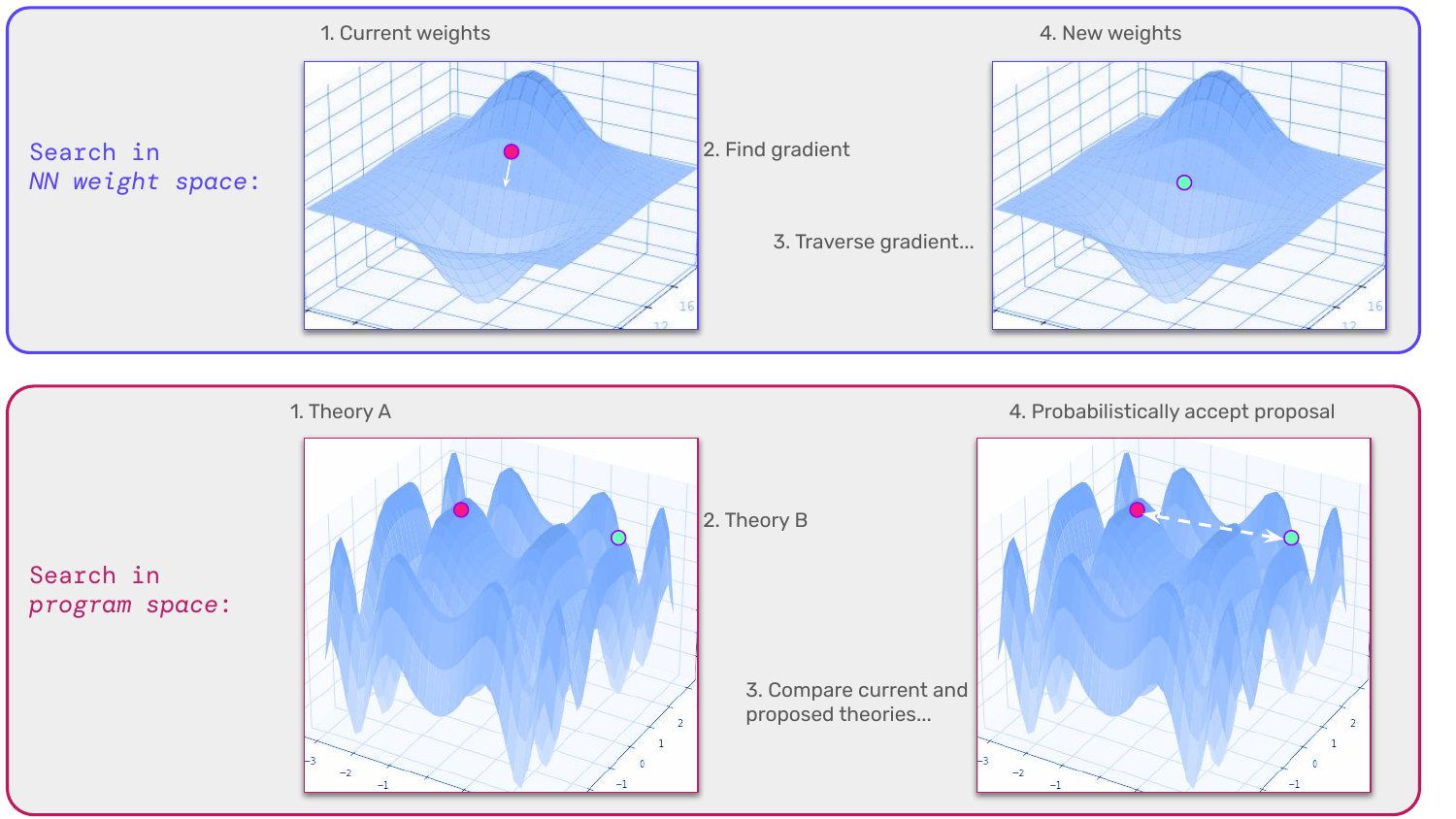}
\caption{Search spaces reflecting the ``hard problem of learning'', where navigating the space of programs (bottom) is far more challenging than the space of neural net parameterizations (top). (Inspired by \cite{Ullman2020BayesianMO}.)
}
\label{fig:pp-induction}
\end{figure}

\subsection*{Future directions}\label{sec:PP_future}

Probabilistic programming and simulation should continue to advance the cognitive- and neuro-sciences work we described above from Tenenbaum et al. The hypothesis, with a growing body of support, is that a simulator or game engine type of representation is what evolution has built into our brains in order to understand the world of objects and agents; babies start off with a minimal proto game engine, learn with the game engine and expand it to cases not necessarily built for, and use this as the subsequent foundation for learning everything else.

Another general area to advance and scale PPL use is in simulation-based inference. Specifically in the context of human-machine inference, using probabilistic programs as the simulator has many potential advantages: modularity in modeling systems, general-purpose inference engines, uncertainty quantification and propagation, and interpretability. See Fig. \ref{fig:activesci} and the simulation-based inference motif.

With the Etalumis project we now have the ability to control existing scientific simulators at scale and to use them as generative, interpretable models, which is highly valuable across simulation sciences where interpretability in model inference is critical. The proof of concept was developed in particle physics with the Large Hadron Collider’s (LHC) ``Sherpa'' simulator at scale. This enables Bayesian inference on the full latent structure of the large numbers of collision events produced at particle physics accelerators such as the LHC, enabling deep interpretation of observed events. Work needs to continue across scientific domains to validate and generalize these methods, which has begun in epidemiology \cite{GramHansen2019HijackingMS,Witt2020SimulationBasedIF} and astrophysics \cite{Acciarini2020SpacecraftCR}. Other promising areas that are bottlenecked by existing simulators include hazard analysis in seismology \cite{Heinecke2014PetascaleHO}, supernova shock waves in astrophysics \cite{Endeve2012TURBULENTMF}, market movements in economics \cite{Raberto2001AgentbasedSO}, blood flow in biology \cite{Perdikaris2016MultiscaleMA}, and many more. Building a common PPL abstraction framework for different simulators would be ideal, for experiment-reproducibility and to further allow for easy and direct comparison between related or competing simulators.

In general, the intersection of probabilistic programming and scientific ML is an actively growing ecosystem. Still, much more progress stands to be made in computational efficiency, both in the inference algorithms and potentially in specialized compute architectures for PPL. We also believe the development of causal ML methods within PPL will be imperative, as the abstractions of probabilistic programming provide opportunity to integrate the mathematics of causality with ML software.
The larger opportunity for probabilistic programming is to expand the scope of scientific modeling: Just as Turing machines are universal models of computation, probabilistic programs are universal probabilistic models -- their model space $M$ comprises all computable probabilistic models, or models where the joint distribution over model variables and evidence is Turing-computable \cite{Ullman2020BayesianMO}.

\hfill \break
\subsection{7. DIFFERENTIABLE PROGRAMMING}

\textit{Differentiable programming (DP)} is a programming paradigm in which derivatives of a program are automatically computed and used in gradient-based optimization in order to tune program to achieve a given objective. DP has found use in a wide variety of areas, particularly scientific computing and ML.

From the ML perspective, DP describes what is arguably the most powerful concept in deep learning: parameterized software modules that can be trained with some form of gradient-based optimization. In the DP paradigm, neural networks are constructed as differentiable directed graphs assembled from functional blocks (such as feedforward, convolutional, and recurrent elements), and their parameters are learned using gradient-based optimization of an objective describing the model’s task \cite{Baydin2017AutomaticDI}. The ML field is increasingly embracing DP and freely composing NN building blocks in arbitrary algorithmic structures using control flow, as well as the introducing novel differentiable architectures such as the Neural Turing Machine \cite{Graves2016HybridCU} and differentiable versions of data structures such as stacks, queues, deques \cite{Grefenstette2015LearningTT}.

DP frameworks for deep learning have dominated the ML landscape during the past decade, most notably auto-differentiation frameworks such as Theano \cite{Bergstra2010TheanoAC}, TensorFlow \cite{Abadi2016TensorFlowAS}, and PyTorch \cite{Paszke2017AutomaticDI}. In contrast to deep learning, which mainly involves compositions of large matrix multiplications and nonlinear element-wise operations, physical simulation requires complex and customizable operations due to the intrinsic computational irregularities, which can lead to unsatisfactory performance in the aforementioned frameworks. To this end, recent works have produced powerful, parallelizable DP frameworks such as JAX \cite{jax2018github, Frostig2018CompilingML} and DiffTaichi \cite{Hu2020DiffTaichiDP} for end-to-end simulator differentiation (and thus gradient-based learning and optimizations), and each can integrate DP into simulator code. Both have promising applications in physical sciences, which we discuss more in the context of examples below. Further, the Julia language and ecosystem \cite{bezanson2017julia} provides system-wide differentiable programming, motivated by scientific and numerical computing, with recent AI integrations for deep learning and probabilistic programming.

\paragraph{\textit{Automatic differentiation (AD)}}
At the intersection of calculus and programming, AD is the workhorse of DP. AD involves augmenting the standard computation with the calculation of various derivatives automatically and efficiently \cite{Baydin2017AutomaticDI}. More precisely, AD performs a non-standard interpretation of a given computer program by replacing the domain of the variables to incorporate derivative values and redefining the semantics of the operators to propagate derivatives per the chain rule of differential calculus. AD is not simply non-manual differentiation as the name suggests. Rather, AD as a technical term refers to a specific family of techniques that compute derivatives through accumulation of values during code execution to generate numerical derivative evaluations rather than derivative expressions. This allows accurate evaluation of derivatives at machine precision with only a small constant factor of overhead and ideal asymptotic efficiency. Availability of general-purpose AD greatly simplifies the DP-based implementation of ML architectures, simulator codes, and the composition of AI-driven simulation, enabling their expression as regular programs that rely on the differentiation infrastructure. See Baydin et al. \cite{Baydin2017AutomaticDI} for a thorough overview.

\begin{figure}[!t]
\centering
\includegraphics[width=0.95\linewidth]{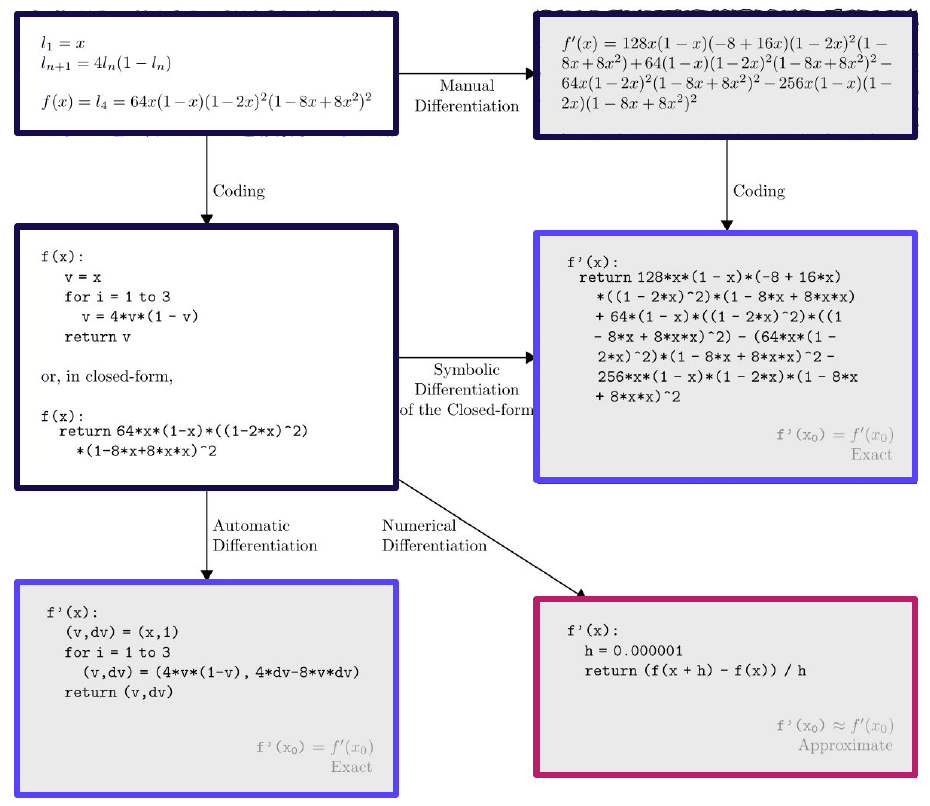}
\caption{
A taxonomy of differentiating mathematical expressions and computer code, taking as an example a truncated logistic map (upper left). Symbolic differentiation (center right) gives exact results but requires closed-form input and suffers from expression swell, numerical differentiation (lower right) has problems of accuracy due to round-off and truncation errors, and automatic differentiation (lower left) is as accurate as symbolic differentiation with only a constant factor of overhead and support for control flow. (Reproduced from \citet{Baydin2017AutomaticDI})
}
\label{fig:autodiff}
\end{figure}

\begin{figure}[!ht]
\centering
\includegraphics[width=0.9\linewidth]{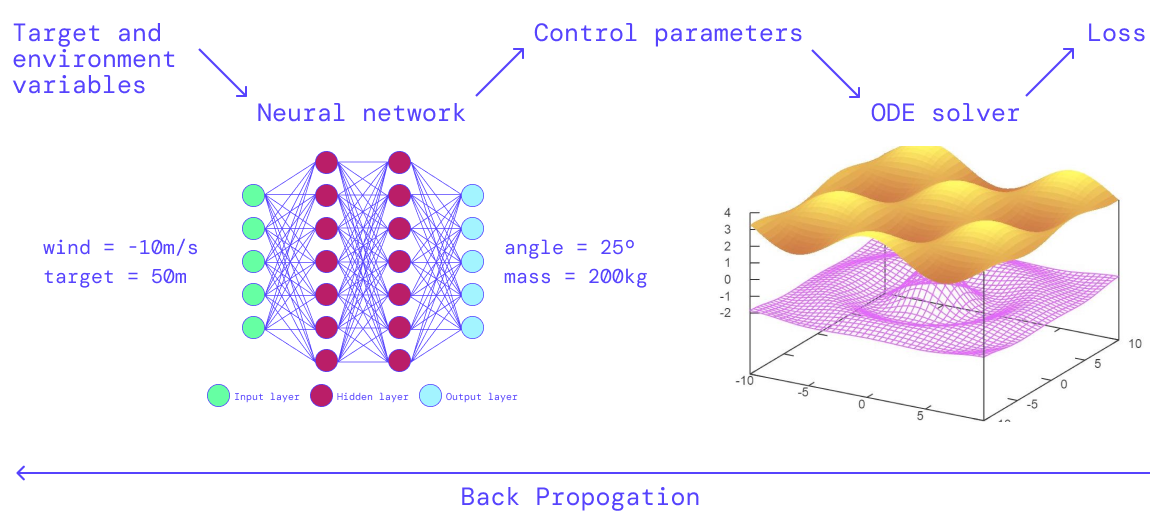}
\caption{
A neural network surrogate expressed in sequence with an ODE solver within a differentiable programming framework, which allows end-to-end learning by backpropagating the output loss through full system. This example illustrates an inverse problem: given the situation (wind, target) we need to solve for optimal control parameters (angle, weight) for the physical system (represented by an ODE-based simulator). (Figured inspired by Chris Rackauckas \cite{Innes2019ADP})
}
\label{fig:ode-inverse}
\end{figure}

In the context of simulators, DP, like probabilistic programming, offers ways to exploit advances in ML \cite{Cranmer2020TheFO} and provides an infrastructure to incorporate these into existing scientific computing pipelines. An intriguing approach with many applications is machine-learned surrogate modeling with scientific simulators and solvers. In Fig. \ref{fig:ode-inverse} we illustrate one example of chaining a neural network surrogate with a simulator (as an ODE solver) to solve an \textit{inverse problem}, or inferring hidden states or parameters from observations or measurements. DP enables this approach because the end-to-end differentiable workflow, including both simulation code and ML model, can backpropagate errors and make updates to model parameters.
We discuss this and related methods in depth in the multi-physics and surrogate modeling motifs, and how all the motifs enable a spectrum of inverse problem solving capabilities.

\subsection*{Examples}\label{sec:DP_ex}

\paragraph{Differentiable pipeline for learning protein structure}

\begin{figure}[!h]
\centering
\includegraphics[width=1.0\linewidth]{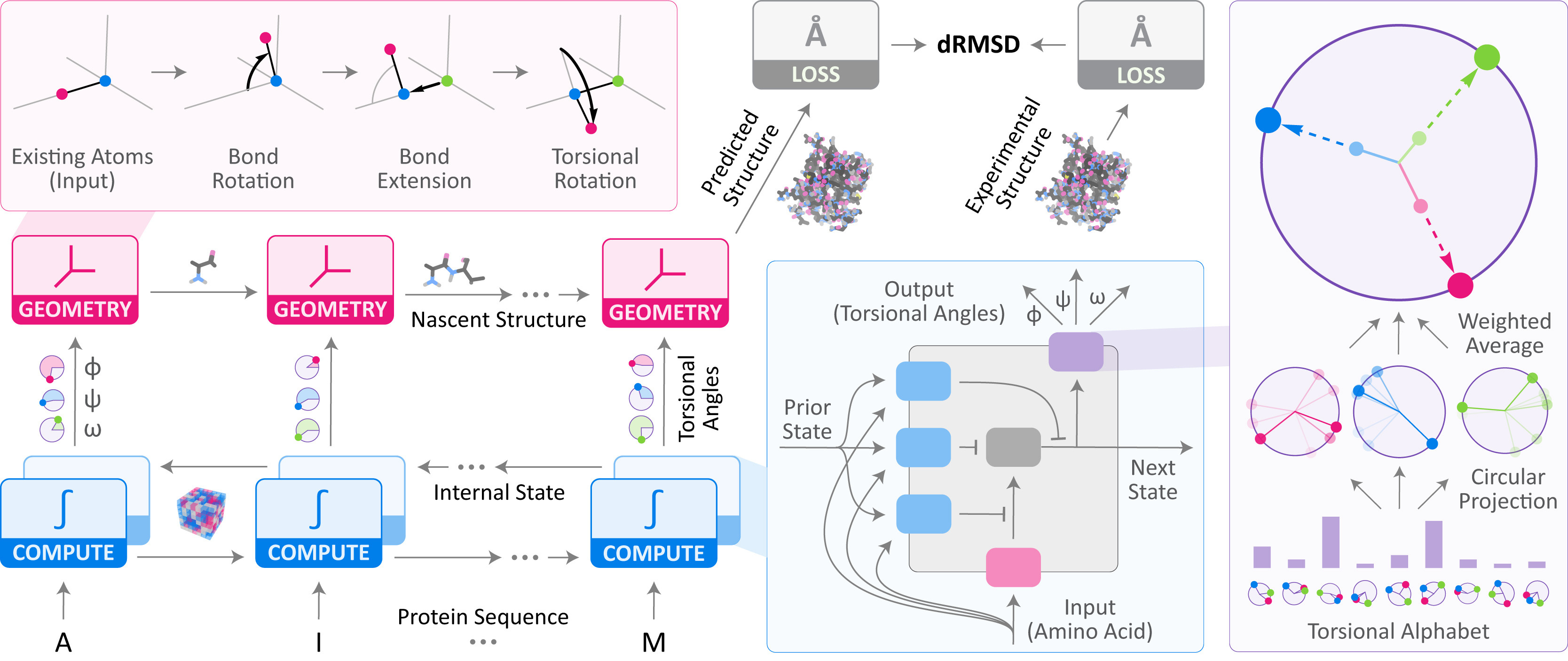}
\caption{
The recurrent geometric network (RGN) of \cite{AlQuraishi2019EndtoEndDL}. The model takes a sequence of amino acids and position-specific scoring matrices as input and outputs a 3D structure by running three stages: computation (left, blue blocks), geometry (left, pink blocks), and assessment (right steps).
(Original figure from Mohammed AlQuraishi \cite{AlQuraishi2019EndtoEndDL})
}  
\label{fig:diff-protein}
\end{figure}

Computational approaches to protein folding not only seek to make structure determination faster and less costly (where protein structure is a fundamental need for rational drug design), but also aim to understand the folding process itself. Prior computational methods aim to build explicit sequence-to-structure maps that transform raw amino acid sequences into 3D structures \cite{Kolinski2011MultiscaleAT}, either with expensive molecular dynamics (MD) simulations or ``fragment assembly'' methods that rely on statistical sampling and a given template solution. Another category of prior methods are ``co-evolution'' where multiple sequence alignment is used to infer geometric couplings based on evolutionary couplings, yet in practice to go from the sequence contact map to a full 3D structure is only reliable 25\%–50\% of the time \cite{Ovchinnikov2017ProteinSD}.
In general there are major shortcomings in the prior work such as inability to predict structural consequences from slight changes or mutations \cite{Hopf2017MutationEP} and inability to work with \textit{de novo} sequences \cite{AlQuraishi2019EndtoEndDL}, not to mention the computational burdens (for example, the prior state-of-art pipeline that combines template-based modeling and simulation-based sampling takes approximately 20 CPU-hours per structure \cite{Zhang2018TemplatebasedAF}).
To address the above limitations, AlQuraishi \cite{AlQuraishi2019EndtoEndDL} proposes differentiable programming to machine-learn a model that replaces structure prediction pipelines with differentiable primitives. The approach, illustrated in Fig. \ref{fig:diff-protein}, is based on four main ideas: (1) encoding protein sequences using a recurrent neural network, (2) parameterizing (local) protein structure by torsional angles to enable a model to reason over diverse conformations without violating their covalent chemistry, (3) coupling local protein structure to its global representation via recurrent geometric units, and (4) a differentiable loss function to capture deviations between predicted and experimental structures that is used to optimize the recurrent geometric network (RGN) parameters. Please see the original work for full details \cite{AlQuraishi2019EndtoEndDL}, including thorough experiments comparing quantitative and qualitative properties of the DP-based approach (RGN) versus prior art. One important highlight is the RGN approach is \textit{6-7 orders of magnitude faster} than existing structure prediction pipelines, although there is the upfront cost of training RGN for weeks to months.
Beyond results that overall surpass the prior methods (including performance on novel protein folds), the differentiable structure approach can be useful towards a bottom-up understanding of systems biology. AlphaFold, the recent breakthrough approach that surpasses RGN and everything else in the field \cite{Jumper2021HighlyAP}, also exploits autodiff (Fig. \ref{fig:dppp} but in a deep-learned way that doesn't necessarily improve scientific knowledge of protein folding; more on AlphaFold is mentioned in the Honorable Mentions subsection later.
A learning-based approach, in contrast to a deterministic pipeline, can provide valuable information towards future experiments and scientific knowledge. In this case of machine-learning protein sequences, the resulting model learns a low-dimensional representation of protein sequence space, which can then be explored, for example with search and optimization algorithms we've discussed in the open-endedness and multi-physics motifs. In a standard pipeline (and to a different degree in AlphaFold), this knowledge is obfuscated or not present -- the full domain knowledge must be encoded \textit{a priori}. The use of simulation intelligence motifs in addition to DP holds great potential in this area, notably with semi-mechanistic modeling to incorporate experimental and structural data into integrative multi-scale simulation methods that can model a range of length and time scales involved in drug design and development \cite{Jagger2020MultiscaleSA}.

\paragraph{JAX: DP for physical simulation from molecular dynamics to rigid body motion} 

We've highlighted JAX several times now as a valuable system for utilizing machine learning to improve simulation sciences. One use-case demonstrating integration of several motifs is the example of building semi-mechanistic surrogate models in the DP framework in order to accelerate complex computational fluid dynamics (CFD) simulations \cite{Kochkov2021MachineLC}. Another use-case we alluded to in Fig, \ref{fig:brainbloodflow} is spanning multiple modeling scales including molecular dynamics (MD). 
MD is frequently used to simulate materials to observe how small scale interactions can give rise to complex large-scale phenomenology \cite{jax2018github}. Standard MD packages used across physics simulations (such as HOOMD Blue \cite{Anderson2013HOOMDblueAP} or LAMMPS \cite{Plimpton1993FastPA,Thompson2021LAMMPSA}) are complicated, highly specialized pieces of code. Poor abstractions and technical debt make them difficult to use, let alone extend for advances in accelerated computing and ML -- for instance, specialized and duplicated code for running on CPU vs. GPU, hard-coded physical quantities, and complex manual differentiation code. For these reasons, JAX MD \cite{Schoenholz2020JAXMA} leverages differentiable programming as well as automatic hardware acceleration (via the compilation scheme in JAX).
Several design choices rooted in functional programming make JAX MD distinct from traditional physics software, and more capable with autodiff and ML methods: transforming arrays of data, immutable data, and first-class functions.
JAX MD primitives are functions that help define physical simulations, such as ``spaces'' (dimensions, topology, boundary conditions) and ``interactions'' (energy, forces). Higher level functions can then be used to compose simulations of particles or atoms.

The initial JAX MD work \cite{Schoenholz2020JAXMA} demonstrated training several state-of-art neural networks to model quantum mechanical energies and forces, specifically to represent silicon atoms in different crystalline phases based on density functional theory (DFT, the standard first principles approach to electronic structure problems).
DFT is prevalent across physics problems but non-trivial to reliably compute or approximate in practice. In recent years there's been interest in utilizing ML to improve DFT approximations, notably towards JAX-DFT \cite{Li2021KohnShamEA}. They leverage the differentiable programming utilities of JAX to integrate the DFT solver equations (i.e., Kohn-Sham equations) directly into neural network training -- that is, leveraging JAX as a physics-informed ML framework. This integration of prior knowledge and machine-learned physics, the authors suggest, implies rethinking computational physics in this new paradigm: in principle, all heuristics in DFT calculations (e.g., initial guess, density update, preconditioning, basis sets) could be learned and optimized while maintaining rigorous physics \cite{Li2021KohnShamEA}. 
JAX has similarly been leveraged for automated construction of equivariant layers in deep learning \cite{Finzi2021APM}, for operating on graphs and point clouds in domains like particle physics and dynamical systems where symmetries and equivariance are fundamental to the generalization of neural networks -- \textit{equivariant neural networks} are detailed later in the Discussion section.
A more AI-motivated extension of JAX is Brax \cite{Freeman2021BraxA}, a differentiable physics engine for accelerated rigid body simulation and reinforcement learning. Brax trains locomotion and dexterous manipulation policies with high efficiency by making extensive use of auto-vectorization, device parallelism, just-in-time compilation, and auto-differentiation primitives of JAX. 
For similar reasons, JAX has been leveraged specifically for \textit{normalizing flows}, a general mechanism for defining expressive probability distributions by transforming a simple initial density into a more complex one by applying a sequence of invertible transformations \cite{rezende2015variational,Papamakarios2019NormalizingFF}. We touch on normalizing flows more in the Discussion section later.

\paragraph{DiffTaichi: purpose-built DP for highly-efficient physical simulation} 
Taichi \cite{Hu2019TaichiAL} is a performance oriented programming language for writing simulations, and DiffTaichi \cite{Hu2020DiffTaichiDP} augments it with additional automatic differentiation capabilities. While JAX MD and the various others in the JAX ecosystem can provide physical simulation workflows out of the box because they're purpose built for those applications, DiffTaichi is ``closer to the metal'' for fast simulations and thus in general better computational performance over JAX. This can be explained by several key language features required by physical simulation, yet often missing in existing differentiable programming tools: 
\begin{itemize}
    \item \textit{Megakernels --} The programmer can naturally fuse multiple stages of computation into a single kernel, which is later differentiated using source code transformations and just-in-time compilation. Compared to mainstream DP frameworks for deep learning (PyTorch \cite{Paszke2017AutomaticDI} and TensorFlow \cite{Abadi2016TensorFlowAS}) which use linear algebra operators, the megakernels have higher arithmetic intensity and are thus more efficient for physical simulation tasks.
    \item \textit{Imperative parallel programming --} DiffTaichi is built in an imperative way for better interfacing with physical simulations than the functional array programming approach of mainstream deep learning frameworks. First, this approach helps provide parallel loops, ``if'' statements and other control flows in a differentiable framework, which are widely used constructs in physical simulations for common tasks such as handling collisions, evaluating boundary conditions, and building iterative solvers. Second, traditional physical simulation programs are written in imperative languages such as Fortran and C++.
    \item \textit{Flexible indexing --} The programmer can manipulate array elements via arbitrary indexing, handy for the non element-wise operations such as numerical stencils and particle-grid interactions that are specific to physical simulations and not deep learning.
\end{itemize}

The initial work by Hu et al. \cite{Hu2020DiffTaichiDP} explains the design decisions and implementations nicely, along with illustrative examples in incompressible fluids, elastic objects, and rigid bodies with significant performance gains over comparable implementations in other DP (and deep learning) frameworks. Nonetheless these are on toy problems and much work remains to be done in complex real-world settings.

\paragraph{Julia \textbf{$\delta P$}: differentiable programming as a scientific ML \textit{lingua franca}}
Dynamic programming languages such as Python, R, and Julia are common in both scientific computing and AI/ML for their implementation efficiency, enabling fast code development and experimentation. 
Yet there's a compromise reflected in most systems between convenience and performance: programmers express high-level logic in a dynamic language while the heavy lifting is done in C/C++ and Fortran. 
Julia \cite{bezanson2017julia}, on the other hand, may give the performance of a statically compiled language while providing interactive, dynamic behavior. The key ingredients of performance are: rich type information, provided to the compiler naturally by multiple dispatch, code specialization against run-time types, and just-in-time (JIT) compilation using the LLVM compiler framework \cite{Lattner2004LLVMAC} -- see Bezanson et al. \cite{bezanson2017julia} for details.

Potentially resolving the tradeoffs and annoyances of dynamic-static systems that couple Python and C, for example, Julia is a natural fit for differentiable programming towards scientific ML (not to mention probabilistic programming, geometric deep learning, and others we've touched on).
The work by Innes et al. \cite{Innes2019ADP} is a DP system that is able to take gradients of Julia programs, making automatic differentiation a first class language feature. Thus Julia can support gradient-based learning and program transforms for the SI motifs and deep learning.
We detailed several examples in the multi-physics and surrogate modeling motifs, in particular the Universal Differential Equations (UDE) system \cite{Rackauckas2020GeneralizedPL} (Fig. \ref{fig:ude}). Other illustrative examples include differentiable programming for finance, where contract valuation can be mixed with market simulations and other DP tools like neural networks; accelerated neural differential equations in financial time series modeling; and end-to-end automatic differentiation of hybrid classical-quantum systems (in simulation) \cite{Innes2019ADP}.

\subsection*{Future directions}\label{sec:DP_future}

In the artificial intelligence field, differentiable programming has been likened to deep learning \cite{lecunDP}. This does not do it justice, as DP can be viewed as a generalization of deep learning. The potential of DP is immense when taken beyond deep learning to the general case of complex programs -- in existing programs taking advantage of the extensive amount of knowledge embedded within them, and enabling never before possible models and \textit{in silico} experiments. 
DP has the potential to be the lingua franca that can further unite the worlds of scientific computing and machine learning \cite{Rackauckas2020GeneralizedPL}. The former is largely governed by mechanistic equations and inefficient numerical solvers, and the latter with data-driven methods that learn high-dimensional spaces. DP provides exciting opportunity at the intersection, as we explored in the multi-physics motif. 

DP tied to ML and simulation presents a relatively new area and there is much to do in both theory and practice. For the former, it would be valuable to derive techniques for working with non-differentiable functions like ReLU used in deep learning \cite{Abadi2020ASD} or with non-smooth functions. It would also be useful to work with approximate reals, rather than reals, and to seek numerical accuracy theorems. For practical use, particularly in ML, it is important for DP to add explicit tensor (multi-dimensional array) types with accompanying shape analysis. In general, richer DP languages are needed to support wide ranges of types and computations -- e.g. recursion, or programming with Riemannian manifolds (to accommodate natural gradient descent). There is also a need to move beyond Python as a host language, with computation runtimes magnitudes more efficient in Julia, C++ (e.g. DiffTaichi), and close-to-the-metal domain-specific languages (DSLs, see Fig. \ref{fig:os}).

Just as AD is the practical power underlying DP, non-standard interpretation is the fundamental power underlying AD. Non-standard interpretation poses the question \textit{how can I change the semantics of this program to solve another problem?}, the theory and formalism of which could provide rich developments beyond AD.

\hfill \break

\section{\textit{The Frontier}}

There are two nascent fields of AI and ML
that we suggest have potentially massive upside for advancing simulation sciences and machine intelligence: open-endedness and machine programming. We now describe these as motifs in the SI stack.

\subsection{8. OPEN-ENDED OPTIMIZATION}
\textit{Open-endedness}, or open-ended search, can be defined as any algorithm that can consistently produce novelty or increases in complexity in its outputs. Such open-endedness is a ubiquitous feature of biological, techno-social, cultural, and many other complex systems \cite{Banzhaf2016DefiningAS}. But its full algorithmic origins remain an open problem \cite{Stepney2021}. This open-endedness stands in contrast to most other kinds of optimization which generally seek to minimize loss functions, or otherwise reduce uncertainty \cite{Friston2017,Krakauer2020a}. By instead seeking to increase complexity, open-ended search tends to increase uncertainty by definition \cite{Rasmussen2019TwoMO}. We believe this is an extremely desirable property in optimization (and intelligence) as it has proven to be a source of robust behavior, and mechanistic innovation \cite{Adams2017} as we will review below. 

\textit{Open-ended learning} algorithms are in general those with characteristics of life-long learning: adaptively and continually pose new objectives, adapt to environment changes, and avoid pitfalls such as local minima, deceptive loss functions \cite{Lehman2011a}, or catastrophic forgetting \cite{Kumaran2016WhatLS}. It follows that \textit{open-ended optimization} is where components of a system continue to evolve new forms or designs continuously, rather than grinding to a halt when some sort of optimal, iteration threshold, or stable position is reached \cite{Taylor1999FromAE}. 
Open-ended optimization also has deep connections to fundamental issues in computer science. In theoretical analysis it has been shown to produce unbounded complexity, via computational universality \cite{Hernandez2018} and to deal with learning strategies in non-differentiable spaces \cite{HernndezOrozco2020AlgorithmicPM} contrary to the old belief that differentiability was fundamental. By these arguments we must embrace computation undecidability and irreducibility in optimization methods \cite{Hernandez2018}, which in turn can lead to practical results, such as candidate explanations for the emergence of modularity in nature based on a simple storage system \cite{Hernandez2018b}.

In general at least two major kinds of open-endedness can be distinguished: (1) processes that do not stop, and (2) processes that do not have specific objectives. It is straightforward to draw the connection to these features and to evolution. 
We believe open-ended algorithms, like those we will outline below, \emph{must} serve as the mechanisms underlying evolution's creativity, at multiple levels of abstraction, from the evolution of protein sub-units, to protein networks, to the development eukaryotic systems, to speciation events, and perhaps the Cambrian explosion \cite{Hernandez2018b}. Rasmussen \& Sibani \cite{Rasmussen2019TwoMO} provide useful discussion to this end, by providing a means for disentangling the processes and algorithms of optimization and expansion within complex evolutionary processes.
We aim to extend these concepts broadly in life sciences and other domains such as physics and engineering, as alluded to in the methods and examples discussed next.

\subsection*{Examples}\label{sec:OE_ex}

\paragraph{Randomness and optimization}
Randomness is, in a sense, the simplest kind of open-ended search: A random process does indefinitely continue, and can generate different values which for a time leads to increases in entropy, and therefore output complexity. But of course for a finite tailed random process, entropy of the samples converges, and so complexity converges. Nevertheless, the right amount of randomness (or noise) can bump an optimization process out of local minima, and so aid in improving the optimization \cite{Burke2005}. 
Randomness offers a simple demonstration of how increasing complexity can later on aid an optimization whose objective is decreasing uncertainty. But, as the measured entropy of a random process does plateau, it is insufficient for the consistent introduction of complexity over the lifetime \cite{Thrun1998}.

Some Bayesian optimization (BO) methods are examples of entropy-based search strategies, where acquisition functions intelligently balance explore-exploit search strategies \cite{Bonawitz2014} -- recall an acquisition function guides the BO search by estimating the utility of evaluating the objective function at a given point or parameterization. A useful class of acquisition functions are information-based, such entropy search \cite{Hennig2012EntropySF,Haarnoja2018} and predictive entropy search \cite{HernndezLobato2014PredictiveES}. Yet these information-based methods aim to reduce uncertainty in the location of the optimizer, rather than increasing entropy continuously; BO is a key example of ML-based optimization that is orthogonal to open-endedness. Some BO strategies, and broadly Monte Carlo sampling, utilize search algorithms based on randomness. Rather, these are pseudo-random algorithms, such as Sobol sequences \cite{Sobol1967OnTD} and other quasi-Monte Carlo techniques \cite{Lemieux2009MonteCA,Letham2019ConstrainedBO,Lavin2018DoublyBO}.

\paragraph{Deceptive objectives and novelty search}
Deceptive optimization problems, are defined as problems for which the global optima appears highly suboptimal over the first few samples, and generally not solved by simple noise injection \cite{Lehman2011}. A concrete example of deception can be found by considering an agent in a maze. If the environment provides a proximity signal to the end of the maze, it would seem such a signal would make the maze trivial. But this is often not so. Consider that in most mazes an agent must walk away from the objective, for a time (to get around a bend, for example). In this case the proximity signal becomes a deceptive signal because while it is the optimal policy to follow the proximity signal overall, for short periods the winning agent must violate this optimal policy.

Stanley and colleagues \cite{Lehman2011a} were the first to suggest that \textit{open-ended novelty search}, defined as open-ended search to increase behavioral complexity, is sufficient to overcome deception. This is because the deceptive states are novel states. Other researchers have gone so far to suggest that novelty and curiosity (discussed below) are sufficiently good search/exploration algorithms that they can and should be used universally, in place of traditional objective-based optimization methods \cite{Peterson2019,Fister2019,Mouret2011b}. While novelty search does converge, there is early evidence novelty search interacts with traditional evolutionary algorithms to make evolvability--modularization and reusability--inevitable \cite{Doncieux2020}.

To increase searching efficiency in open-endedness, some efforts combine novelty search with \textit{interactive evolution (IEC)}, where a human intervenes on the process to guide search \cite{Takagi2001InteractiveEC,Woolley2014ANH,Lwe2016AcceleratingTE}. IEC is particularly well suited for domains or tasks for which it is hard to explicitly define a metric, including when criteria are inherently subjective or ill-defined (e.g., salience, complexity, naturalism, beauty, and of course open-endedness). 

Woolley \& Stanley \cite{Woolley2014ANH} implement novelty-assisted interactive evolutionary computation (NA-IEC) as a human-computer interaction system, where a human user is asked to select individuals from a population of candidate behaviors and then apply one of three evolutionary operations: a traditional IEC step (i.e. subjective direction by the human), a short term novelty search, or a fitness-based optimization. Experiments in a deceptive maze domain show synergistic effects of augmenting a human-directed search with novelty search: NA-IEC exploration found solutions in fewer evaluations, at lower complexities, and in significantly less time overall than the fully-automated processes.

\paragraph{Curiosity and self-training artificial systems}
Developmental artificial intelligence, defined as the invention and use of training algorithms which force agents to be self-teaching, relies on giving agents a sense of curiosity. Curiosity, loosely defined as learning for learning's sake \cite{Berlyne1954}, has been difficult to pin down experimentally both in psychological research \cite{Kidd2015,Gottlieb2018} and in AI research \cite{Savinov2019,Burda2018,Ecoffet2019,Kosoy2020,Colas2019,Oudeyer2018a}. That said, there is recent progress toward a unified mathematical view \cite{Peterson2019}. Details aside, the broad goal of \textit{self-teaching-by-curiosity} is to lessen, or remove, the need for carefully curated datasets which are both expensive to create and often innately and unfairly biased \cite{Gebru2017,Jo2020}. 

According to leading practitioners, effective self-teaching \cite{Sigaud2021} requires an agent go about self-generating goals \cite{Colas2019}, discovering controllable features in the environment \cite{Laversanne-Finot2018}, and as a result make causal discoveries \cite{Gopnik2012,Sontakke2020}. It also requires agents to learn to avoid catastrophic mistakes \cite{Lipton2018}, and handle environmental non-stationary \cite{Thrun1998}. Self-learning inverse models may also prove important \cite{Baranes2013}; in other words, self-teaching agents must generate self-organizing long-term developmental trajectories. 

\paragraph{Quality-diversity and robustness to unknown in robotic systems}
The broad goal of \textit{quality-diversity (QD)} algorithms is to produce a diverse set of high performing solutions to any given optimization problem \cite{Pugh2016a}. Having diverse potential solutions for AI-driven research and decision making is desirable as it enables the practitioner or downstream algorithms to be robust in problem solving \cite{Pugh2016a,Colas2020}. In practice, this is done by combining novelty search (or other open-ended search approaches) and objective-based optimization. One of the most widely used algorithms is known as MAP-elites, developed by Mouret \& Clune \cite{Mouret2015}. 
Clune and colleagues use MAP-elites to generate generalizable action plans in robotic systems \cite{Velez2014}. Remarkably though, QD also seems to solve a long-standing problem in robotics research: adaption to the unexpected \cite{Cully2015}. Unlike agents in the natural world, robotic systems which work perfectly in trained environments, will fail frequently and catastrophically when even small but unexpected changes occur. They especially struggle when there are changes or damage to their appendages. However, robots trained using QD methods showed a remarkable ability to recover from the unknown, as well as to purposeful damage inflicted by the experimenters. In other words, seeking novelty, prioritizing curiosity, and building complexity, prepares agents to handle the unexpected \cite{Jaderberg2021}; open-ended learning and optimization can produce more robust AI agents.

\paragraph{Robotic systems for scientific discovery}
In addition to creating robust, flexible, and self-training robotics platforms, open-ended optimization methods have significant future potential for use in directly exploring scientific unknowns. For instance, Häse and colleagues have produced several objective-based optimization schemes for autonomous chemical experimentation, including approaches that combine objectives with user-defined preferences (with similar motivations as NA-IEC) \cite{Hse2018ChimeraEH}, and in BO schemes mentioned above \cite{Hse2018PhoenicsAB}. These are developed in the context of ``Materials Acceleration Platforms'' for autonomous experimentation in chemistry \cite{FloresLeonar2020MaterialsAP}, where the specific implementations include he ``Chemputer'' \cite{Steiner2019OrganicSI} for organic synthesis (an automated retrosynthesis module is coupled with a synthesis robot and reaction optimization algorithm) and ``Ada'' \cite{MacLeod2020SelfdrivingLF} (Autonomous discovery accelerator) for thin-film materials. We continue discussing open-ended SI systems of this nature in the Integrations section later.

\paragraph{Lifelong machine learning and efficient memory}
A \textit{lifelong learning} system is defined as an adaptive algorithm capable of learning from a continuous stream of information, with such information becoming progressively available over time and where the number of tasks to be learned are not predefined. In general the main challenge is \textit{catastrophic forgetting}, where models tend to forget existing knowledge when learning from new observations \cite{Thrun1998}. This has been an active area of research in the neural network community (and connectionist models broadly, such as Hopfield Networks \cite{Hopfield1982NeuralNA}) for decades, and most recently focused on deep learning. For a comprehensive review refer to Parisi et al. \cite{Parisi2019ContinualLL}. Here, in the context of open-endedness, we focus on the interplay of resource limitations and lifelong (continual) learning. 

Memory-based methods write experience to memory to avoid forgetting \cite{Gepperth2016IncrementalLA}, yet run into obvious issues with fixed storage capacity. Further, it is non-trivial to predefine appropriate storage without strong assumptions on the data and tasks, which by definition will evolve in continual learning settings.
The open-endedness setting would exacerbate this issue, as increasing novelty over time would increasingly require writing new knowledge to memory.
Central to an effective memory model is the efficient updating of memory. 
Methods with learned read and write operations, such as the differentiable neural computer (DNC) \cite{Graves2016HybridCU}, use end-to-end gradient-based learning to simultaneously train separate neural networks to encode observations, read from the memory, and write to the memory. For continual learning problem domains, a DNC could hypothetically learn how to select, encode, and compress knowledge for efficient storage and retrieve it for maximum recall and forward transfer. Design principles for DNC and related external NN memories \cite{Fortunato2019GeneralizationOR,Banino2020MEMOAD,Munkhdalai2019MetalearnedNM} are not yet fully understood; DNC is extremely difficult to train in stationary datasets, let alone in open-ended non-stationary environments\cite{Hadsell2020EmbracingCC}. 

Associative memory architectures can provide insight into how to design efficient memory structures, in particular using overlapping representations to be space efficient. For example, the Hopfield Network \cite{Hopfield1982NeuralNA,Ramsauer2021HopfieldNI} pioneered the idea of storing patterns in low-energy states in a dynamic system, and Kanerva's sparse distributed memory (SDR) model \cite{Kanerva1988SparseDM}, which affords fast reads and writes and dissociates capacity from the dimensionality of input by introducing addressing into a distributed memory store whose size is independent of the dimension of the data.
The influence of SDR on modern machine learning has led to the Kanerva Machine \cite{Wu2018TheKM,Marblestone2020ProductKM} that replaces NN-memory slot updates with Bayesian inference, and is naturally compressive and generative. 
By implementing memory as a generative model, the Kanerva Machine can remove assumptions of uniform data distributions, and can retrieve unseen patterns from the memory through sampling, both of which behoove use in open-ended environments.
SDR, initially a mathematical theory of human long-term memory, has additionally influenced architectures derived directly from neocortex studies, namely Hierarchical Temporal Memory (HTM) \cite{Hawkins-et-al-2016-Book}.
HTM is in a class of generative memory methods that is more like human and rodent intelligence than the others we've mentioned, and implements open-ended learning in sparse memory architecture. HTM networks learn time-based sequences in a continuous online fashion using realistic neuron models that incorporate non-linear active dendrites. HTM relies on SDR as an encoding scheme that allows simultaneous representation of distinct items with little interference, while still maintaining a large representational capacity, which has been documented in auditory, visual and somatosensory cortical areas \cite{Cui2017TheHS}.
Leveraging the mathematical properties of sparsity appears to be a promising direction for continual learning models, and open-ended machine intelligence overall.

\subsection*{Future directions}\label{sec:OE_future}
Curiosity, novelty, QD, and even memory augmentation algorithms are in practice quite limited in the kinds of complexity they generate, and all converge rather quickly such that complexity generation will plateau and then end. This is in striking contrast to even the simplest of biological systems (for example, E. Coli monocultures \cite{Rasmussen2019TwoMO}). And the approaches for working, short term, and long term memories remain limited compared to the observed characteristics of their biological counterparts.
We seem to be missing some central insight(s); this is a ripe area for extending the domain of artificial intelligence, where complexity sciences will have significant roles to play \cite{Wolpert2019TheEO,nmi,fest}.

In the field of \textit{artificial life (ALife)}, open-endedness is a crucial concept. ALife is the study of artificial systems which exhibit the general characteristics of natural living systems. Over decades of research practitioners have developed several kinds of Alife systems, through synthesis or simulation using computational (software), robotic (hardware), and/or physicochemical (wetware) means. A central unmet goal of ALife research has been to uncover the algorithm(s) needed to consistently generate complexity, and novelty, consistent with natural evolution \cite{Frans2021}. Discovering and developing these algorithms would have significant implications in open-endedness and artificial intelligence broadly \cite{Stanley2019WhyOM}.

In the context of simulation more generally, a fully implemented open-ended system would mean not only solving optimization problems as posed by practitioners, but also posing new questions or approaches for practitioners to themselves consider. We believe this would be revolutionary. It would allow AI systems to leverage their prior expertise to create new problem formulations, or insights, and eventually, scientific theories. As important as investing in R\&D to build such systems, we need to invest in understanding the ethical implications of this approach to science, and in monitoring the real-world, downstream effects.

\begin{figure}[ht]
\centering
\includegraphics[width=0.95\linewidth]{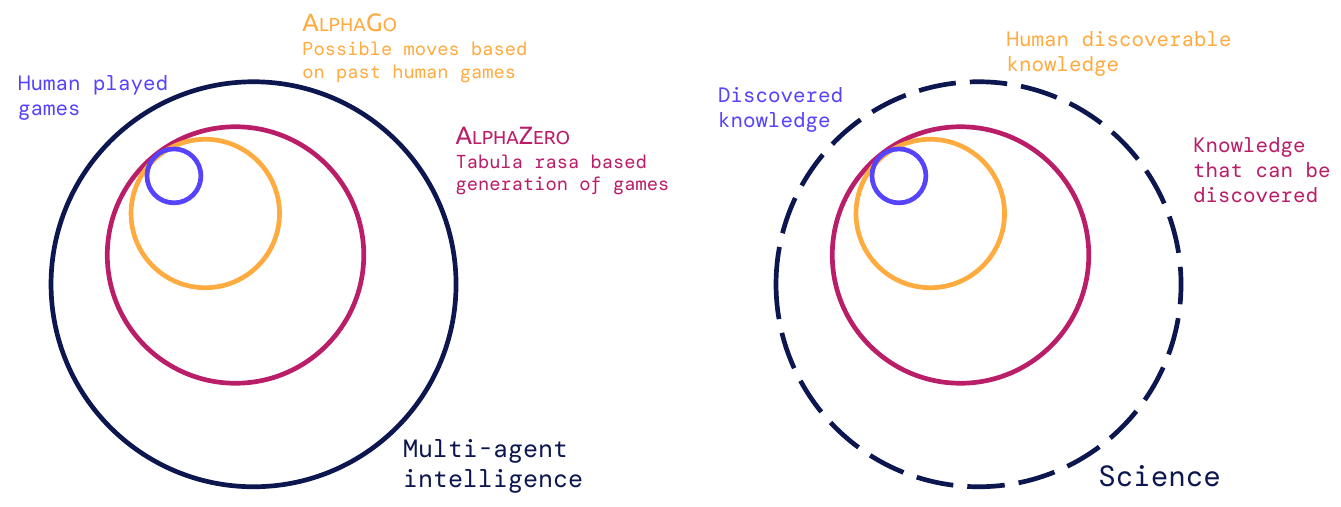}
\caption{
Left: Overall search space for multi-agent intelligence (black) -- more precisely, perfect information games -- and the subsets achieved by humans (purple), AlphaGo \cite{Silver2016MasteringTG} (yellow), and AlphaZero \cite{Schrittwieser2020MasteringAG} (red).
Right: Open-ended search space for science (black), and subsets representing what has been achieved thus far (purple), what is achievable by humans (yellow), and what is achievable in general (red).
(Diagrams are not to scale, inspired by \cite{Kitano2021NobelTC}.)
}
\label{fig:knowledge}
\end{figure}

\hfill \break
\subsection{9. MACHINE PROGRAMMING}

\textit{Machine programming (MP)} is the automation of software (and potentially hardware) development.  Machine programming is a general technology, in ways that are quite different than the other simulation intelligence motifs. As systems based on the simulation intelligence motifs grow in complexity, at some point these systems will necessitate automating aspects of software development. Machine programming can understand code across motifs, compose them into working systems, and then optimize these systems to execute on the underlying hardware -- this interaction is illustrated in Fig. \ref{fig:os} at a high-level. Hence, while MP is not tied to any specific class of SI problems, MP will be integral as the glue that ties the other motifs together, serving as a bridge to the hardware. 

\begin{figure}[h!]
\begin{center}
\includegraphics[width=0.5\columnwidth]{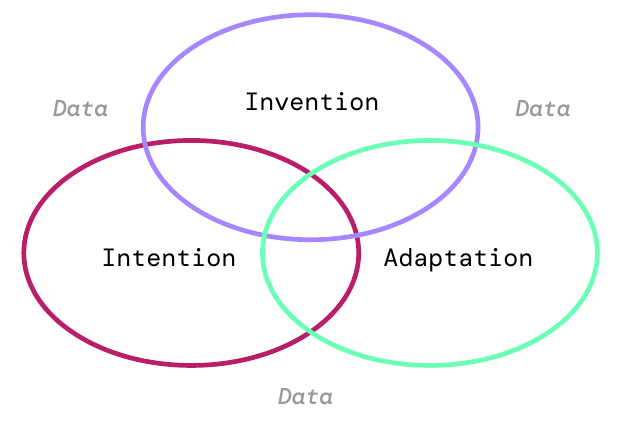}
\caption{The ``Three Pillars of Machine Programming'': intention, invention, adaptation.
Intention is quite literally to discover the intent of the programmer, and to lift meaning from software. Example sub-domains include program synthesis and inductive programming, which share territory with the invention pillar, for creating new algorithms and data structures. Sketch~\cite{nye:2019:infer}, FlashFill~\cite{Gulwani2011AutomatingSP}, and DeepCoder~\cite{Balog2017DeepCoderLT} are notable examples, with many others noted in text and in \cite{gottschlich:2018:mapl}.
The adaptation pillar is for evolving in a changing hardware/software world, and more broadly in dynamic systems of data, sensors, and people \cite{lavin2021technology} -- examples include the DSL ``GraphIt''~\cite{Zhang2018GraphItA} for computing on graphs of various sizes and structures on various underlying hardware, and ``Tiramisu''~\cite{Baghdadi2019TiramisuAP} compiler for dense and sparse deep learning and data-parallel algorithms. 
(Figure reproduced from Gottschlich et al.~\cite{gottschlich:2018:mapl})
}
\label{fig:pillars}
\end{center}
\end{figure}

The term \emph{machine programming} made its debut in a paper by Gottschlich et al.~\cite{gottschlich:2018:mapl} that defined the MP in terms of three distinct pillars: \emph{(i)} \emph{intention}, \emph{(ii)} \emph{invention}, and \emph{(iii)} \emph{adaptation} (Fig. \ref{fig:pillars}). \emph{Intention} is principally concerned with identifying novel ways and simplifying existing ways for users to express their ideas to a machine. It is also concerned with lifting meaning (i.e., semantics) from existing software and hardware systems~\cite{ahmad:2019:siggraph, kamil:2016:pldi, lee:2021:plp}. \emph{Invention} aims to construct the higher-order algorithms and data structures that are necessary to fulfill a user's intention. In many cases, inventive systems simply fuse existing components together to form a novel solution that fulfills the user's intent. However, there are some corner cases, which are likely to become more prevalent in the future, where an inventive system constructs a truly novel solution  not yet been discovered by humans (e.g., Alam et al.'s inventive performance regression testing system~\cite{alam:2019:neurips} and Li and Malik's inventive optimization algorithm system~\cite{li:2017:iclr}). \emph{Adaptation} takes an \emph{invented} high-order program and adapts it to a specific hardware and software system. This is done to ensure certain quality characteristics are maintained such as correctness, performance, security, maintainability, and so forth.

If trends continue, we anticipate programming languages will become more intentional by design. The byproduct of such intentionality is that such languages will be capable of more holistic automation of the inventive and adaptive pillars of MP, while simultaneously achieving super-human levels in key quality characteristics (e.g., software performance, security, etc.). Early evidence has emerged demonstrating this such as Halide~\cite{adams:2019:siggraph}, a programming language (more precisely a domain-specific language embedded in C++) designed for high-performance image and array processing code on heterogeneous modern hardware (e.g., various CPU and GPU architectures) and operating systems (Linux, Android, etc.). For example, Halide, in concert with verified lifting, has demonstrably improved the performance of Adobe Photoshop by a geometric mean of $3.36x$ for approximately 300 transpiled functions~\cite{ahmad:2019:siggraph}.
Moreover, other forms of intentional communication from humans to machines are already emerging, such as translation from natural language to programming language and translation of visual diagrams to software programs. Two embodiments of such systems are GitHub's Co-Pilot~\cite{Chen2021EvaluatingLL} and Ellis et al.'s automated synthesis of code for visual diagrams~\cite{ellis:2018:neurips}.

%
%

\begin{figure}[h!]
\begin{center}
\includegraphics[width=1.0\columnwidth]{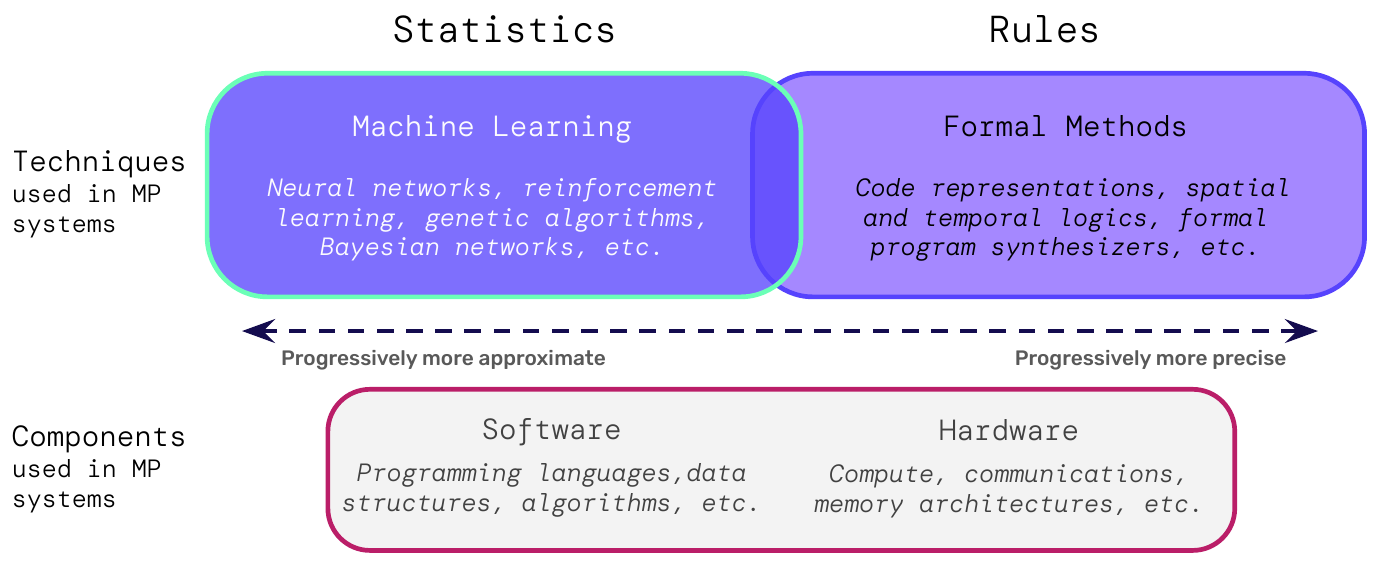}
\caption{Illustrating the ``bifurcated space'' of machine programming, made up of \emph{statistical} approaches, including various ML techniques like deep neural networks, and \emph{rules-based} approaches, which include formal methods, deterministically reproducible representations, and so forth.}
\label{fig:mpbifurcation}
\end{center}
\end{figure}

In machine programming, there appear to be at least two fundamental approaches for automation. The first is \emph{statistical}, which consists of methods from machine learning and probabilistic analysis~\cite{alam:2019:neurips, hasabnis:2021:maps}. The second is \emph{rules-based}, which consist of approaches using formal methods that include techniques such as satisfiability solvers (SAT solvers) or satisfiability modulo theory (SMT) solvers~\cite{alur:2013:fm, kamil:2016:pldi} -- SAT is concerned with the problem of determining if there exists an interpretation that satisfies a given Boolean formula, and SMT is a generalization of SAT to more complex formulas involving real numbers, integers, and/or various data structures in order determine whether a mathematical formula is satisfiable.

As the field of MP matures, we are already seeing an emergence of systems that consist of both statistical and rules-based elements. A concrete example of this is Solar-Lezama et al.'s work fusing formal sketching techniques (where a \textit{program sketch} is the schematic outline of a full program) with deep neural networks to improve computational tractability~\cite{nye:2019:infer}. Another example is the Machine Inferred Code Similarity (MISIM) system by Ye et al., an automated engine for quantifying the semantic similarity between two pieces of code, which uses a deterministic and configurable context-aware semantics structure (CASS) in conjunction with a learned similarity scoring system driven by a Siamese deep neural network with an extensible neural back-end~\cite{ye:2020:misim}. 

The Department of Energy's 2020 report on Synthesis for Scientific Computing \cite{Lopes2021PROGRAMSF} provides an overview of where and how MP can have significant impact for scientific programming. In general, the consortium of experts anticipate increases in productivity of scientific programming by orders of magnitude by making all parts of the software life cycle more efficient, including reducing the time spent tracking down software quality defects, such as those concerned with correctness, performance, security, and portability \cite{Alam2019AZL,Alam2016ProductionRunSF}.
It is also believed that MP will enable scientists, engineers, technicians, and students to produce and maintain high quality software as part of their problem-solving process without requiring specialized software development skills, which today creates several choke points in organizations of researchers and engineers in many disconnected capacities.

%

\subsection*{Future directions}\label{sec:MP_future}

One of the core directions the field of MP is heading towards is \emph{intentional programming}. Intentional programming is principally concerned with separating the intention of the program from the components that are used to realize that intention. A critical reason for this \emph{separation of concerns} is that it tends to enable the machine to more exhaustively explore the inventive and adaptive components of software development -- the programmer focuses on supplying the core ideas, while the machine handles the other two MP pillars, invention and adaptation.


A concrete embodiment of this is seen in the Halide programming language. Halide was designed with two primary components for two different types of programmers: \emph{(i)} an algorithmic DSL for domain experts and \emph{(ii)} the scheduling component for programmers with a deep understanding of building optimal software for a given hardware target. Due to the separation of the algorithm and schedule, the Halide team was able to automate the scheduling component to automatically extract performance. This is currently done using the Halide auto-scheduler using a learned cost model, which can synthesize more performant software (in the form of code, not neural networks) than the best humans it was pit against, which includes those individuals who developed the language itself~\cite{adams:2019:siggraph}. 


We anticipate the field of MP to help accelerate the development and deployment of the SI stack. Referring back to the SI operating system conceptualization in Fig. \ref{fig:os}, without MP one can imagine bottom-up constraints on the types of data structures and algorithms that can be implemented above the hardware layers
This is not unlike a ``hardware lottery'', where a research idea or technology succeeds and trumps others because it is suited to the available software and hardware, not necessarily because it is superior to alternative directions \cite{Hooker2021TheHL}.
Historically this has been decisive in the paths pursued by the largely disconnected hardware, systems, and algorithms communities; the past decade provides an industry-shaping example with the marriage of deep learning and GPUs. The various mechanisms of MP, however, provide opportunity to mitigate such hardware lottery effects for SI, and broadly in the AI and software fields. With MP in the SI operating system, as a motif that interfaces hardware and software, the integration of SI motifs can become far more efficient (auto HW-SW optimizations and codesign), robust (less prone to human and stochastic errors), and scalable (flexibly deployable modules from HPC to the edge).

\hfill \break
 
\section{Simulation Intelligence Themes}\label{sec_themes}

We first discuss several common themes throughout the SI motifs, namely the utility of the motifs for solving inverse problems, the advantages (and scientific necessities) of implementing uncertainty-aware motif methods, synergies of human-machine teaming, and additional value to be realized from integrating motifs in new ways.
Each of these themes may prove significant in the development of SI-based scientific methods.
The discussion then continues to cover important elements for SI in practice, critically the data engineering and high-performance computing elements of SI at scale.






\subsection{INVERSE-PROBLEM SOLVING}\label{sec_inverse}

\begin{figure}[ht]
\centering
\includegraphics[width=0.6\linewidth]{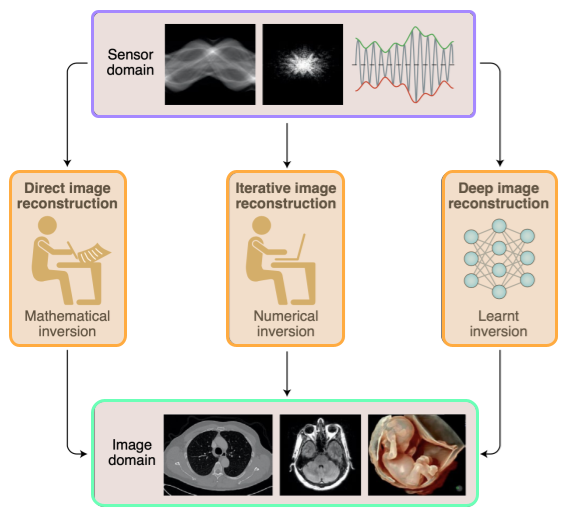}
\caption{
Three types of tomographic reconstruction methods as inverse problem-solving (orange). With direct or analytic reconstruction, a mathematical inverse of the forward transform is used to compute images from sensor data. In iterative reconstruction, the current image estimate is iteratively updated such that its forward transform gradually approaches the sensor data. In deep tomographic reconstruction, the inversion does not need to rely on any explicit transform model but is learnt from representative big data. 
(Original figure credit: Molly Freimuth (CT image); Christopher Hardy (mR image); Dawn Fessett (ultrasound image) \cite{wang2020deep}).  Examples of inverse problem-solving in other areas such as cell and molecular biology can be found in\cite{iscience}.
}
\label{fig:ct}
\end{figure}

An \textit{inverse problem} is one of inferring hidden states or parameters from observations or measurements. If the forward simulation (or data-generation) process is $x \rightarrow y$, we seek to infer $x$ given $y$. In scientific settings, an inverse problem is the process of calculating from observations the causal factors that produced them. One typical example is computational image reconstruction, including MRI reconstruction, CT tomographic reconstruction, etc., as shown in Fig.  \ref{fig:ct}. Most reconstruction methods could be classified as direct, analytical, or iterative reconstruction \cite{wang2020deep}. Based on the knowledge and information, a forward model can be derived to predict data given an underlying objective, either a simulation model or a mathematical/physical model\cite{iscience}. The reconstruction is addressed by computing the mathematical inverse of the forward model via nonlinear programming and optimization with regularization. Recently, deep learning based approaches provide a more promising way to inverse problem solving when conventional mathematical inversion is challenging: Rather than relying entirely on an accurate physics or mathematical model, the new data-driven machine learning methods can leverage large datasets to learn the inverse mapping end-to-end. 

\begin{figure}[ht]
\centering
\includegraphics[width=0.9\linewidth]{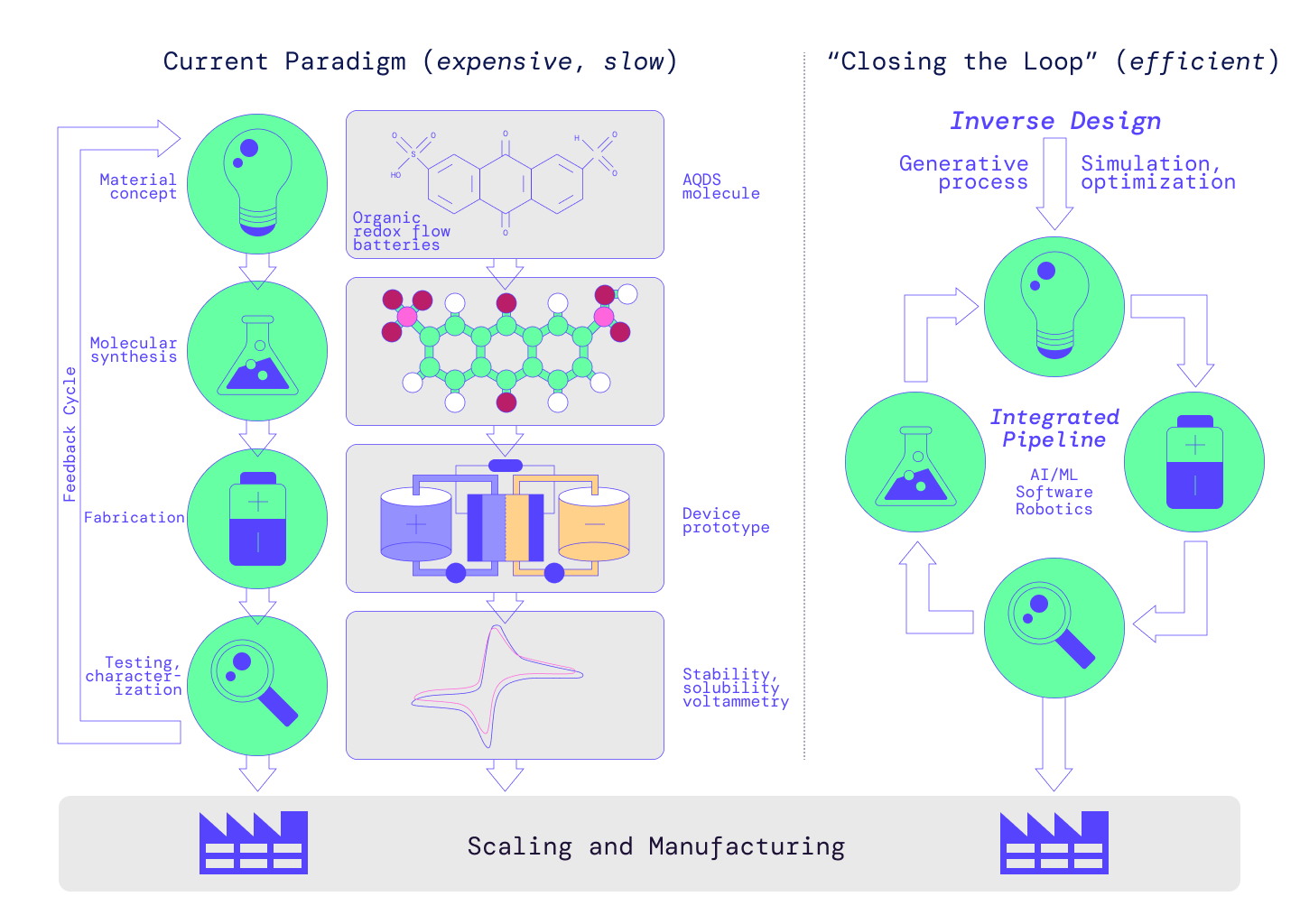}
\caption{
In many domains we can realize efficient inverse design brought about by simulation and AI technologies. This example schematic shows the current material discovery paradigm (here for organic redox flow batteries \cite{SnchezLengeling2018InverseMD}) versus a closed-loop, inverse design counterpart \textit{in silico} that maximizes efficiency and minimizes \textit{in situ} resource-utilization.
}
\label{fig:inverse-batteries}
\end{figure}

Beyond the classic inverse problems, such as image reconstruction, inverse scattering, and computed tomography, much recent effort has been dedicated to the design of new materials and devices, and to the discovery of novel molecular, crystal, and protein structures via an inverse process. This can be described as a problem of \textit{inverse design:} discovering the optimal structures or parameters of a design based on the desired functional characteristics -- for example, computational inverse design approaches are finding many use-cases in optimizing nanophotonics \cite{Molesky2018InverseDI,So2020DeepLE, zhang2021directional}, crystal structures \cite{corpinot2018practical,noh2019inverse,noh2020machine,fung2021inverse} and molecular biology\cite{iscience}. Fig. \ref{fig:inverse-batteries} illustrates an example of how inverse design cycles can replace the existing, deductive paradigm that can be expensive and inefficient, and yield suboptimal solutions. In vast, complex spaces like materials science, inverse design can yield de novo molecules, foreshadowing what could be an SI-driven, inductive scientific method.

Inverse design differing from conventional inverse problems poses several unique challenges in scientific exploration. First, the forward operator is not explicitly known. In many cases, the forward operator is modeled by first-principles calculations, including molecular dynamics and density functional theory (DFT). This challenge makes inverse design intractable to leverage recent advances in solving inverse problems, such as MRI reconstruction \cite{wang2020deep}, implanting on images through generative models with large datasets \cite{asim2020invertible}. Second, the search space is often intractably large. For example, small drug-like molecules have been estimated to contain between $10^{23}$ to $10^{60}$ unique cases \cite{ertl2009estimation}, while solid materials have an even larger space. This challenge results in obvious obstacles for using global search, such as Bayesian optimization \cite{snoek2015scalable}, evolutionary strategies \cite{hansen2003reducing, salimans2017evolution, zhangenabling}, or Bayesian inference via Markov Chain Monte Carlo (MCMC), since either method is prohibitively slow for high-dimensional inverse problems. Third, we encounter multi-modal solutions with one-to-many mapping. In other words, there are multiple solutions that match the desirable property. This issue leads to difficulties in pursuing all possible solutions through gradient-based optimization, which converges to a single deterministic solution and is easy trapped into local minima. Probabilistic inference also has limitations in approximating the complex posterior, which may cause the simple point estimates (e.g., maximum a posterior (MAP)) to have misleading solutions \cite{sun2020deep}. 
Generally speaking, inverse design methods include Bayesian and variational approaches, deep generative models, invertible learning, and surrogate-based optimization. 

{\em Bayesian and variational approaches.}
From the inference perspective, solving inverse design problems can be achieved by estimating the full posterior distributions of the parameters conditioned on a target property. Bayesian methods, such as approximate Bayesian computing \cite{yang2018predictive}, are ideal choices to model the conditional posterior but this idea still encounters various computational challenges in high-dimensional cases. An alternative choice is variational approaches, e.g., conditional GANs \cite{wang2018high} and conditional VAE \cite{sohn2015learning}, which enable the efficient approximation of the true posterior by learning the transformation between latent variables and parameter variables. However, the direct application of both conditional generative models for inverse design is challenging because a large number of data is typically required \cite{tonolini2020variational}.

{\em Deep generative models.} 
Many recent efforts have been made on solving inverse problems via deep generative models \cite{asim2020invertible, whang2021composing, whang2021solving, daras2021intermediate, sun2020deep, kothari2021trumpets, zhangflow}. For example, Asim et al. \cite{asim2020invertible} focuses on producing a point estimate motivated by the MAP formulation and \cite{whang2021composing} aims at studying the full distributional recovery via variational inference. A follow-up study from \cite{whang2021solving} is to study image inverse problems with a normalizing flow prior which is achieved by proposing a formulation that views the solution as the maximum \textit{a posteriori} estimate of the image conditioned on the measurements. For MRI or implanting on images, strong baseline methods exist that benefit from explicit forward operator \cite{sun2020deep, asim2020invertible, kothari2021trumpets}. 

{\em Surrogate-based optimization.}
Another way is to build a neural network surrogate and then conduct surrogate-based optimization via gradient descent. This is more common in scientific and engineering applications \cite{forrester2009recent, gomez2018automatic, white2019multiscale, xue2020amortized, brookes2019conditioning}, specifically for the inverse design purpose. The core challenge is that the forward model is often time-consuming so a faster surrogate enables an intractable search. A recent study in the scope of surrogate-based optimization is the neural-adjoint (NA) method \cite{ren2020benchmarking} which directly searches the global space via gradient descent starting from random initialization, such that a large number of interactions are required to converge and its solutions are easily trapped in the local minima \cite{deng2021neural}. Although the neural-adjoint method boosts the performance by down selecting the top solutions from multiple starts, the computational cost is significantly higher, especially for high-dimensional problems \cite{khatib2021deep}. 

{\em Invertible models and learning.} 
Flow-based models \cite{rezende2015variational,dinh2016density,kingma2018glow,grathwohl2018ffjord,wu2020stochastic,nielsen2020survae}, may offer a promising direction to infer the posterior by training on invertible architectures. Some recent studies have leveraged this unique property of invertible models to address several challenges in solving inverse problems \cite{ardizzone2018analyzing,ardizzone2019guided, kruse2021benchmarking}. However, these existing inverse model approaches suffer from limitations \cite{ren2020benchmarking} in fully exploring the parameter space, leading to missed potential solutions, and fail to precisely localize the optimal solutions due to noisy solutions and inductive errors, specifically in materials design problems \cite{deng2020neural}.  To address this challenge, Zhang et al. \cite{zhanginverse} propose a novel approach to accelerate the inverse design process by leveraging probabilistic inference from deep invertible models and deterministic optimization via gradient descent.  Given a target property, the learned invertible model provides a posterior over the parameter space; they identify these posterior samples as an intelligent prior initialization which enables us to narrow down the search space. Then a gradient descent is performed to calibrate the inverse solutions within a local region. Meanwhile, a space-filling sampling \cite{shields2016generalization} is imposed on the latent space to better explore and capture all possible solutions. More discussion on flow-based models follows in the context of ``honorable mention motifs'' later in this section.

Across sciences one can consider the simulation as implicitly defining the distribution $p(X, Z|y)$, where $X$ refers to the observed data, $Z$ are unobserved latent variables that take on random values inside the simulation, and $y$ are parameters of the forward model. As we've suggested, there exist many use-cases and workflows for the inverse, where one wishes to infer $Z$ or $y$ from the observations $X = x$. The inverse design loop in Fig. \ref{fig:inverse-batteries} represents a class of this problem in a practical application. In many cases it is possible the solution to the inverse problem is ill-posed \cite{Carleo2019MachineLA}: small changes in observed outputs lead to large changes in the estimate, implying the inverse problem solver will have high variance. For this reason, and to meet the needs of scientific applications in general, we often require uncertainty quantification, specifically given limited datasets \cite{zhang2018quantification, zhang2018effect,zhang2020quantification} from high-fidelity simulations or experiments. 
Moreover, to fully assess the diversity of possible inverse solutions for a given measurement, an inverse solver should be able to estimate the complete posterior of the parameters (conditioned on an observation). This makes it possible to quantify uncertainty, reveal multi-modal distributions, and identify degenerate and unrecoverable parameters -- all highly relevant for applications in science and engineering \cite{Ardizzone2019AnalyzingIP}. 
The challenge of inverse problems is also key in the context of causal discovery in science. In\cite{nmi,iscience}, a framework connecting these challenges, requiring the parallel simulation of a large number of computable programs, based on the foundations of \textit{algorithmic probability}, is introduced as capable of producing generative models and used as the foundation of a model-driven approach to inverse problems.

\hfill \break
\subsection{UNCERTAINTY REASONING}

How to design a reliable system from unreliable components has been a guiding question in the fields of computing and intelligence \cite{Neumann1956ProbabilisticLA}. In the case of simulators and AI systems, we aim to build reliable systems with myriad unreliable components: noisy and faulty sensors, human and AI error, unknown or incomplete material properties, boundary conditions, and so on. There is thus significant value to quantifying the myriad uncertainties, propagating them throughout a system, and arriving at a notion or measure of reliability.

Many of the recommended methods for each motif are in the class of probabilistic ML, first and foremost probabilistic programming, which naturally represents and manipulates uncertainty about models and predictions \cite{Ghahramani2015ProbabilisticML}. 
Probabilistic ML encompasses Bayesian methods that use probabilities to represent \textit{aleatoric uncertainty}, measuring the noise inherent in the observations, and \textit{epistemic uncertainty}\cite{philo}, accounting for uncertainty in the model itself (i.e., capturing our ignorance about which model generated the data). 
For example, Zhu \& Zabaras~\cite{Zhu2018BayesianDC} develop a probabilistic encoder-decoder neural network in order to quantify the predictive uncertainty in fluid dynamics surrogate models, and recent extensions to physics-informed ML methods aim to implement Bayesian counterparts, such as Bayesian Neural ODE \cite{Dandekar2020BayesianNO} and Bayesian PINN \cite{Yang2021BPINNsBP} -- interestingly, some findings suggest that typical NN uncertainty estimation methods such as MC-dropout \cite{Gal2016DropoutAA} do not work well with PDEs, which is counter to other PINN uncertainty quantification work \cite{Zhang2019QuantifyingTU}, implying there is still much work to be done in this area.
Gaussian processes, on the other hand, used in the same physics-modeling and emulation scenarios quantify aleatoric and epistemic uncertainties naturally \cite{Bilionis2015BayesianUP}.


\textbf{\textit{Probabilistic numerics}} \cite{Hennig2015ProbabilisticNA, Poincare1896} is a highly relevant class of methods that provide statistical treatment of the errors and/or approximations that are made en route to the output of deterministic numerical methods, such as the  approximation of an integral by quadrature, or the discretized solution of an ordinary or partial differential equation \cite{Oates2019AMR}. The many \textit{numerical algorithms} we've discussed in this paper estimate quantities not directly computable by using the results of more readily available computations, and the probabilistic numeric viewpoint provides a principled way to manage the parameters of such numerical computations.
By comparison, epistemic uncertainty arises from the setup of the computation rather than the computation itself, and aleatory is concerned with the data.
Hennig et al. \cite{Hennig2015ProbabilisticNA} describe this class of methods in detail, along with practical uses such as Bayesian quadrature in astronomy. The more recent overview by Oates \& Sullivan \cite{Oates2019AMR} provides more context, not to mention the suggestion that probabilistic programming will be critical for  integration of probabilistic numerics theory and software, as well as enabling the tools from functional programming and category theory to be exploited in order to automatically compile codes built from probabilistic numerical methods (e.g., Ref.~\cite{Scibior2015PracticalPP}).

In nearly all experimental sciences at all scales, from particle physics to cosmology, an important use of machine learning and simulation is to generate samples of labeled training data. 
For example, in particle physics, when the target $y$ refers to a particle type, particular scattering process, or parameter appearing in the fundamental theory, it can often be specified directly in the simulation code so that the simulation directly samples $X \sim p(\cdot|y)$~\cite{Carleo2019MachineLA}. 
In some cases this may not be feasible in the simulation code, but the simulation may provide samples $(X, Z) \sim p(\cdot)$, where $Z$ are latent variables that describe what happened inside the simulation, but which are not observable in an actual experiment. Either case enables the researchers to generate accurate labeled training data. Yet while the data-generating simulator has high fidelity, the simulation itself has free parameters to be tuned and residual uncertainties in the simulation must be taken into account in downstream tasks. Comprehensively quantifying uncertainties in data-generators and simulators in general remains an area of active research -- notably it must be interdisciplinary, as the uncertainty measures for ML do not necessarily suffice for physical sciences, and vice versa.

We have implied that development of robust AI and simulation systems require careful consideration of dynamic combinations of data, software, hardware, \textit{and} people \cite{lavin2021technology}. AI and simulation systems in sciences and decision-making often construct human-machine teams and other variations of cooperative AI \cite{Dafoe2020OpenPI}, where it is valuable to calculate the various uncertainties in communications between combinations of humans, machines, agents, and sources of data and information broadly. Uncertainty-aware cooperative AI is a relatively new area of research, which we touch on more in the human-machine teaming section.

An additional direction for our work on uncertainty reasoning is to investigate how the motifs provide more powerful uncertainty methods than currently used in probabilistic ML, for instance the use of differentiable programming (more specifically autodiff) towards automatic uncertainty quantification. Models built in probabilistic programming frameworks are particularly useful because not only is uncertainty quantification trivial, the modeler is able to encode their expected uncertainties in the form of priors. Modern neural networks, on the other hand, are often miscalibrated (in the sense that their predictions are typically overconfident) due to ignoring epistemic uncertainty: NNs are typically underspecified by the data and thus can represent many different models that are consistent with the observations. Uncertainty estimation methods for NNs exist, although some research has demonstrated difficulties in translating these methods to physics-informed ML use-cases \cite{Yang2021BPINNsBP}.
One promising direction is uncertainty-driven open-ended search. We have described the future benefits in optimization settings, and also suggest these same methods may be key in the hypothesis-generation part of automating the scientific method -- Kitano \cite{Kitano2021NobelTC} describes these pursuits -- and helping shape new scientific methods (for instance, built on principles of uncertainty and complexity, rather than falsifiability).

\hfill \break
\subsection{INTEGRATIONS}

Throughout this paper we've suggested there’s significant and valuable overlap between the module motifs. For instance, the use of multi-scale and agent-based modeling in applications such as systems biology and sociopolitics, 
the use of physics-infused ML for building surrogate models in multi-physics scenarios, using simulation-based causal discovery in agent-based network models, and more.
We've also discussed some of the many ways that engine motifs integrate with and enable the module motifs. For instance, probabilistic programming languages (PPL) provide useful representations for multi-modal multi-scale simulation \cite{Pfeffer2009FigaroA}, agent-based modeling \cite{agentmodels}, semi-mechanistic modeling \cite{Lavin2018DoublyBO}, and causal inference \cite{Perov2019MultiVerseCR}. Another type of integration would be using probabilistic programs within the broader workflow of a module motif: e.g., within simulation-based inference (not to mention ``human-machine inference'' broadly) as shown in Fig.~\ref{fig:sbi}, or in scenarios for open-endedness.
Same goes for the other engine motif, differentiable programming: Semi-mechanistic modeling, which is critical in multi-physics simulation amongst other scenarios, is only useful because of differentiable programming. And as we explained earlier, differentiable programming is a generalization of deep learning, and these neural network methods play significant roles in many motifs and workflows, e.g. multi-physics and inverse design, respectively.
Not to mention, at the engine level, consider that a common approach for implementing PPL is embedding stochastic macros within existing languages such as Python or C++, and utilizing automatic differentiation features from differentiable programming.
And there are valuable implications for leveraging both in the development of machine programming and open-endedness: for instance, auto-differentiating software to better enable learning and optimization in intentional programming, and utilizing PPL for readily encoding and tuning priors into probabilistic generative models in open-ended environments, respectively.
Naturally, both machine programming and open-endedness will make heavy use of the module motifs such as causal discovery, ABM, multi-physics and multi-scale modeling.

The Nine Motifs of Simulation Intelligence (SI) are useful independently, yet when integrated they can provide synergies for advanced and accelerated science and intelligence. 
It is through these motif synergies that human-machine teams and AI agents can broaden the scientific method and potentially expand the space of achievable knowledge (Fig.~\ref{fig:knowledge}).
Implementation and some integration of the methods in existing machine learning frameworks and programming languages is doable, especially with growing ecosystems around frameworks based on autodiff and vector operations (notably PyTorch~\cite{Paszke2017AutomaticDI} and Tensorflow~\cite{Abadi2016TensorFlowAS}), as well as probabilistic programming systems at multiple levels of abstraction (such as high-level Pyro~\cite{Bingham2019PyroDU} and PyMC3~\cite{Salvatier2016ProbabilisticPI} based on Python, and lower level Gen~\cite{CusumanoTowner2019GenAG} based on Julia~\cite{bezanson2017julia}).
For future work, we aim to develop a more purpose-build framework for integrating the motifs. It would be advantageous to use a functional paradigm where models are represented (implicitly or explicitly) as functions; in homage to Pearl, functions could be pi functions (generative direction), lambda functions (inference direction), or coupled pairs of pi and lambda functions. Functional representations lend naturally to composition, abstraction, and multi-scale reasoning, and SI methods work by analyzing these functions.
Such a framework would bring multiple motifs together by composing these analyses -- e.g. probabilistic programming, multi-scale physics, differentiable programming, and causal discovery used together in a principled way in the same application. In general this can provide a unifying framework for machine intelligence, enabling the integration of our best methods for learning and reasoning.

Next we highlight several more motif integrations, amongst the many synergies through the SI stack to further develop:

\paragraph{\textit{Causal reasoning with probabilistic programs}}
The execution of a simulator results in one or more simulation traces, which represent various paths of traversing the state space given the underlying data-generating model, which is not unlike the execution traces resulting from probabilistic programming inference. Formalizing the sequence of simulator states as a sequence of (probabilistic) steps, we can define a trace-based definition of causal dependency, which we introduce in an example below. With a simulator that produces a full posterior over the space of traces, as in probabilistic programming, we can attempt to draw causal conclusions with causal inference and discovery algorithms.

As alluded to in the probabilistic programming motif, the abstractions of probabilistic programming languages (PPL) provide ample opportunity to integrate the mathematics of causality with machine learning, which thus far has been non-trivial to say the least \cite{Scholkopf2021TowardCR,HernndezOrozco2020AlgorithmicPM}. There are several non mutually exclusive ways we suggest viewing this integration:
\begin{enumerate}
    \item A probabilistic program is fundamentally a simulator that represents the data-generating process. As mentioned above, to address causal queries we must know something about the data generation process. A promising approach is to build a probabilistic program in Box's model-inference-criticism-repeat loop \cite{Blei2014BuildCC} to arrive at a model that robustly represents the true data generating process, and use this data generator to infer causal relationships in the system -- see Fig.~\ref{fig:causal-invert}. Said another way, one can conceptualize a structural causal model as a program for generating a distribution from independent noise variables through a sequence of formal instructions \cite{Hardt2021PatternsPA}, which also describes a probabilistic program. The Omega PPL is a newer framework that provides initial steps towards this type of causal interpretation \cite{Tavares2019SoftCF}. A resource-bounded approach to generate computable models was also introduced in the context of algorithmic information dynamics\cite{nmi} leading to large sets of generative models sorted by algorithmic likeliness.

    \item PPL by definition must provide the ability to draw values at random from distributions, and the ability to condition values of variables in a program via observations \cite{Gordon2014ProbabilisticP}. These two constructs are typically implemented within a functional or imperative host language as $sample$ and $observe$ macros, respectively. Similarly, one can readily implement the \textit{do}-operator from the aforementioned do-calculus of causality \cite{Pearl1995CausalDF, Tucci2013IntroductionTJ}, to force a variable to take a certain value or distribution. This allows simulating from interventional distributions, provided the structural causal model (including the governing equations) is known. Witty et al. \cite{Witty2019BayesianCI} have provided intriguing work in this direction, although limited both in expressiveness and scope -- more development in robust PPL is needed. In general, via \textit{do}-calculus or otherwise, the abstractions of PPL enable interventions, notably in the context of PPL-based simulators: the simulation state is fully observable, proceeds in discrete time steps \cite{Wood-AISTATS-2014}, and can be manipulated such that arbitrary events can be prevented from occurring while the simulation is running.

    \item PPL provide the mechanisms to easily distinguish intervention from observational inference, and thus implement counterfactual reasoning. A counterfactual query asks ``what-if'': \textit{what would have been the outcome had I changed an input?} More formally: \textit{what would have happened in the posterior representation of a world (given observations) if in that posterior world one or more things had been forced to change?} This is different from observational inference as it (1) fixes the context of the model to a ``posterior world'' using observational inference, but (2) then intervenes on one or more variables in that world by forcing each of them to take a value specified by the programmer or user. Interventions in the ``posterior'' world can cause variables — previously observed or otherwise — to take new values (i.e., ``counter'' to their observed value, or their distribution). Thus for counterfactual reasoning, we build a probabilistic program that combines both conditioning and causal interventions to represent counterfactual queries such as \textit{given that X is true, what if Y were the case?} Recent extensions to the Omega language mentioned above aim to provide counterfactual queries via probabilistic programming, again with a version of Pearl's \textit{do}-operator \cite{Tavares2019ALF}.
\end{enumerate}


\paragraph{\textit{Causal surrogates and physics-infused learning}}

Similar to the perspective that a probabilistic program encodes a causal data-generating process, one can view a system of differential equations as encoding causal mechanisms of physical systems. To be more precise, consider the system of differential equations $dx/dt = g(x)$ where $x$, $x(0) = x_0$, and $g(x) := nonlinearity$. The Picard–Lindelöf theorem \cite{Nevanlinna1989RemarksOP} states there is a unique solution as long as $g$ is Lipschitz, implying that the immediate future of $x$ is implied by its past values. With the Euler method we can express this in terms of infinitesimal differentials: $x(t + \delta t) = x(t) + dt g(x)$. This represents a causal interpretation of the original differential equation, as one can determine which entries of the vector $x(t)$ cause the future of others $x(t+dt)$ \cite{Scholkopf2019CausalityFM}. A generalization of this approach in the context of algorithmic probability covers all computable functions including differential equations and mathematical models~\cite{nmi}. And in \cite{HernndezOrozco2020AlgorithmicPM}, it was shown that differentiability is not needed to perform an effective search optimization. The implication is then the algorithmic or physics-infused ML methods we discussed can represent the causal dynamics of physical systems, whereas neural networks fundamentally cannot, and differentiability is not the essential feature believed to allow neural networks to minimize loss functions -- see Table \ref{tab:causal}, and also Mooij et al. \cite{Mooij2013FromOD} for a formal link between physical models and structural causal models. This provides theoretical grounding that algorithmic and physics-infused ML can be used as causal surrogate models which are a more accurate representation of the true causal (physical) processes than universal function approximators (including physics-informed neural networks). This insight also suggests that using algorithmic physics-infused surrogates may provide grounded counterfactual reasoning. 
We highlight several specific approaches: 

\textit{Continuous-time neural networks} are a class of deep learning models with their hidden states being represented by ordinary differential equations (ODEs) \cite{Funahashi1993ApproximationOD}. Compared to discretized deep models, continuous-time deep models can enable approximation of functions classes otherwise not possible to generate -- notably the Neural ODE \cite{Chen2018NeuralOD} which showed CTNN can perform adaptive computations through continuous vector fields realized by advanced ODE solvers. Vorbach et al. \cite{Vorbach2021CausalNB} investigate various CT network formulations and conditions in search of causal models. 
The popular Neural ODEs do not satisfy the requisite conditions for causal modeling simply because the can't account for interventions; the trained Neural ODE statistical model can predict in i.i.d. settings and learn from data, but it cannot predict under distribution shifts nor implicit/explicit interventions.
Further, Neural ODEs cannot be used for counterfactual queries \cite{Mooij2013FromOD}, which naturally follows from the causal hierarchy discussed previously (see Fig. \ref{fig:causallevels}). 
Liquid time-constant networks (LTCs) \cite{Hasani2021LiquidTN} do satisfy the properties of a causal model: The uniqueness of their solution and their ability to capture interventions make their forward- and backward- mode causal. Therefore, they can impose inductive biases on their architectures to learn robust and causal representations \cite{Lechner2020NeuralCP}.
By this logic, we suggest UDEs can also be used as causal models of physical systems and processes.
We also suggest the use of these ``causal'' models in practice may be more elusive than structural causal models that make assumptions and confounders explicit (whereas these NN-based approaches do not).

\textit{Physics-infused Gaussian processes in causal discovery --} 
Earlier we described a ``active causal discovery'' approach where an agent iteratively selects experiments (or targeted interventions) that are maximally informative about the underlying causal structure of the system under study. The Bayesian approach to this sequential process can leverage a Gaussian process (GP) surrogate model, effectively implementing a special case of Bayesian optimization. The use of a physics-infused GP can provide robustness in the sense that the physical process being simulated is grounded as a causal surrogate -- i.e., encoding causal mechanisms of said physical system -- rather than a nonparametric function approximator. The several physics-infused GP methods we highlighted are \textit{Numerical GPs} that have covariance functions resulting from temporal discretization of time-dependent partial differential equations (PDEs) which describe the physics \cite{Raissi2018NumericalGP, Raissi2018HiddenPM}, modified Mat\'{e}rn GPs that can be defined to represent the solution to stochastic partial differential equations \cite{Lindgren2011AnEL} and extend to Riemannian manifolds to fit more complex geometries \cite{Borovitskiy2020MaternGP}, and the \textit{physics-informed basis-function GP} that derives a GP kernel directly from the physical model \cite{Hanuka2020PhysicsinformedGP}.

\textit{Algorithmic Information Dynamics (AID)} \cite{scholarpedia2020} is an algorithmic probabilistic framework for causal discovery and causal analysis. It enables a numerical solution to inverse problems based on (or motivated by) principles of algorithmic probability. AID studies dynamical systems in software space where all possible computable models can be found or approximated under the assumption that discrete longitudinal data such as particle orbits in state and phase space can approximate continuous systems by Turing-computable means. AID combines perturbation analysis and algorithmic information theory to guide a search for sets of models compatible with observations, and to precompute and exploit those models as testable generative mechanisms and causal first principles underlying data and processes.

\begin{table}[!t]
\centering
\vskip 10em
\begin{tabular}{l|ccccccccccccccccccccccc}
& \begin{rotate}{60} Predict in I.I.D. setting \end{rotate}
& \begin{rotate}{60} Predict under shifts/interventions \end{rotate}
& \begin{rotate}{60} Answer counterfactual queries \end{rotate}
& \begin{rotate}{60} Obtain physical insight \end{rotate}
& \begin{rotate}{60} Learn from data \end{rotate}
\\
\hline

\rowcolor{purp}
{Mechanistic / Physical} & \checkmark & \checkmark & \checkmark & \checkmark & ? \\

{Structural causal (SCM)} & \checkmark & \checkmark & \checkmark & ? & ? \\

\rowcolor{purp}
{Causal graph}& \checkmark & \checkmark & - & ? & ? \\

{Physics-infused} & \checkmark & \checkmark & ? & \checkmark & \checkmark \\

\rowcolor{purp}
{Probabilistic graph (PGM)} & \checkmark & \checkmark & - & ? & \checkmark \\

{Physics-informed} & \checkmark & \checkmark & - & - & \checkmark \\

\rowcolor{purp}
{Statistical} & - & - & - & - & \checkmark \\

\hline
\end{tabular}
\vskip 1em
\caption{
Expanding on the taxonomy of models in Peters et al. \cite{Peters2017ElementsOC} to show characteristics of the most detailed / structured models (top) to the most data-driven / \textit{tabula rasa} models (bottom). The former, a mechanistic or physical model, is usually in terms of differential equations. The latter is often a neural network or other function approximator that provides little insight beyond modeling associations between epiphenomena. Causal models are in between with characteristics from both ends (as are the semi-mechanistic models we discussed in the surrogate modeling motif).
}
\label{tab:causal}
\end{table}


\paragraph{\textit{Multi-fidelity simulation of electron paths for chemical reactions}}
In an earlier example of differentiable and probabilistic programming we introduced how generative modeling in molecular synthesis can help mitigate inefficiencies in the standard design-make-test-analyze molecular discovery process. In that specific method we assumed a ``reaction outcome predictor'' exists \cite{Bradshaw2020BarkingUT}. That ability, to reliably predict the products of chemical reactions, is of central importance to the manufacture of medicines and materials, and to understand many processes in molecular biology. In theory all chemical reactions can be described by the stepwise rearrangement of electrons in molecules \cite{Herges1994CoarctateTS}. In practice this sequence of bond-making and breaking is known as the \textit{reaction mechanism}, which can be described as several levels of abstraction (corresponding do multiple spatial scales) \cite{Bradshaw2019AGM}: computationally expensive quantum-mechanical simulations of the electronic structure at the lowest level, and rules-based methods for transforming reactant molecules to products in a single step that is often far too simplifying. A conceptual abstraction in the middle of these levels is used by chemists in practice: a qualitative model of quantum chemistry called \textit{arrow pushing}, or sequences of arrows which indicate the path of electrons throughout molecular graphs \cite{Herges1994CoarctateTS}. Bradshaw et al. aim to leverage the domain knowledge embedded in arrow-pushing diagrams to machine-learn the mechanism predictions -- in contrast to other ML approaches for directly predicting the products of chemical reactions \cite{Coley2017PredictionOO,Jin2017PredictingOR,Segler2017NeuralSymbolicML,Wei2016NeuralNF, Lee2019MolecularTU}. Much like the interpretability the emerges from causal structural models versus purely data-driven ones, the resulting mechanism models are more interpretable than product prediction models. 

\begin{figure}[!ht]
\centering
\includegraphics[width=0.8\linewidth]{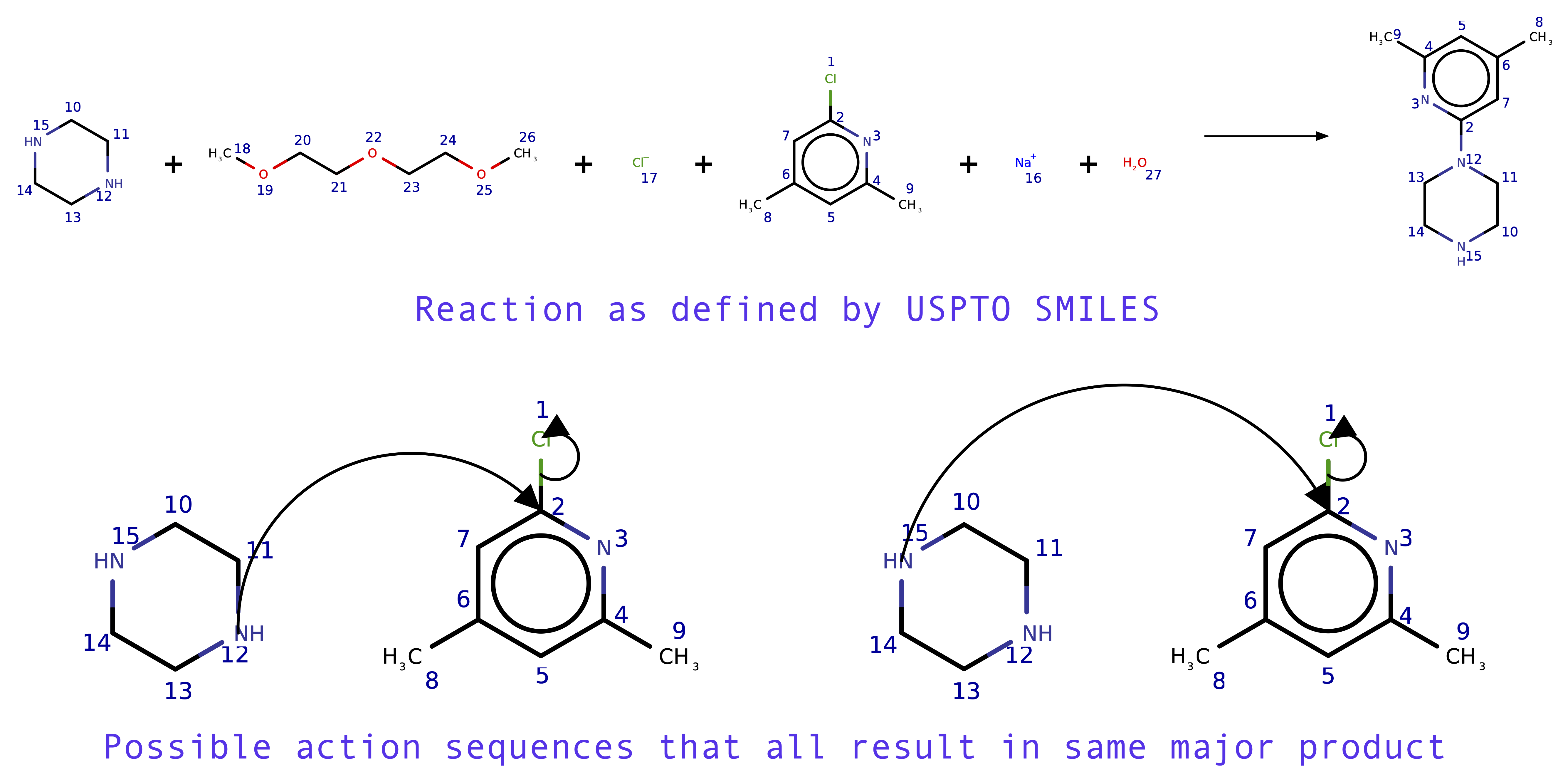}
\caption{
Example showing how different electron paths can form the same products, for instance due to symmetry. Thus modeling and interpreting the possible electron actions sequences are necessary for reliable evaluation of paths and prediction of products. The generative modeling approaches described in the text (using GNNs and probabilistic programming) are effective in overcoming this challenge. 
(Figure reproduced from Ref.~\cite{Bradshaw2019AGM})
}
\label{fig:electronpaths}
\end{figure}

The aim of their approach is to describe the problem of predicting outcomes of chemical reactions -- i.e., identifying candidate products, given a particular set of reactants - as a generative process which simulates arrow-pushing diagrams conditional on the reactant set. The generative modeling approach comes in handy with the next ``choice'' of atom in the sequence, which is something domain-expert humans have essentially zero intuition about. Since this would simulate a mechanism, even though it is an approximate one, it is useful as a predictive model which does more than simply produce an answer (at the very least, the predicted mechanism can function as a type of explanation). The arrow pushing diagrams are sequences of arrows from atom to atom, which can be viewed as sequentially defining conditional distributions over the next atoms (i.e., a discrete distribution over atoms in the reactants), subject to some simple chemical constraints. The aim is to learn a distribution over model parameters for the probability of a series of electron actions that transform a set of reactant graphs into product graphs, which can also include graphs of reagents that aid the reaction process. 
In the current implementation the generative model is parameterized by a graph neural network and implemented in the autodiff (deep learning) library PyTorch. However, the well-practiced probabilistic programmer will recognize the direct analogy between series of conditional distributions and probabilistic programming inference, in particular sequential Monte Carlo based methods. The implementation in a PPL has notable advantages, for instance trace-based interpretability: different electron paths can form the same products (see Fig. \ref{fig:electronpaths}), for instance due to symmetry, which can be inspected with PPL traces.

With both GNN and probabilistic programming, we can more easily encode the constraints imposed by chemistry. For GNN, this is the geometric deep learning and equivarience properties we've discussed. Both also enable reasoning about multiple levels of fidelity: instead of simulating reactions at a low-level representation (involving atomic coordinates and other representations that are not understandable) we can simulate reactions at an intermediate granularity, one which we know domain experts are already familiar with and trust as an abstraction.
Further for probabilistic programming, this again exemplifies the value of PP for ``partially specified'' models, which are the rich area between data-driven (black-box, deep learning) and fully-specified simulators.
See the paper \cite{Bradshaw2019AGM} for details, which has impressive results by training on large, unannotated reaction datasets (i.e., learning unsupervised), and further learns chemical properties it was not explicitly trained on.

\paragraph{\textit{SI-based materials acceleration platforms and open-ended search}}

One property to exploit with open-ended search and optimization is the potential to discover entirely novel solutions, in spaces that otherwise would never have been explored. For instance, in socioeconomics as a variation on the multi-scale agent-based modeling AI Economist framework we discussed above -- the dynamic optimization scheme suggests an alternative to typical objective-based convergence criteria (corresponding to Nash equilibria \cite{Holt2004TheNE} inthe multi-agent setting) and can be extended with open-ended search algorithms to find non-obvious or even counter-intuitive socioeconomic solutions. 
A different sector has been gaining interest in AI and simulation tools to search for \textit{de novo} solutions, pursuing ``materials acceleration platforms'' for autonomous experimentation in chemistry \cite{FloresLeonar2020MaterialsAP}. 
A fundamental goal of materials science and chemistry is to learn structure–property relationships and from them to discover \textit{de novo} materials or molecules with desired functionalities. 
Yet the chemical space is estimated to be ~$10^{60}$ organic molecules \cite{Drew2012SizeEO} and the space of materials is even larger \cite{FloresLeonar2020MaterialsAP}. 
The traditional approach is based on human expert knowledge and intuition, resulting in a process that is slow, expensive, biased, and limited to incremental improvements.
Even with high-throughput screening (HTS), the forward search process is limited to predesignated target areas \cite{CorreaBaena2018AcceleratingMD}.
Inverse design methods can provide significant efficiency gains \cite{sanchez2018inverse}, representing a key component for the molecule/hypothesis-generation phase of automating the discovery of chemicals and materials -- see Fig. \ref{fig:inverse-batteries}. As we described, inverse design focuses on efficient methods to explore the vast chemical space towards the target region, and fast and accurate methods to predict the properties of a candidate material along with chemical space exploration \cite{noh2020machine}.


\begin{figure}[!ht]
\centering
\includegraphics[width=0.9\linewidth]{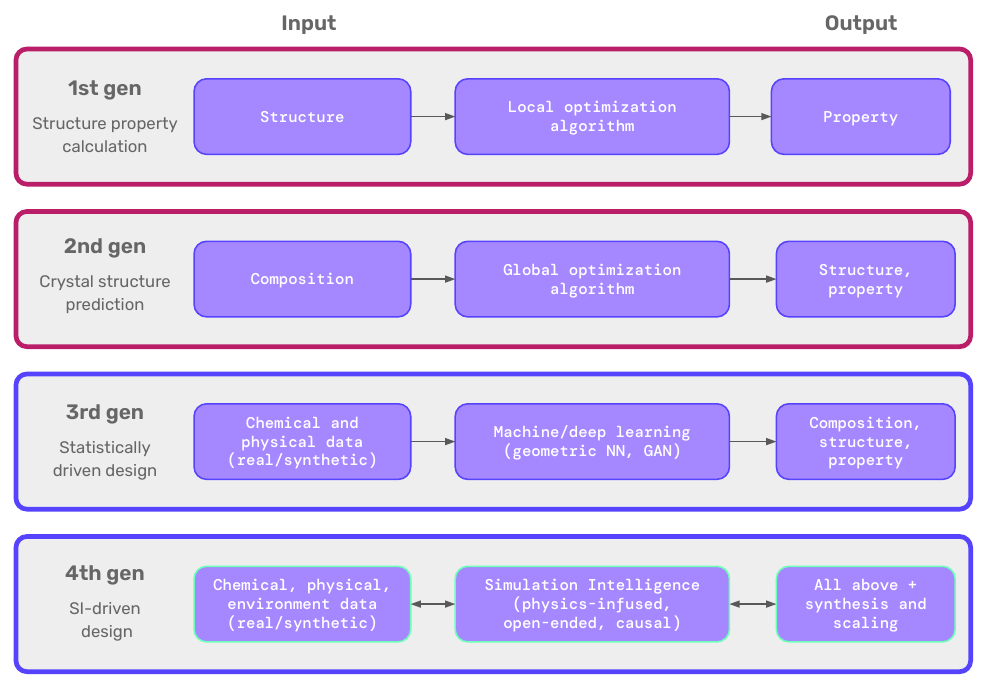}
\caption{Evolution of the research workflow in computational chemistry (extended from the ``3rd generation'' graphic in \cite{Butler2018MachineLF}). The standard paradigm in the first-generation approach is to calculate the physical properties of an input structure, which is often performed via an approximation to the Schrödinger equation combined with local optimization of the atomic forces. In the second-generation approach, by using global optimization (for example, an evolutionary algorithm), an input of chemical composition is mapped to an output that contains predictions of the structure or ensemble of structures that the combination of elements are likely to adopt. The third-generation approach is to use machine-learning techniques with the ability to predict composition, structure, and properties provided that sufficient data are available and an appropriate model is trained.
The outdated first and second generations (red outline) are replaced by the third and fourth generation approaches (blue outline) that excel in efficiency and producing optimal solutions. We suggest these two vectors -- \textit{efficiency and optimality} -- are where SI-based methods can bare fruit. For efficiency, in practice we see a significant roadblock in the synthesis and scaling steps of new chemicals and materials -- see Fig. \ref{fig:pipeline} -- which can be addressed with the integrated, closed loop approaches of SI, such as simulation-based inference (and ``human-machine inference'' in Fig. \ref{fig:activesci}), physics-\textit{infused} learning, causal discovery (Fig. \ref{fig:causal-invert}), and inverse problem solving (Fig. \ref{fig:inverse-batteries}). This interconnectedness is loosely reflected in the figure with forward-backward arrows to contrast with the standard forward computational approaches.
As for optimality, an overarching goal for SI is to both expand and intelligently navigate the space of potential solutions. The implementation of open-ended optimization in this context has exciting prospects. It remains to be seen where and how ``quantum learning'' (i.e., quantum computing, ML, and data) plays a role in this evolution. We discuss quantum learning in the Accelerated Computation and Honorable Mention sections. 
}
\label{fig:chemvolution}
\end{figure}

The focus in materials acceleration platforms has thus far been on objective-based optimization schemes for autonomous chemical experimentation, including approaches that combine objectives with user-defined preferences (with similar motivations as the NA-IEC approach we described for open-ended search) \cite{Hse2018ChimeraEH}, and in Bayesian optimization schemes \cite{Hse2018PhoenicsAB}. Specific implementations include he ``Chemputer'' \cite{Steiner2019OrganicSI} for organic synthesis (where an automated retrosynthesis module is coupled with a synthesis robot and reaction optimization algorithm) and ``Ada'' \cite{MacLeod2020SelfdrivingLF} (Autonomous discovery accelerator) for thin-film materials (with a purpose-built software package to coordinate researchers with lab equipment and robots \cite{chemos}). These operate on closed-loop optimization schemes, which some suggest is a requirement for fully autonomous discovery systems \cite{FloresLeonar2020MaterialsAP,chemos}. 
It is interesting to explore if open-ended search and optimization methods can advance these discovery platforms, and how they could be implemented in the environment.
In such an expansive space of molecules, one can leverage open-ended search to not only explore more broadly, but also in ways that increase uncertainty (and entropy) towards potentially unknown territories that can match target characteristics. We are not yet aware of work in this direction, and even so it may call for a rethinking of the broader discovery pipeline. That is, the closed loop setting of current approaches may not support open-ended search in optimal ways. Continual learning methods may be usable, as the modeling and search algorithms aim to improve on subsequent passes through the cycle, but that is still very much objective-based and does not allow for exploration with ongoing, targeted increases in entropy. Perhaps an offline process for open-ended search that is decoupled from the main process is needed, along with some mechanism for information flow between the two (so the open-ended search can learn from the materials synthesis process, for example). And to consider integrating robotics with open-endedness would introduce additional challenges (not to mention risks). One can potentially look to surrogate modeling to alleviate some of the challenges implementing open-ended search and optimization, but at this point we are well into speculative territory and there is still much platform engineering and baseline experiments to be done.

It's straightforward to connect materials acceleration platforms with nearly all the SI motifs, as well as inverse design and human-machine teaming workflows. For instance, semi-mechanistic modeling can be advantageous for incorporating domain expert priors into the AI-driven search, and well as physics-infused learning. A significant challenge with chemical and materials design is the translation of properties towards production scaling, which can potentially be alleviated with improved multi-scale modeling techniques. With such a massive, complex environment that cannot possibly be understood by humans, causal modeling and inference can prove valuable for interpreting the solution space and navigating cause-effect relationships in the overall process. 

% ...

\paragraph{\textit{Inducing physics programs with open-ended learning}}

One would suggest that open-ended methods train in real-time rather than batches or episodes like mainstream deep learning -- online learning could be a requirement of any model developed for open-ended use. However, ``DreamCoder'' \cite{Ellis2020DreamCoderGG} provides an interesting counterexample with ``wake-sleep'' \cite{Dayan2000HelmholtzMA} probabilistic program learning.

\begin{figure}[!ht]
\centering
\includegraphics[width=1.0\linewidth]{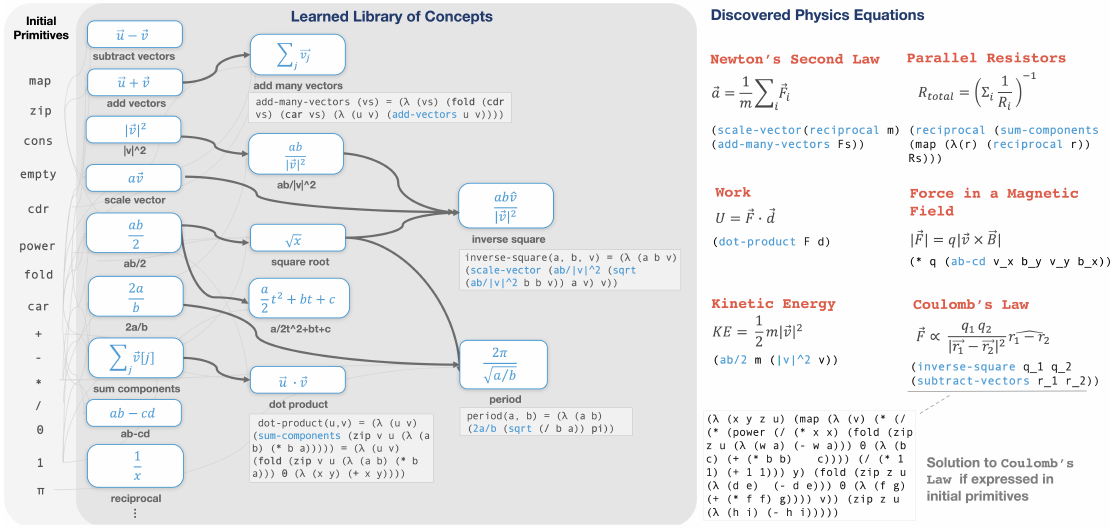}
\caption{Learning a language for physical laws starting with recursive list routines such as map and fold. DreamCoder observes numerical data from 60 physical laws and relations, and learns concepts from vector algebra (e.g., dot products) and classical physics (e.g., inverse-square laws). (Original figure from Ellis et al. \cite{Ellis2020DreamCoderGG})
}
\label{fig:dreamcoder}
\end{figure}

DreamCoder is a system that learns to solve problems by writing programs -- i.e., a program induction perspective on learning, where learning a new task amounts to searching for a program that solves it. 
This perspective provides a number of advantages:
Nonetheless practical use is limited, as applications of program induction require not only carefully engineered domain-specific language (DSL) with strong biases/priors on the types of programs to find, but also a search algorithm hand-designed to exploit that DSL for fast inference as the search space is otherwise intractable. 
The wake-sleep approach in DreamCoder resolves these issues by learning to learn programs: its learned language defines a generative model over programs and tasks, where each program solves a particular hypothetical task, and its inference algorithm (parameterized by a neural network) learns to recognize patterns across tasks in order to best predict program components likely to solve any given new task. This approach produces two distinct kinds of domain expertise: (1) explicit declarative knowledge, in the form of a learned DSL that captures conceptual abstractions common across tasks in a domain; and (2) implicit procedural knowledge, in the form of a neural network that guides how to use the learned language to solve new tasks, embodied by a learned domain-specific search strategy \cite{Ellis2020DreamCoderGG}. This succinct description does not do DreamCoder justice, as the paper thoroughly discusses the approach and learned representations, and presents illustrative examples like that in Fig. \ref{fig:dreamcoder}.
The authors show this setup allows learning to scale to new domains, and to scale within a domain provided there are sufficiently varied training tasks. 
DreamCoder is open-ended in the sense that it learns to generate domain knowledge ceaselessly, implements a continual learning algorithm, and doesn't have a specific optimization objective.
Another important distinction from deep neural networks is the system is not learning \textit{tabula rasa}, but rather starting from minimal bases and discovering the vocabularies of functional programming, vector algebra, and physics.

\paragraph{\textit{Physical computing}}
Here we highlight several promising directions for improving physical computation with the incorporation of SI methods: optimizing the design of sensors and chips.

\textit{ML-enabled intelligent sensor design} is the use of inverse design and related machine learning techniques to optimally design data acquisition hardware with respect to a user-defined cost function or design constraint.
Here is a general outline of the process (motivated by Ballard et al.~\cite{Ballard2021MLSensorD}):
\begin{enumerate}
    \item Standard engineering practices produce an initial design $\alpha_0$, or a randomized initialization of parameters and spatial configuration (in terms of transduction elements or sensing components, for example).
    \item Acquisition of raw sensing data ${X_i, y_i}$ from the prior $\alpha_{i-1}$, and data preprocessing and transformations to $X'_i$ such that the data is amenable to machine learning (e.g., normalized, Gaussian noise, etc.).
    \item Train a ML model (typically a neural network or Gaussian process, potentially a probabilistic program) to output sensing results as $\hat{y}_i$
    \item A cost function $J(y_i , \hat{y}_i)$ is used to evaluate the learned model, using the ground-truth sensing information $y_i$ -- here it is useful to develop a problem-specific cost function with physics-informed constraints (see the multi-physics motif section). 
    \item The results from step (4) inform a redesign $\alpha_{i+1}$ that consists of adjusting or eliminating the less informative/useful features, and possibly replacing such features with alternative sensing elements. Ideally one can use a Bayesian optimization (BO) outer-loop to inform the subsequent design, where the learned NN or GP serves as the surrogate (as in the surrogate modeling motif). Future work may investigate BO-based neural architecture search methods (e.g. \cite{White2021BANANASBO}) applied to sensor components and modules.
    \item Repeat from step (2) for $i = 1, .., N$, where termination conditions may be the number of budgeted cycles ($N$), convergence on design or outcome, or achieving a down-stream cost function. Depending on the surrogate model choice and optimization scheme, it may be useful to employ transfer learning techniques to reduce the data burden of subsequent iterations.
\end{enumerate}
Such an approach can drastically improve the overall performance of a sensor and broader data ingestion/modeling system, perhaps with non-intuitive design choices. 
Consider an intriguing example in the field of synthetic biology, aiming to intelligently design RNA sequences that synthesize and execute a specific molecular sensing task: Angenent et al.~\cite{AngenentMari2020ADL} developed a surrogate-based design framework with a neural network trained on RNA sequence-to-function mappings, resulting in an \textit{in silico} design framework for engineering RNA molecules as programmable response elements to target proteins and small molecules. The data-driven approach outperforms the prediction accuracy resulting from standard thermodynamic and kinetic modeling.
In general data-driven, ML-based sensor designs can outperform ``intuitive'' designs that are solely based on analytical/theoretical modeling, but there are non-trivial limitations: generally large, well-characterized training datasets are needed to engineer and select sensing features that can statistically separate out various inherent noise terms or artifacts from the target signals of interest. 
Not to mention the the curse of dimensionality, where the high-dimensional space of training data may drown-out the meaningful correlations to the target sensing information.
See Ballard et al.~\cite{Ballard2021MLSensorD} for more discussion on ML-enabled intelligent sensor design.


\textit{ML-based chip design} aims to automate and optimize the various processes in designing, fabricating, and validating computer chips.
In particular, recent work has made significant strides in ML-based ``chip floorplanning'' \cite{Mirhoseini2021AGP}, or designing the physical layout of a computer chip. This slow, costly process involves placing hypergraphs of circuit components (such as macros (memory components) and standard cells (logic gates such as NAND, NOR and XOR)) onto chip canvases (two-dimensional grids) so that performance metrics (e.g., power consumption, timing, area, and wire length) are optimized, while adhering to hard constraints on density and routing congestion.
Engineers approach this manual design task over the course of months per chip layout, where performance evaluation of each design takes multiple days.
Mirhoseini et al. \cite{Mirhoseini2021AGP} developed an reinforcement learning (RL) approach, not unlike those described in the agent-based modeling motif, by framing the task as a sequential Markov decision process (MDP) as follows:
\begin{itemize}
    \item States: hypergraph (as an adjacency matrix), node features (width, height, type), edge features (number of connections), current node (macro) to be placed, and metadata of the netlist graph (routing allocations, total number of wires, macros and standard cell clusters).
    \item Actions: all possible locations (grid cells of the chip canvas) onto which the current macro can be placed subject to predefined hard constraints on density or blockages.
    \item Transitions: probability distribution over next states, given a state and an action.
    \item Rewards: always zero but for the last action, where the reward is a negative weighted sum of proxy wirelength, congestion, and density. (See the paper for details \cite{Mirhoseini2021AGP})
\end{itemize}

A graph convolutional NN is used to learn feature embeddings of the macros and other hypergraph components -- this architecture provides advantageous geometric properties for this design space, which we discuss more in the ``honorable mention motifs'' later.
The benchmark results of this approach show the generation of chip floorplans that are comparable or superior to human experts in under six hours, compared to multiple months of manual human effort.
This is a nice practical example of AI-driven knowledge expansion in Fig. \ref{fig:knowledge}: the artificial agents  approach the problem with chip placement experience that is magnitudes greater in size and diversity than any human expert.

As alluded to above, this is one step of many in a resource-intensive process of computer chip development. There are many ways automation and human-machine teaming can potentially optimize and accelerate the process, including additional ways simulation-AI can advance chip design: for example, the Modulus~\cite{Hennigh2020NVIDIASA} platform of NN-solvers has been applied to the problem of design optimization of an FPGA heat sink, a multi-physics problem that can be approached with multiple parallel ML surrogates (with one NN for the flow field and another NN for the temperature field).

\textit{Physical Neural Networks --}
In the simulation intelligence spirit of minimizing the gaps between the \textit{in silico} and \textit{in situ} worlds, \textit{physical neural networks (PNNs)} are an intriguing direction for physical learning hardware. That is, PNNs describe hierarchical physical computations that execute deep neural networks, and trained using backpropagation through physics components. The approach is enabled by a hybrid physical-digital algorithm called \textit{physics-aware training (PAT)}, which enables the efficient and accurate execution of the backpropagation algorithm on any sequence of physical input-output transformations, directly \textit{in situ} \cite{Wright2021DeepPN}.
For a given physical system, PAT is designed to integrate simulation-based and physical modeling in order to overcome the shortcomings in both. That is, digital models (or simulations) describing physics are always imperfect, and the physical systems may have noise and other imperfections. Rather than performing the training solely with the digital model (i.e., gradient-based learning with the simulator), the physical systems are directly used to compute forward passes, which keep the training on track -- one can view this as another type of constraint in the physics-infused learning class of methods we discussed earlier.
Naturally this necessitates a differentiable simulator of the physical system, one to be built with differentiable programming and perhaps additional SI motifs.
Wright et al. \cite{Wright2021DeepPN} trained PNNs based on three distinct physical systems (mechanical, electronic, and optical) to classify images of handwritten digits, amongst other experiments, effectively demonstrating PAT and the overall physical computing (or physical learning) concept. They also describe in depth the mathematical foundations of all components of the approach, including the implementation of autodiff for physics-aware training.
The overall approach is very nascent but could expand possibilities for inverse design, such as inverse design of dynamical engineering systems via backpropagation through physical components and mechanisms. Physical neural networks may also facilitate cyber-physical simulation environments with a variety of machines and sensor systems, and may lead to unconventional machine learning hardware that is orders of magnitude faster and more energy efficient than conventional electronic processors.

\hfill \break
\subsection{HUMAN-MACHINE TEAMING}

The stack in Fig. \ref{fig:os} indicates the motifs and applications interact with people of diverse specialties and use-cases. Not to mention the human-machine inference methods we've described (in simulation-based inference, causal reasoning, and other motifs) call for human-in-the-loop processes. To this end we define \textit{human-machine teaming (HMT)} to be the mutually beneficial coordination of humans and machine intelligence systems (which include AI and ML algorithms and models, various data and information flows, hardware including compute architecture and sensor arrays, etc.).
HMT overlaps with the more well-established field of \textit{human-computer interaction (HCI)}, which integrates the disciplines of computer science, cognitive science, and human-factors engineering in order to design technologies that enable humans to interact with computers in efficient and novel ways.\footnote{
We intentionally use the name ``machine'' in a general sense, overlapping with and extending beyond HCI, human-AI interaction, and human-robot interaction. The development of HMT is to include the variety of machines we've discussed in this paper, from \textit{in silico} agents and algorithms, to \textit{in situ} nuclear fusion reactors and particle accelerators, and of course the nascent field of machine programming.
}

Common practice in machine learning is to develop and optimize the performance of models in isolation rather than seeking richer optimizations that consider human-machine teamwork, let alone considering additional human-in-the-loop challenges affecting model performance, such as changes in policies or data distributions. 
Existing work in HMT includes systems that determine when to consult humans, namely when there's low confidence in predictions \cite{Horvitz2007ComplementaryCP,Kamar2012CombiningHA,Raghu2019TheAA}, or informing the human annotator which data to label for active learning \cite{Settles2012ActiveL,Hu2019ActiveLW}.


HMT research addresses several fundamental issues with applicability across domains of science and intelligence:
\begin{itemize}
    \item How humans and machines communicate
    \item How to model the relationship between humans and machines for accomplishing closed feedback loops
    \item How to quantify and optimize human-machine teamwork
    \item Machines learning how to best compliment humans \cite{Wilder2020LearningTC}
    \item How human-machine dynamics can influence and evolve the characteristics of errors made by humans and AI models, especially in continual learning contexts
    \item How to quantify human-machine uncertainties and predict reliability, especially in the context of broader and dynamic systems \cite{lavin2021technology}
    \item How to provide theoretical guarantees for various HMT subclasses (e.g., online optimization of experiments, active causal discovery, human-machine inference (Fig. \ref{fig:activesci})), and how to empirically validate them
    \item How to capture and express the human-machine interactions for a particular application domain \textit{-- where do general methods suffice, and where are domain-specific algorithms and interfaces needed?}
    \item Synergistic human-AI systems can be more effective than either alone -- we should understand this tradeoff precisely, which will likely vary in metrics and dimensions depending on the domain and use-case
    \item Laboratory processes and equipment are currently designed for the traditional, deductive scientific method, where tools assist human scientists and technicians. Simulation intelligence provides the opportunity to rethink this human-machine relationship towards new science methods that are inductive, data-driven, and managed by AI (beyond standard automation).
\end{itemize}

The standard HMT approach is to first train a model $m$ on data $x$ for predicting $y$, then $m$ is held fixed when developing the human-interaction components (and broader computational system) and when training a policy $q$ for querying the human. 
One of the more glaring issues with this general approach is the introduction of $q$ can cause the $x$ and $y$ distributions to shift, resulting in altered behavior from $m$ that necessitates retraining. 
Joint training approaches could mitigate this concern, and further provide a more optimal teaming workflow. For instance, Wilder et al.~\cite{Wilder2020LearningTC} propose several joint training approaches that consider explicitly the relative strengths of the human and machine. A \textit{joint discriminative} approach trains $m$ while learning $q$ end-to-end from features to decisions, and a 
a new \textit{decision-theoretic} approach to additionally construct probabilistic models for both the AI task and the human response. The latter allows a follow-up step that calculates the expected value of information for querying the human, and joint training optimizes the utility of the end-to-end system. 

\begin{figure}[!ht]
\centering
\includegraphics[width=1.0\linewidth]{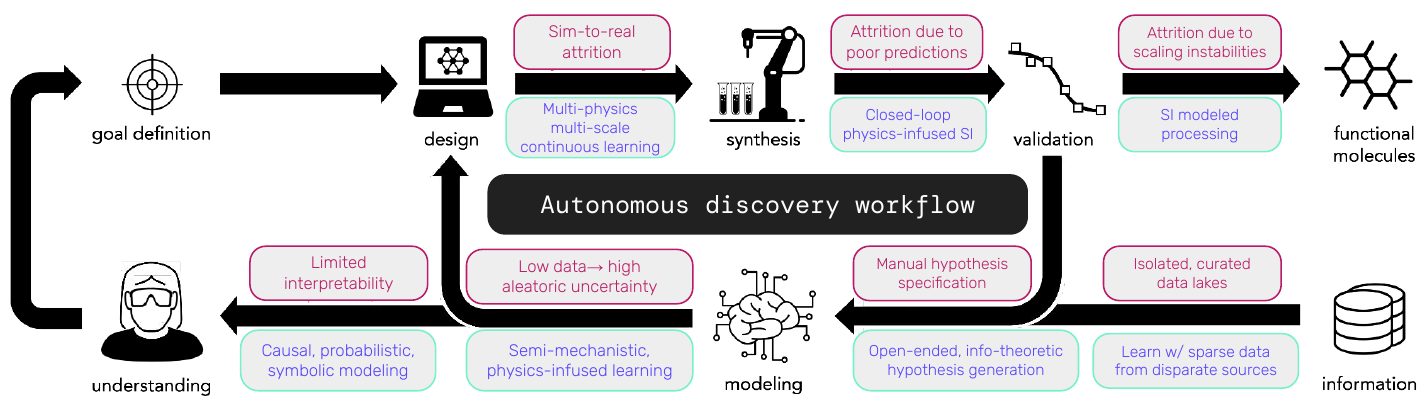}
\caption{
The AI field largely focuses on modeling and design, which represent only a subset of the AI and data systems required for highly-autonomous science. Here we show one high-level example for molecule discovery, where modeling and design must be integrated with various information flows and coordinate with upstream and downstream tasks such as human-input and material synthesis.
Plugging in mainstream AI technologies would not suffice, with shortcomings noted in red along the workflow phases. Also noted in purple (and green) are corresponding advantages that SI motifs and methods alternatively bring to the workflow. The additional use of the uncertainty-aware methods we discussed would improve the reliability of such a system.
From this view one can also envision some of the SI data challenges we describe. For example, the multitude of data consumers and generators in the workflow amplifies the risks of data inconsistencies and errors, which can be mitigated with proper data versioning and provenance. Further, provenance and data readiness steps are necessary to ensure reproducibility of the experiment process and results. We stress this is a simplified workflow that assumes data fusion and synthetic data needs are taken care of, and doesn't shed light on the significant data engineering throughout each phase.
}
\label{fig:pipeline}
\end{figure}

In addition to quantifying the value of information in human-machine dynamics, using probabilistic ML methods can also enable the quantification and propagation of uncertainties in HMT (and overall throughout SI-based workflows for automating science -- Fig. \ref{fig:pipeline}). 
We've described candidate methods in the previous section, and here point out another class of methods for uncertainty estimation in the human-machine coalition setting: \textit{Evidential learning} maps model inputs to the parameters of a Dirichlet distribution over outputs (classes), where smaller parameter values represent less evidence for a class and thus produce a broader distribution representing greater epistemic uncertainty \cite{Sensoy2018EvidentialDL,Meinert2021MultivariateDE}. Evidential learning enjoys a direct mapping to \textit{subjective logic}, a calculus for probabilities expressed with degrees of uncertainty where sources of evidence can have varying degrees of trustworthiness \cite{Jsang2016SubjectiveLA}. 
In general, subjective logic is suitable for modeling and analysing situations where the available information is relatively incomplete and unreliable, which can be utilized in HMT scenarios for uncertainty-aware logical reasoning.
One can extend this to trust calibration, a valuable yet overlooked factor in many HMT scenarios: \textit{automation bias} and \textit{automation complacency} can occur when humans too readily accept AI or machine outputs. In these cases, humans either replace their own thinking or judgement, or are less sensitive to the possibility of system malfunctions \cite{Goddard2012AutomationBA,Tomsett2020RapidTC}. Conversely, \textit{algorithm aversion} occurs when humans disregard algorithms for lack of trust, despite the algorithms consistently outperforming humans.
In general for feedback systems powered by ML, one cannot simply improve the prediction components and reliably improve performance. As in control theory, the ``Doyle effect''~\cite{Doyle1978GuaranteedMF,Petersen2014RobustCO} suggests that an improved prediction system can increase sensitivity to a modeling or data flaw in another part of the engineering pipeline -- this certainly holds for human-in-the-loop ML, and scientific ML pipelines in general.

The HMT concerns are relevant and perhaps exacerbated in the broader perspective of a full workflow like that of Fig. \ref{fig:pipeline}, where we illustrate a scientist in the context of a largely autonomous workflow for AI-driven science.

\hfill \break
\section{Simulation Intelligence in Practice}\label{sec_practical}

The implementation and integration of simulation intelligence in workflows such as this can address the noted issues in important ways. For instance, the semi-mechanistic modeling and physics-infused learning we shared earlier are essential in the modeling-understanding phase, and are also critical in the sim-to-real transfer phase(s); we suggest these need to be integrated in closed-loop ways that coordinate learning and info-sharing between upstream and downstream phases. The inverse design approaches we presented (such as Fig. \ref{fig:inverse-batteries}) could be useful to this end, and inverse problem solving may prove useful for mitigating losses in the synthesis steps -- this is an area of future SI work for us. 
More generally it can be advantageous to reshape such workflows from the perspective of SI. For instance, human-machine inference can in essence replace the left side of the diagram, and SI-based inverse design can flip/rewire much of the inner modeling-design-synthesis-validation loop. Nonetheless the data and engineering challenges remain non-trivial in practice, which we discuss next.

\subsection{Data-intensive science and computing}


As we've described, scientific datasets are different from the data typically available to ML research communities; scientific data can be high-dimensional, multi-modal, complex, sparse, noisy and entangled in ways that differ from image, video, text and speech modalities. Scientific questions often require the combination of data from multiple sources, including data to be fused from disparate and unknown sources, and across supercomputing facilities. Further, many science applications where ML and AI can be impactful are characterized by extreme data rates and volumes, highly diverse and heterogeneous scenarios, extremely low latency requirements, and often environment-specific security and privacy constraints.

For these challenges and more, data processing, engineering, sharing, governance, and so on are essential to scientific AI and simulation intelligence, which are inherently \textit{data-intensive}: 
problems characterized by data volume, velocity, variety, variability (changes both in the structure of data and their underlying semantics, especially in real-time cases), and veracity (data sources are uncontrolled and not always trustable) \cite{HPCS2015datahpc}. 
Admittedly we do not discuss nearly enough the importance of data in the SI motifs and in simulation sciences broadly -- one or more survey papers on this is warranted. Here we highlight several key concepts and references.

\paragraph{Data challenges and opportunities in merging scientific computation and AI/ML}
In recent years there have been numerous reports on data and AI/ML best-practices from various perspectives and fields.
Argonne, Oak Ridge, and Lawrence Berkeley national laboratories have recommended that computational science communities need AI/ML and massive datasets \cite{Stevens2020AIFS}, including the suggestion that a lack of data is by far the largest threat to the dawn of strongly AI-enabled biology, advocating for advances in synthetic data generation, privacy-preserving ML, and data engineering.
Leaders in the nuclear fusion community collaborate on a set of needs for AI to move the field forward \cite{Humphreys2020AdvancingFW}, calling for an industry-wide ``fusion data machine learning platform''. NASA perspectives on big data systems for climate science \cite{Schnase2015BigDC,Schnase2017MERRAAS} cover topics such as data ETL (extract-transform-load) in massive multi-tier storage environments, and the need to discover and reuse data workflows. The National Science Foundation is pushing for domain-inspired AI architectures and optimization schemes to enable big-data driven discovery \cite{Huerta2020ConvergenceOA} across various domain-specific fields such as chemistry and materials sciences \cite{Himanen2019DataDrivenMS,Blaiszik2019ADE}, and agriculture \cite{Janssen2017TowardsAN}. 
Here we highlight several of the common themes across those reports, which are data themes shared with simulation intelligence. We also expand on them and clarify what is meant by concepts such as ``ML-ready data'' and ``open data'': 

\begin{itemize}
    \item \textbf{Interoperability} is one of the tenets of ``FAIR'' \cite{Wilkinson2016TheFG}, guiding principles that define 15 ways in which data should be \textbf{f}indable, \textbf{a}ccessible, \textbf{i}nteroperable, and \textbf{r}eusable. In general these are rather straightforward: for example, to be findable data are assigned a globally unique and persistent identifier, to be reusable data are released with a clear and accessible data usage license, and so on. Although generally agreed upon, FAIR is rather high-level and warrants more specific details for the tenets. Double-clicking on interoperable is particularly important for the data-intensive applications of SI. \textit{Data interoperability}, as it pertains to AI/ML and simulation sciences, addresses the ability of computational systems to exchange data with unambiguous, shared meaning and content. The medical ecosystem is unfortunately emblematic of the main data interoperability challenges \cite{Lehne2019WhyDM} that stymie ML and scientific progress: incompatible standards, isolated and legacy databases, incompatible systems and proprietary software, inconsistent privacy and security concerns, and datasets that are inherently difficult to exchange, analyze, and interpret. Many of these interoperability challenges are a consequence of business incentives or legal needs. Other problems could be solved by at least a common ontology and a standardized representation of knowledge within this ontology -- for example, the ``SyBiOnt'' ontology for applications in synthetic biology design \cite{Misirli2016DataIA}, to bring together the collective wealth of existing biological information in databases and literature. We further suggest that such an ontology should integrate with a unanimous data lifecycle management framework in that field, where a main value-adds are proper semantic versioning for datasets and lingua franca amongst stakeholders \cite{lavin2021technology}; more on such frameworks is mentioned in the readiness entry. When considering distributed computing and HPC (both highly important in scientific computing and simulation), these challenges can obstruct the sharing and interconnection of data such that scaling the experiments is infeasible.

    \item \textbf{Data readiness} at one time could have described the preparedness and availability of data for programmer queries and data science or analytics use. The set of concerns are far more complex and dynamic with ML. And further, scientific use-cases can impose stricter requirements. Data-readiness describes the successful execution of the data tasks ahead of modeling: collating, processing, and curating data for the intended use(s). Challenges include poor and undocumented data collection practices, missing values, inconvenient storage mechanisms, intellectual property, security and privacy -- notice the overlap with the interoperability issues above. To help overcome these challenges, several frameworks for data readiness have been developed \cite{Afzal2021DataRR,Lawrence2017DataRL,lavin2021technology}. These provide data handling structure and communication tools for ML practitioners, including required implementations such as data profiling and quality analyses as a key step before data enters a ML pipeline, and continuous testing for shifts in the data and environments of deployed models. These frameworks are highly relevant for SI practitioners and systems as well.

    \item \textbf{Provenance} in scientific workflows is paramount: a data product contains information about the process and data used to derive the data product, and provenance provides important documentation that is key to preserving the data, to determining the data's quality and authorship, and to reproduce as well as validate the results. These are all important requirements of the scientific process \cite{Davidson2008ProvenanceAS}. Several overviews on the topic point to methods for scientific data provenance as a means of reproducability \cite{Davidson2008ProvenanceAS} and integrity \cite{Huang2018FromBD}. We suggest including \textit{knowledge provenance} in this theme, referring to the audit trail, lineage, and pedigree of the computed information and results. In recent years the ML community has prioritized systems with tools for versioning datasets and models, which should be embraced by all in scientific computing and simulation, and extended to versioning of environments and testbeds.
    
    \item \textbf{Synthetic data} is paramount in scientific simulation and AI/ML. 
    The primary use-case is data generation to make up for shortcomings in real data: availability, annotation, distribution (variance), representation of various modes and anomalies, computational constraints, and of course sample size. With the popularity of data-hungry deep learning in recent years, there have been many works on synthetic data generation for ML, largely in general computer vision and robotics \cite{Bousmalis2018UsingSA,JohnsonRoberson2017DrivingIT}, and also scientific areas such as medical imaging and clinical records \cite{Shin2018MedicalIS,Abay2018PrivacyPS,Torfi2020DifferentiallyPS}, social and economic systems \cite{Burgard2017SyntheticDF,Zheng2020TheAE}, particle physics \cite{Deja2019GenerativeMF} and astrophysics \cite{DeRose2019TheBF}, and human cognition and behavior \cite{Battaglia2013SimulationAA,Hamrick2016InferringMI,Gerstenberg2021ACS}.
    The validation of generated data against the real environment, and the reliability of subsequent models to perform in said environment, are of principal importance -- Rankin et al. \cite{Rankin2020ReliabilityOS}, for example, demonstrate the necessary analysis in the context of medical ML. A variety of statistical methods exist for quantifying the differences in data distributions \cite{Kar2019MetaSimLT} (and relatedly for dataset drifts \cite{Rabanser2019FailingLA})
    In some use-cases, domain-specific metrics may be called for, imposing physics-informed or other environment constraints on the synthetic data \cite{Ltjens2021PhysicallyConsistentGA}.
    
    An overarching challenge is the integrated use of synthetic and real datasets. Techniques for ``sim-to-real'' transfer learning help ML models train synthetically and deploy with real data \cite{Kar2019MetaSimLT}. 
    The sim-to-real perspective of the AI and ML field is largely dominated by simulation testbeds (including video games) for robotics use-cases.
    However in scientific use-cases we encounter numerous scenarios where 
    simulated data can be used as a catalyst in machine-based experiment understanding:
    generating synthetic diagnostics that enable the inference of quantities that are not directly measured in machine-based experiments -- e.g., extracting meaningful physics from extremely noisy signals in particle physics, and automatically identifying important nuclear fusion plasma states and regimes for use in supervised learning. Humphreys et al. \cite{Humphreys2020AdvancingFW} refer to this as one means of ``machine learning boosted diagnsotics'', which also includes the systematic integration of multiple data sources that encompasses the next theme.
    \item \textbf{Data fusion} corresponds to the integration of information acquired from multiple sources (sensors, databases, manual trials, experiment sites, etc.). Many challenges exist that are highly dependent on the complexity of application environments and scales of sensors and systems. For example, data imperfections and inconsistencies due to noise in sensors and environments, the alignment or registration of data from different sensors or machines with different frames, and more discussed in the data fusion and ML survey by Meng et al. \cite{Meng2020ASO}.
    With open data, an ongoing challenge is finding true values from conflicting information when there are a large number of sources, among which some may copy from others \cite{Dong2013DataFR}.
    Large-scale computer simulations of physical phenomena must be periodically corrected by using additional sources of information, including snapshots of reality and model forecasts. The fusion of observations into computer models is known as \textbf{\textit{data assimilation}} and is extensively used in many areas, such as atmospheric prediction, seismology, and energy applications. Assimilating multiple sources of information into computer simulations helps maintain an accurate representation of the evolution of the physical phenomena and promotes realistic long-term prediction. Yet developing efficient and scalable assimilation algorithms capable of handling complex environments with nonlinear and non-Gaussian behavior is still an open research area, which is strongly aligned with several SI research directions we've discussed: optimal design of experiments and sensor placement,  physics-infused optimization algorithms, and inverse problem solving.

    \item \textbf{Openness} -- ``open science'' can be defined as a new approach to the scientific process based on cooperative work and new ways of diffusing knowledge by using digital technologies and new collaborative tools \cite{Burgelman2019OpenSO}. Hesitancy to share can come from several places including security restraints, concerns of intellectual property (IP), and fear of being scooped by other researchers. Openness for data can be thought of as making data searchable, accessible, flexible, and reusable -- open data is thus FAIR data. Open data also invites concerns with ethics and accountability, in addition to those prevalent in AI and scientific computing broadly. Please see Hutchinson et al.~\cite{Hutchinson2021TowardsAF} and references therein for details.
    We further suggest that practical openness requires industry or domain specific standards that are agreed upon and adhered to -- from appropriate metrics and storage formats, to procedures for gathering and labeling data -- in order to resolve challenges in data acceptance, sharing, and longevity. These are also essential for multi-source databases, for instance the IDA (Image and Data Archive, \href{https://ida.loni.usc.edu/login.jsp}{ida.loni.usc.edu}) with over 150 neuroimaging studies, most of which collect data from multiple clinical sites and over several years such that shared, agreed-upon data procedures are necessary for open data.
    
\end{itemize}


\paragraph{Data pipelines and platforms for SI}

Along with the SI motifs and their integrations that we've presented, the advancement of SI calls for cohesive data platforms that exemplify the above principles.
The report on advancing nuclear fusion with ML research \cite{Humphreys2020AdvancingFW} urges that ``the deployment of a Fusion Machine Learning Data Platform could in itself prove a transformational advance, dramatically increasing the ability of fusion science, mathematics, and computer science communities to combine their areas of expertise in accelerating the solution of fusion energy problems'' \cite{Humphreys2020AdvancingFW}.
Many fields in physical and life sciences could make the same assertion, including all the domains we've discussed in the context of SI.
A recent example of such a platform is Microsoft's ``Planetary Computer'', an open-source cloud platform for multi-petabye global environmental data: \href{https://planetarycomputer.microsoft.com/}{planetarycomputer.microsoft.com}. Largely an Earth observation catalogue and development hub, the datasets exemplify the above list of principles, and the ecosystem is designed to cultivate interdisciplinary and open science. The open-source code should eventually include pipelines for data processing and engineering, including modules for the principled use of real and synthetic data. 

Hutchinson et al. \cite{Hutchinson2021TowardsAF} provide a thorough regimen for ML dataset lifecycles and supporting software infrastructure, while the design of SI data platforms should also consider how the new scientific workflows like Fig. \ref{fig:pipeline} can have different requirements and challenges from ML in general. Consider that each phase of such a workflow may have several implicit dataflows, each executing a set of chained data transformations consuming and producing datasets; 
A challenge is these phases are often heterogeneous in a few ways: First, software, hardware, and development practices are inconsistent between collaborating people, teams, and organizations. Second, various fields call for domain-specific stuff such as specialized data types and structures.
And mainly, most of the available software has grown through accretion, where code addition is driven by the need for producing domain science papers without adequate planning for code robustness and reliability. \cite{Grannan2020UnderstandingTL}
The Department of Energy's 2019 AI for Science Report states ``Today, the coupling of traditional modeling and simulation codes with AI capabilities is largely a one-off capability, replicated with each experiment. The frameworks, software, and data structures are distinct, and APIs do not exist that would enable even simple coupling of simulation and modeling codes with AI libraries and frameworks.'' 
The interdisciplinary nature of simulation intelligence, along with the many possible motif integrations, offer opportunity for SI data platforms to address these concerns.

The implementation and engineering of such data platforms is far from trivial \cite{Chervenak2000TheDG,Simmhan2009BuildingRD}.
First consider that the curation and preparation of data are often the most time-consuming and manual tasks, which end up being disconnected and distributed in scientific endeavors. For the former, much manual and highly specialized work is performed by domain scientists to gather the relevant scientific data, and/or to use ML to achieve automated knowledge extraction from scientific data. Even then there is a significant gap between raw (or even cleaned) scientific data and useful data for consumption by ML and other parts of the workflow including simulators. Not to mention most scientific domains have specialized data formats that may require industry-specific software and domain-specific knowledge to inspect, visualize, and understand the data. 
See, for example, Simmhan et al. \cite{Simmhan2009BuildingRD} discuss pipeline designs for reliable, scalable data ingestion in a distributed environment in order to support the Pan-STARRS repository, one of the largest digital surveys that accumulates 100TB of data annually to support 300 astronomers. Blamey et al. \cite{Blamey2020RapidDO} also provide an informative walk through  the challenges and implementations of cloud-native intelligent data pipelines for scientific data streams in life sciences.
Deelman et al. \cite{Deelman2018TheFO} advocate for co-located (or \textit{in situ}) computational workflows, to minimize inefficiencies with distributed computing. One motivation for co-located workflows is to minimize the cost of data movement by exploiting data locality by operating on data in place. Other supporting reasons include supporting human-in-the-loop interactive analysis, and capturing provenance for interactive data analysis and visualization (as well as any computational steering that results \footnote{\textit{Computational steering} methods provide interactivity with an HPC program that is running remotely. It is most commonly used to alter parameters in a running program, receive real-time information from an application, and visualize the progress of an application. The idea is to save compute cycles by directing an application toward more productive areas of work, or at a minimum stopping unproductive work as early as possible. In the numerical simulation community it more specifically refers to the practice of interactively guiding a computational experiment into some region of interest.}).
Data pipelines need to provide efficient, scalable, parallel, and resilient communication through flexible coupling of components with different data and resource needs, and utilize extreme-scale architectural characteristics such as deep memory/storage hierarchy and high core count. 

Technologies in large scale data processing and engineering have flourished in the previous decades of ``big data'', and even more so with the accelerating adoption of AI and ML. It can be expected that SI will be a similar accelerant of data technologies and ecosystems. 
Large scale computing (i.e., peta-to-exa scale simulation) imposes significant challenges in engineering data pipelines \cite{Fiore2018OnTR,Chang2019NISTBD}, in particular for the HPC community which we discuss in the next section.

\hfill \break
\subsection{Accelerated computing}\label{sec_infra}

High performance computing (HPC) plays a critical role in simulation systems and scientific computing. HPC systems are mainly used in domains with complex data-intensive applications, applying advanced mathematics and sophisticated parallel algorithms to solve large scale problems appearing in physics, medicine, mathematics, biology, and engineering~\cite{Chen2013SynergisticCI}. 
Established supercomputer codes (where a supercomputer is system optimized for HPC applications) build on many years of research in parallelization \cite{Mezmaz2007AGB,Melab2018MulticoreVM,Oger2016OnDM} and numerical methods such as linear and nonlinear solvers, algorithms for continuous and discrete variables, automatic mesh refinements \cite{Bryan2014ENZOAA,Hopkins2015ANC} and others -- for example, the PETSc (\href{http://petsc.org}{petsc.org}) suite of data structures and routines for the scalable (parallel) solution of scientific applications modeled by partial differential equations \cite{petsc-user-ref}.
Advances in ML and AI have impacted supercomputers in significant ways: enabling lower precision data types, leveraging tensor operations, and distributed machine learning frameworks. Further, it is becoming easier to solve traditional problems in a non-traditional way, leveraging neural networks. For instance, ML has proven highly successful for analyzing large-scale simulation data produced by supercomputers, such as extracting climate extremes \cite{Kurth2018ExascaleDL}, and towards efficient surrogate models or emulators based on HPC ensemble simulations \cite{Lu2019EfficientSM}. 
One prominent example of a problem requiring massive computing resources is molecular dynamics (MD). ``Anton''~\cite {7012191} is a family of supercomputers designed from scratch to solve precisely this one problem; recall from Fig.~\ref{fig:brainbloodflow} that MD is one subset of the broader spatiotemporal space needed to simulate dynamics of our world. 
Recently, researchers at Argonne National Laboratory developed an AI-driven simulation framework for solving the same MD problem, yielding ~50x speedup in time to solutions over the traditional HPC method \cite{10.1145/3468267.3470578}.
And some work suggests these approaches need not be mutually exclusive: it has been shown in the context of computational fluid dynamics that traditional HPC workloads can be run alongside AI training to provide accelerated data-feed paths \cite{stencil}.
Co-locating workloads in this way may be necessary for petascale (approaching exascale) scientific simulations: Compute aside, supercomputers or large clusters (distributed compute nodes) are the only way to host some of the largest currently available models -- as their memory requirements go into trillions of parameters, partitioning the model is the only way. In particular, the ensemble simulations can only be done in HPC. 
That said, a unified, two-way coupling between such large scale simulators and ML (which calls for data assimilation at each time step, or updating parameters and state variables) remains a significant challenge due to the complexity of the system's mathematical equations (which are often nonlinear) as well as the existing input-output (I/O) structures that are often built in a black-box way. 

Many HPC experts suggest that the highest performing computing systems of the future will be specialized for the  application at hand, customized in the hardware itself \textit{and} in the algorithms and abstractions underlying the software.\footnote{A different perspective can be drawn from the example of ``Folding@home''~\cite{Larson2009FoldingHomeAG}, a distributed computing project for running protein dynamics simulations on networks of volunteers' personal computers that consist of heterogeneous hardware components. The project aims to help scientists develop new therapeutics for a variety of diseases by simulating the processes of protein folding and the movements of proteins at scale.} Both of these considerations -- distributed compute and HW-SW specialization, not to mention heterogeneous systems incorporating a variety of specialized components -- can be powerful but make application development more difficult \cite{Stoller2019FutureDF}. 
At the hardware level, Fiore et al. \cite{Fiore2018OnTR} suggest new processors, memory hierarchy, and interconnects have to be taken into consideration when designing HPC systems to succeed existing petascale systems. Some works have stated that silicon photonic interconnects could provide large bandwidth densities at high-energy efficiencies to help solve some issues related to the increased parallelism and data intensity and movement \cite{Rumley2017OpticalIF, Cheng2018RecentAI}. At the software level \cite{Grannan2020UnderstandingTL}, compilers, libraries, and other middleware, should be adapted to these new platforms -- perhaps in automated ways enabled by machine programming and program synthesis (e.g. in ``Arbor'' for neuronal network simulations on HPC \cite{Akar2019ArborA}). At the applications level, algorithms, programming models, data access, analysis, visualization tools and overall scientific workflows \cite{Deelman2018TheFO}, have to be redesigned when developing applications in petascale (potentially to exascale) computing infrastructures -- for instance, HPC systems (and teams) would do well to adopt the cycles and tooling of the DevOps and newer MLOps~\cite{Renggli2021ADQ} ecosystems, although unique characteristics related to HPC-level data storage and resource sharing will present new challenges. The interdisciplinary nature of these computing topics is paramount: collaboration and synergy among the domain science, mathematics, computer science and engineering, modeling and simulation, etc., an optimal system and solution can be built. Like the integration of SI motifs and the broader SI stack, more comprehensive and holistic approaches are required to tackle the many complex real-world challenges.

An interesting and active use-case for HPC is quantum computing, where researchers use supercomputers to simulate quantum circuits \cite{haner2016high,smelyanskiy2016qhipster,villalonga2020establishing,liu2021closing,liu2021redefining} and quantum computers more generally. The invention of quantum computing was motivated \cite{feynman1982simulating} by the fact that quantum computers are exponentially (in the number of quantum bits, circuit depth, or both) expensive to simulate---this both gives QCs their power and also makes it very challenging to simulate even intermediate-scale QC \cite{preskill2018quantum}. Classical simulations of QCs are important in at least two areas: quantum-algorithm design (enabling empirical testing and verification of new quantum algorithms) and quantum-system design (enabling the simulation of both quantum hardware and layers of the quantum stack \cite{jones2012layered} above hardware but below algorithms). Accelerated and accurate simulation of quantum computers will play an important role in the development of practical, useful quantum computers. In parallel to the growth of conventional HPC acceleration of QC over the past several years, there has been an explosion of interest in the application of machine-learning tools to problems in quantum physics in general \cite{carleo2017solving,torlai2018neural,carrasquilla2020machine} and to the simulation of QC in particular \cite{carrasquilla2021probabilistic}. We are already seeing SI approaches applied to the domain of QC. Another potential application of HPC to QC, which is essentially entirely separate from the simulation use case, is that of performing classical co-processing to aid in the execution of error-correction protocols for quantum hardware. Already there have been efforts to apply neural networks to help with error decoding \cite{carrasquilla2020machine}, and it seems likely that quantum error correction will ultimately be implemented at scale using a combination of HPC and AI methods. Finally, the practical uses of QC typically involve both quantum and classical processing \cite{nielsen2011quantum}, with many use cases requiring very substantial amounts of classical computation that themselves warrant the use of HPC resources. As a concrete example, quantum-chemistry simulations are anticipated to involve a combination of both classical and QC computations \cite{malone2021towards} (which we've alluded to earlier in several motif examples).

As the field makes advances in the specific motifs and SI overall, there are several open questions to keep an eye on regarding computation:
\begin{itemize}
    \item Will advances in the motifs lead to new breeds of algorithms that require radically new systems and/or architectures? Perhaps these can be implemented as conventional ASICs (Application-Specific Integrated Circuits), or will new substrates beyond CMOS electronics\cite{shalf2015computing} be preferred?
    \item Will SI inverse design produce novel compute architectures and substrates? 
    \item How will the open-endedness paradigm affect computing systems (memory allocation, load balancing, caching, etc.)?
    \item How will Machine Programming influence hardware-software co-design for simulation intelligence (and high-performance computing)?
    \item In addition to the traditional goals of performance and scalability (latency, throughput, etc.), energy efficiency, and reliability, how will the advancing technologies build abstractions to facilitate reasoning about correctness, security, and privacy \cite{Stoller2019FutureDF}? 
    \item Will SI-based surrogate modeling enable a landscape shift where computationally lightweight surrogates running on a small number of GPUs replace massive supercomputing workloads (particularly in physical and life sciences simulations)? What will be the metrics and benchmarks to quantify the tradeoffs between such orthogonal computation approaches? 
\end{itemize}

To the first question, there has already been a resurgence of interest in novel hardware platforms for both simulation \cite{athale2016optical,estakhri2019inverse} and machine learning \cite{tanaka2019recent,hughes2019wave,xiao2020analog,wetzstein2020inference,markovic2020physics,wright2021deep,furuhata2021physical}. Although CMOS platforms are still used essentially exclusively due to their performance and reliability, we anticipate that special-purpose hardware built on non-CMOS physical substrates may gain traction over the next decade.
It remains to be seen if and how appropriate benchmarks are developed for comparing different compute substrates, which may ultimately be infeasible in a general, application-agnostic way. This echoes a significant benchmarking issue we discussed earlier in the context of physics-infused ML: It is customary that distributed training algorithms in HPC platforms are benchmarked using idealized neural network models and datasets \cite{Huerta2020ConvergenceOA}, which does not impart any insights regarding the actual performance of these approaches in real deployment scenarios -- i.e., when using domain-inspired AI architectures and optimization schemes for data-driven discovery in the context of realistic datasets (which are noisy, incomplete, and heterogeneous), and when using purpose-built HPC stacks that can have application-specific customizations up-and-down the layers we described above.

\subsection{Domains for use-inspired research}

Much of the work to develop the motif methods and their many integrations is \textit{foundational SI research}: Understanding the mechanisms underlying thought and intelligent behavior and their implementation in machines.
The main descriptions of each motif fall in this category, from the mathematics of causality to multi-agent cooperation.
Pursuing foundational research in-sync with use-inspired research is essential, in SI and broadly in simulation, AI, and data technologies. \textit{Use-inspired research} is basic research with uses for society in mind, thus a main driver of breakthrough innovations. Sometimes referred to as ``Pasteur's quadrant''\footnote{\textit{Pasteur's quadrant} is a classification of scientific research projects that seek fundamental understanding of scientific problems, while also having immediate use for society. Louis Pasteur's research is thought to exemplify this type of method, which bridges the gap between ``basic'' and ``applied'' research.}, the many example motif applications we've discussed align with this division of research and development. The National Science Foundation similarly describes their AI Institutes as operating in cycles of foundational and use-inspired AI research: \href{https://www.nsf.gov/cise/ai.jsp}{nsf.gov/cise/ai.jsp}.

Table \ref{tab:motifs} provides a high-level, non-exhaustive look of the domains for use-inspired research per motif. Note  many  methods  and  use-cases  arise  from  the  integrations  of  motifs, as  discussed  throughout  the  text, which are not necessarily elucidated in this high-level table view.

\begin{table}[!t]
\centering
\vskip 10em
\begin{tabular}{l|ccccccccccccccccccccccc}
& \begin{rotate}{60} Machine programming \end{rotate}
& \begin{rotate}{60} Differentiable programming \end{rotate}
& \begin{rotate}{60} Probabilistic programming \end{rotate}
& \begin{rotate}{60} Multi-physics multi-scale \end{rotate}
& \begin{rotate}{60} Simulation-based inference \end{rotate}
& \begin{rotate}{60} Surrogate modeling, emulation \end{rotate}
& \begin{rotate}{60} Causal reasoning \end{rotate}
& \begin{rotate}{60} Agent-based modeling \end{rotate}
& \begin{rotate}{60} Open-endedness \end{rotate}
\\
\hline
\rowcolor{purp}
{HEPhysics - theoretical}& - & \checkmark & \checkmark & \checkmark & \checkmark & - & \checkmark & - & \checkmark \\
\rowcolor{purp}
{HEPhysics - experimental}& \checkmark & \checkmark & \checkmark & \checkmark & \checkmark & \checkmark & \checkmark & - & - \\

{Complexity}& - & - & - & \checkmark & - & - & \checkmark & \checkmark & \checkmark \\

\rowcolor{purp}
{Synthetic biology}& \checkmark & \checkmark & \checkmark & \checkmark & \checkmark & \checkmark & \checkmark & - & - \\

{Chemistry}& - & \checkmark & \checkmark & \checkmark & \checkmark & \checkmark & - & - & - \\

\rowcolor{purp}
{Materials}& - & \checkmark & \checkmark & \checkmark & \checkmark & \checkmark & - & - & - \\

{Medicine}& - & \checkmark & \checkmark & \checkmark & \checkmark & \checkmark & \checkmark & - & - \\

\rowcolor{purp}
{Systems biology}& - & \checkmark & \checkmark & \checkmark & - & \checkmark & \checkmark & \checkmark & \checkmark \\

{Neuro and Cog sciences}& \checkmark & \checkmark & \checkmark & \checkmark & \checkmark & \checkmark & \checkmark & \checkmark & \checkmark \\

\rowcolor{purp}
{Energy - nuclear fission \& fusion}& - & \checkmark & \checkmark & \checkmark & \checkmark & \checkmark & - & - & \checkmark \\
\rowcolor{purp}
{Energy - materials \& storage}& - & \checkmark & \checkmark & \checkmark & \checkmark & \checkmark & - & - & \checkmark \\

{Manufacturing}& \checkmark & \checkmark & - & \checkmark & - & \checkmark & - & - & \checkmark \\

\rowcolor{purp}
{Energy systems}& - & - & - & \checkmark & - & \checkmark & \checkmark & \checkmark & \checkmark \\

{Transportation \& infrastructure}& - & - & - & \checkmark & - & \checkmark & \checkmark & \checkmark & \checkmark \\

\rowcolor{purp}
{Agriculture}& - & \checkmark & - & \checkmark & - & \checkmark & \checkmark & \checkmark & \checkmark \\

{Ecology}& - & \checkmark & \checkmark & \checkmark & \checkmark & \checkmark & \checkmark & \checkmark & \checkmark \\

\rowcolor{purp}
{Socioeconomics \& markets}& - & \checkmark & - & \checkmark & - & - & \checkmark & \checkmark & \checkmark \\

{Finance}& \checkmark & \checkmark & \checkmark & - & - & - & \checkmark & \checkmark & \checkmark \\

\rowcolor{purp}
{Geopolitics}& - & - & - & \checkmark & - & - & \checkmark & \checkmark & \checkmark \\

{Defense}& \checkmark & \checkmark & \checkmark & \checkmark & \checkmark & \checkmark & \checkmark & \checkmark & \checkmark \\

\rowcolor{purp}
{Climate}& - & \checkmark & - & \checkmark & \checkmark & \checkmark & \checkmark & \checkmark & \checkmark \\

{Earth systems}& - & \checkmark & - & \checkmark & \checkmark & \checkmark & \checkmark & \checkmark & \checkmark \\

\rowcolor{purp}
{Astrophysics \& cosmology}& - & \checkmark & \checkmark & \checkmark & \checkmark & \checkmark & \checkmark & - & \checkmark \\


\hline
\end{tabular}
\vskip 1em
\caption{
The motifs can be applied for science and intelligence workflows in diverse domains and interdisciplinary use-cases. Here we itemize some main application areas (approximately ordered from smallest physical scales to largest). Note many methods and use-cases arise from the integrations of motifs (as discussed throughout the text), which are not necessarily apparent in this table view.
Some useful information on the listed domains:
\textit{HEP} stands for high-energy physics (also called particle physics);
\textit{Complexity} includes systems and intelligence as defined by the Santa Fe Institute at \href{https://www.santafe.edu/what-is-complex-systems-science}{\textit{What is complex systems science?}}
and \href{https://www.santafe.edu/research/themes/complex-intelligence-natural-artificial-and-collec}{\textit{Complex intelligence: natural, artificial, collective}}, respectively; 
\textit{Manufacturing} notably includes ML-based design of sensors and chips; 
\textit{Earth systems} includes oceans, land, air, and near space (see \href{https://earthdna.org}{earthdna.org}).
}
\label{tab:motifs}
\end{table}

\paragraph{Brain simulation}
We've discussed several neuroscience (and cognitive science) use-cases of SI motifs, notably with simulation-based inference and probabilistic programming (as well as ``working memory'' and inverse design). The intersections of neuroscience and simulation, including big data and machine learning, have had profound effects on the various fields for decades. The review by Fan \& Markrum \cite{Fan2019ABH} provides a thorough overview that is out of scope for the current paper. The authors detail the last century of brain research, building up to today's supercomputer whole-brain models \cite{Markram2006TheBB}, with a focus on the recent phases of ``simulation neuroscience'' and the ``multi-scale brain''. The former describes a collaborative, integrated research setting where \textit{in silico} and \textit{in situ} experiments interplay to drive mathematical and predictive models (as well as produce hypotheses), all leveraging a wealth of biological datasets on a cohesive, open platform. The latter describes the need for causally linking molecular, cellular and synaptic interactions to brain function and behavior. We aim for SI to play critical roles in these phases of brain research for decades to come.

Simulations provide the crucial link between data generated by experimentalists and computational (as well as theoretical) neuroscience models; comparing predictions derived from such simulations with experimental data will allow systematic testing and refinement of models, and also inform theory in ways that will directly influence neuro-inspired AI (e.g., as in neural circuitry models for computer vision \cite{George2017AGV}, working memory \cite{Hawkins2016WhyNH,Soto2008AutomaticGO}, predictive coding \cite{Bastos2012CanonicalMF,Friston2010TheFP}, and others such as spiking NNs \cite{Taherkhani2020ARO}, language of thought \cite{Ullman2012TheoryLA}, and neuromorphic computing \cite{Markovi2020PhysicsFN}) -- see Refs.~\cite{Ullman2019UsingNT, George2020FromCT, Hassabis2017NeuroscienceInspiredAI} for more discussion.


\hfill \break

\section{Discussion}\label{sec_disc}

In addition to the nine SI motifs, a number of advances in ML also provide encouraging directions for simulation-based sciences and ML, which we describe below as ``honorable mention'' methods in scientific AI and ML.
We then close with discussion on the science of science: methods, breakthroughs, and the role for SI going forward.

\subsection{Honorable mention motifs}

\paragraph{\textit{Geometric and equivariant neural networks}}
In various physics we encounter symmetries, or \textit{equivariance}, defined as the property of being independent of the choice of reference frame. 
A function is equivariant to a transformation if applying that transformation to its inputs results in an equivalent transformation of its outputs. A special case of equivariance that has been exploited largely in machine learning is \textit{invariance}, where any transformation of the inputs results in an identity transformation of the outputs: the architectures of CNNs and RNNs make them invariant to image translations and sequence position shifts, respectively. Thus neural networks with equivarience properties are an intriguing direction of research. Further, these methods share maths with category theory, which may help define equivariant architectures beyond CNN and RNN. To this end, Cohen \& Welling \cite{Cohen2016GroupEC} generalize CNNs to G-CNNs, where generalized-convolution replace the position shifting in a discrete convolution operation with a general group of transformations (like rotations and reflections, in addition to translations). More recent developments by Cohen et al. \cite{Cohen2019GaugeEC} extend neural network equivariance beyond group symmetries, developing neural networks that consume data on a general manifold dependent only on the intrinsic geometry of the manifold. 

Graph-structured data, one of the more prevalent types of data used in ML, presents many opportunities and challenges for exploiting symmetries. \textbf{Graph neural networks (GNNs)} are an extremely active area in the ML community for learning and predicting on flexible, graph-based inputs, with many applications in chemistry, material sciences, medicine, and others \cite{Fung2021BenchmarkingGN,Wu2019ACS}.
Because of their scalability to large graphs, graph convolutional neural networks or message passing networks are widely used \cite{Haan2020NaturalGN}. Yet these networks, which pass messages along the edges of the graph and aggregate them in a permutation invariant way, are fundamentally limited in their expressivity \cite{Xu2019HowPA}. Equivariant graph networks are more expressive \cite{Maron2019InvariantAE}, thanks to the generalization properties (rooted in category theory). Recent work from Haan et al. \cite{Haan2020NaturalGN} leverage category theory to derive a more expressive class of ``natural graph networks'' that can be used describe maximally flexible global and local graph networks -- some of this relationship is discussed in \cite{Shiebler2021CategoryTI}.

Specializing GNN architectures for the task at hand has delivered state-of-art results in areas like drug discovery \cite{Stokes2020ADL}. One can view such architecture specialities as inductive biases, specifically on the structure of the input (i.e., the input's dimensions are related such that they form a graph structure). 
This has taken shape as a class called \textbf{\textit{Geometric Deep Learning}}, where encoding the symmetries present in the problem domain are required for building neural network models of geometric objects.
Said another way, this direction of work aims to generalize equivariant and invariant neural network operators to non-Euclidean domains (e.g. graphs and manifolds) \cite{Bronstein2017GeometricDL}. Bronstein et al. \cite{Bronstein2017GeometricDL} illustrate that convolution commutes with the Laplacian operator, and they use this insight to define both spatially and spectrally motivated generalizations of convolution for non-Euclidean domains.
In particular, equivariant neural networks have architectures designed for vector-transforms under symmetry operations (on reference frames) \cite{Cohen2019GaugeEC}. By constraining NNs to be symmetric under these operations, we've seen successful applications in areas like molecular design \cite{Simm2020ReinforcementLF} and quantum chemistry \cite{Qiao2021UNiTEUN}.
There is also recent exploration in tying these structural constraints to classes of structural casual models (SCMs) that we discussed earlier, which could leade to theoretical foundations of how GNNs can be used to perform causal computations \cite{Zecevic2021RelatingGN}.
This is a robust area of ML and applied maths research -- the interested reader can explore \cite{Gerken2021GeometricDL} for thorough overview, and \cite{Esteves2020TheoreticalAO} for theoretical foundations of group equivariance and NNs.

\paragraph{\textit{Learning mesh-based simulators}}
Software for modeling of physical bodies and systems typically use mesh-based representations for the many geometries of surfaces and volumes. Meshes are used to solve underlying differential equations in the context of finite element analysis (FEA), computational fluid dynamics (CFD), and related statics and dynamics solvers. These present significant computational bottlenecks, which are weakly mitigated by adaptive meshes that are finer in areas that need high precision and coarser otherwise.
New approaches from the machine learning community aim to alleviate such bottlenecks by learning the dynamics of deforming surfaces and volumes rather than the expensive discretization schemes \cite{blazek2015principles} -- that is, machine learned emulators for CFD and FEA. Pfaff et al. \cite{Pfaff2021LearningMS} develop a GNN model that can be trained to pass messages on a mesh graph and to adapt the mesh discretization during forward simulation. They show early results on common physical dynamics simulations of deforming plate, cloth in wind field, and flow are cylinders and airfoils. Similarly using learned message passing to compute dynamics, Sanchez-Gonzalez et al. \cite{SanchezGonzalez2020LearningTS} implement another GNN-based approach for efficiently simulating particle-based systems (and they suggest the same approach may apply to data represented as meshes as well).
Both works point to the value of differentiable simulators, providing gradients that can be used for design optimization, optimal control tasks, and inverse problems.

\paragraph{\textit{Differentiable rendering}}
Algorithms from the computer graphics community can have intriguing implications for ML, for instance for the generation of synthetic training data for computer vision \cite{wood2021fake,behl2020autosimulate,jaipuria2020deflating} and the class of methods referred to as ``vision as inverse graphics'' \cite{Romaszko2017VisionasInverseGraphicsOA}. 
Rendering in computer graphics is the process of generating images of 3D scenes defined by geometry, textures, materials, lighting, and the properties and positions of one or more cameras. 
Rendering is a complex process and its differentiation is not uniquely defined, yet differentiable rendering (DR) techniques based on autodiff can tackle such an integration for end-to-end optimization by obtaining useful gradients \cite{Kato2020DifferentiableRA}.
A driving interest from the ML perspective is towards computer vision models (largely deep neural networks and PGMs) that understand 3D information from 2D image data. One expectations is the improvement in vision applications while reducing the need for expensive 3D data collection and annotation.
One general approach is an optimization pipeline with a differentiable renderer, where the gradients of an objective function with respect to the scene parameters and known ground-truth are calculated. 
Another common approach is self-supervision based on differentiable rendering, where the supervision signal is provided in the form of image evidence and the neural network is updated by backpropagating the error between the image and the rendering output.
See Kato et al. \cite{Kato2020DifferentiableRA} for an overview of DR algorithms, data representations, evaluation metrics, and best practices.

\paragraph{\textit{Normalizing and Manifold Flows}} 

Although very expressive generative models, a main limitation of the popular VAE and GAN approaches is intractable likelihoods (probability densities) which make inference challenging. 
They essentially represent a lower-dimensional data manifold that is embedded in the data space, and learn a mapping between the spaces.
A class of generative models called \textit{normalizing flows} \cite{rezende2015variational,Papamakarios2019NormalizingFF,kobyzev2020normalizing} are based on a latent space with the same dimensionality as the data space and a diffeomorphism; their tractable density permeates the full data space and is not restricted to a lower-dimensional surface \cite{Brehmer2020FlowsFS}. Normalizing flows describe a general mechanism for defining expressive probability distributions by transforming a simple initial density into a more complex one by applying a sequence of invertible (bijective) transformations \cite{rezende2015variational,Papamakarios2019NormalizingFF}, can provide a richer, potentially more multi-modal distributions for probabilistic modeling.
In theory the expressiveness is unlimited: Papamakarios et al. \cite{Papamakarios2019NormalizingFF} show that for any pair of well-behaved distributions $p_{x}(\mathrm{x})$ and $p_{u}(mathrm{u})$ (the target and base, respectively), there exists a diffeomorphism that can turn $p_{u}(\mathrm{u})$ into $p_{x}(\mathrm{x})$; $x$ and $u$ must have the same dimensionality and every sub-transformation along the way must preserve dimensionality. With distributions of larger dimensions there is larger computational cost per sub-transform; various classes of composable transformations exist, such as residual and radial flows, with known computational complexities. Multi-scale architectures can be used mitigate the growing complexity by clamping most dimensions in the chain of transforms, thus only applying all steps to a small subset of dimensions, which is less costly than applying all steps to all dimensions \cite{Dinh2017DensityEU}.
The same approach can be utilized for multi-scale modeling, which is a natural modeling choice in granular data settings such as pixels and waveforms \cite{Papamakarios2019NormalizingFF}, and in many scientific environments as we've discussed.
A class of flows particularly relevant to SI may be \textit{continuous-time flows}, where the transformation from noise $u$ to data $x$ is described by an ordinary differential equation (rather than a series of discrete steps) \cite{Salman2018DeepDN,Grathwohl2019FFJORDFC,Durkan2019NeuralSF}. There are also \textit{equivariant manifold flows} \cite{Katsman2021EquivariantMF} which aim to learn symmetry-invariant distributions on arbitrary manifolds, to overcome issues with the main generalized manifold-modeling approaches (both Euclidean and Riemannian). Much like the other equivariant methods we've discussed, we need to properly model symmetries for many applications in the natural sciences -- e.g., symmetry requirements arise when sampling coupled particle systems in physical chemistry \cite{Khler2020EquivariantFE}, or sampling for $SU(N)$ lattice gauge theories in theoretical physics in order to enable the construction of model architectures that respect the symmetries of the Standard Model of particle and nuclear physics and other physical theories \cite{Boyda2021SamplingUS}.
For a thorough review of normalizing flows for probabilistic modeling and inference, refer to Papamakarios et al. \cite{Papamakarios2019NormalizingFF}.

\begin{figure}[!ht]
\centering
\includegraphics[width=1.0\linewidth]{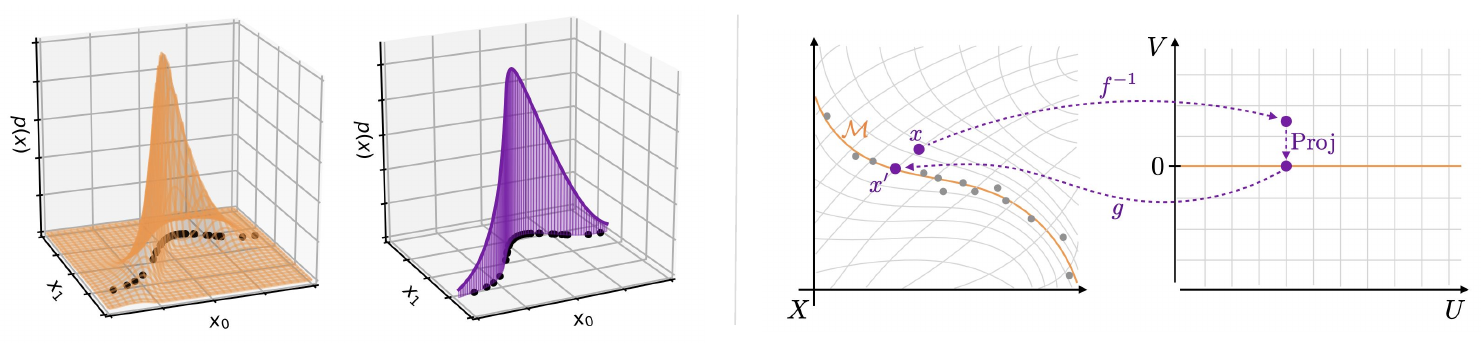}
\caption{
(Left) Sketch of how a standard normalizing flow in the ambient data space (left, orange surface) and an M-flow (right, purple) both model the same data (black dots).
(Right) Sketch of how an M-flow evaluates arbitrary points on or off the learned manifold: First we show the data space $X$ with data samples (grey) and the embedded manifold $M$ (orange). Then the latent space $U × V$ is shown. In purple we sketch the evaluation of a data point $x$ including its transformation to the latent space, the projection onto the manifold coordinates, and the transformation back to the manifold.
(Figures reproduced from \cite{Brehmer2020FlowsFS})
}
\label{fig:mflows}
\end{figure}

Brehmer \& Cranmer \cite{Brehmer2020FlowsFS} highlight a limitation: 
Normalizing flows may be unsuited to data that do not populate the full ambient data space they natively reside in, and can learn a smeared-out version rather than the relatively higher dimensional manifold exactly (illustrated in Fig. \ref{fig:mflows}).
They introduce \textit{manifold-learning flows (M-flows)}: normalizing flows based on an injective, invertible map from a lower dimensional latent space to the data space. The key novelty is that M-flows simultaneously learn the shape of the data manifold, provide a tractable bijective chart, and learn a probability density over the manifold, thus more faithfully representing input data that may be off manifold (in ambient data space). 
M-flows describe a new type of generative model that combines aspects of normalizing flows, autoencoders, and GANs; flows that simultaneously learn the data manifold and a tractable density over it may help us to unify generative and inference tasks in a way that is tailored to the structure of the data.
In many scientific cases, domain knowledge allows for exact statements about the dimensionality of the data manifold, and M-flows can be a particularly powerful tool in a simulation-based inference setting \cite{Cranmer2020TheFO}.

\paragraph{\textit{AlphaFold}}

The recent landmark results from DeepMind's AlphaFold \cite{Jumper2021HighlyAP} on the Critical Assessment of Techniques for Protein Structure Prediction (CASP) competition showed dominance over the field for state-of-art in three-dimensional protein structure modeling.\footnote{See \href{https://www.nature.com/articles/d41586-020-03348-4}{``It will change everything: DeepMind's AI makes gigantic leap in solving protein structures''} in Nature News, and \href{https://deepmind.com/blog/article/alphafold-a-solution-to-a-50-year-old-grand-challenge-in-biology}{``AlphaFold: a solution to a 50-year-old grand challenge in biology''} from DeepMind.}
AlphaFold's approach to computational protein structure prediction is the first to achieve near-experimental accuracy in a majority of cases. Overall there were several deep learning innovations to assemble the network purpose-built for protein structures \cite{Jumper2021HighlyAP}: a new architecture to jointly embed multiple sequence alignments and pairwise features, a new output representation and associated loss that enable accurate end-to-end structure prediction, a new equivariant attention architecture, and others including curated loss functions. While the success of AlphaFold is undeniable, questions remain as to how much of the AlphaFold modeling framework, and knowledge about it, can be adapted to solve other, similar fundamental challenges in molecular biology, such as RNA folding and chromatin packing -- how much effort would be needed to customize the tool to other areas to make a fresh contribution to them, and how to evaluate the contribution of the AI tool itself versus the human in the loop at most stages?
A more philosophical outcome of AlphaFold is the implication that a scientific challenge can be considered partially or totally solved even if human scientists are unable to understand or gain any new knowledge from the solution, thus leaving open the question of whether the problem has really been solved.
It will be intriguing to explore this perspective and the related area of AI-anthropomorphism \cite{Salles2020AnthropomorphismIA} as more scientific, mathematical, and engineering challenges are overcome by various types of machine intelligence.

\paragraph{\textit{Quantum computing}}

Quantum computing \cite{nielsen2011quantum} has strong connections with both simulation and machine learning. With the recent experimental demonstration of quantum computational supremacy \cite{arute2019quantum} and rapid progress in the development of noisy intermediate-scale quantum (NISQ) machines \cite{preskill2018quantum} and algorithms\cite{bharti2021noisy}, there is optimism that quantum computers might even deliver an advantage in tasks involving the simulation of quantum systems in the foreseeable future. In the longer term, once large-scale, fault-tolerant quantum computation is achieved, quantum simulation is widely expected to be a transformative use case \cite{tacchino2020quantum,mcardle2020quantum,cerezo2021variational}.

There also exists a large body of work investigating diverse ways in which quantum computers might be used to perform machine-learning tasks -- the subfield of \textit{quantum machine learning (QML)} \cite{biamonte2017quantum,dunjko2020non}. QML methods exist both for tasks involving classical data as well as for tasks involving quantum data, although there is a general expectation that it will be easier to obtain an advantage with tasks on quantum data. The notion of differentiable programming has also been extended to quantum computing, first as part of QML, but now also applied to other areas of quantum algorithms \cite{mitarai2018quantum,bergholm2018pennylane,broughton2020tensorflow,cerezo2021variational}. Since quantum algorithms -- including quantum-simulation algorithms, but also algorithms simulating classical systems on quantum computers -- can generate quantum data, there is a natural point of connection between simulation on quantum computers and QML for quantum data, and one might expect even closer interplay between simulation and machine learning methods in quantum computing, as we have seen develop in classical computing, and which this paper describes the merger of.

\hfill \break
\subsection{Conclusion}\label{sec_conc}

Donald Braben, a pioneer in innovation practices and the culture of scientific research, defines \textit{transformative research} as ``research that sets out radically to change the way we think about an important subject''~\cite{Braben2008ScientificFT}. 
Advancing the Nine Motifs and realizing synergistic artificial intelligence and simulation sciences can provide the requisite tools for transformative research across domains. 

Yet simulation in conjunction with artificial intelligence does not necessarily yield solutions. Rather, when integrated in the ways we've described, it enables intelligent navigation of complex, high-dimensional spaces of possible solutions: bounding the space helps to guide the user to optimal solutions, and, in many cases, expanding the space to new territories can enable solutions beyond existing knowledge or human priors (Fig.~\ref{fig:knowledge}).

Unlike current statistical machine learning approaches, simulation-based approaches can better deal with cause-effect relationships to help find first principles and the generating mechanisms required to make progress in areas of scientific discovery. Whether this is for problems in inverse design of synthetic biology compounds, predicting the results of interventions in social systems, or optimizing experimental design in high-energy physics, the Nine Motifs of SI explored here enable one to intelligently and tractably navigate the search space of possibilities.

This is why we put forth the concept of new SI-based scientific methods. While the traditional scientific method can be described as \textit{hypothesis-driven deduction}, 
21st century advances in big data brought about a major shift affecting nearly all fields, opening the door for a second dimension of the scientific method: \textit{data-mining-inspired induction} \cite{Voit2019PerspectiveDO}. More recently, machine learning along with high performance computing shaped this inductive reasoning class of scientific method to lean on the patterns and predictions from data-driven models.


It is clear the traditional scientific method provided a useful step-by-step guide for knowledge gathering, but this guide has also fundamentally shaped how humans think about the exploration of nature and scientific discovery. That is, presented with a new research question, scientists have been trained to think immediately in terms of hypotheses and alternatives, and how to practically implement controlled tests \cite{Voit2019PerspectiveDO}. While effective for centuries, the process is also very slow, and subjective in the sense that it is driven by the scientist's ingenuity as well as bias -- this bias has sometimes been a hindrance to necessary paradigm shifts \cite{Kuhn1962TheSO}.
It is possible that additional dimensions (or classes) of scientific methods can shift this classical regime to additional ``modes'' of knowledge gathering, which is what we aim to motivate in this work with simulation intelligence.

One can expect significant advances in science and intelligence with the development and adoption of these modes, but it is also possible that additional ethical and societal implications will arise.
Automation bias and algorithm aversion are concerns we raised in the context of human-machine teaming, and can appear in varied ways when exploring new dimensions of scientific methods with simulation and artificial intelligence technologies.
High costs associated with data collection, data access, and computing power are already problematic in today's society, particularly in the creation and exacerbation of inequalities in research output between state and private institutions as well as between developed and developing nations. Not to mention the ethics and safety challenges we encounter with AI in high-stakes settings, such as biomedical and social research, are similarly relevant with SI-based methods in these settings; performance variability of healthcare and sociopolitical applications across subsets of the population are already a challenge of conventional data-driven methods, and increasing the reliance on such methods at the research stage might exacerbate these issues further.

The continued blending of data-driven, AI, simulation, and large-scale computing workflows is a promising path towards major progress and the emergence of a new scientific methodology \cite{Succi2019BigDT}.
It is not unreasonable to expect that the SI motifs and the integrations we've described can broaden the scientific method in general, with reliable and efficient hypothesis--simulation--analysis cycles \cite{Stevens2020AIFS}, including machine-driven hypothesis generation for highly autonomous science.
Each of these motifs, particularly in combinations described by the SI stack, allows this process to happen faster and more effectively -- from probabilistic programming decoupling modeling and inference, to surrogate modeling allowing for rapid and powerful simulation of systems when detailed mechanistic modeling is infeasible, to simulation-based causal discovery deriving the true mechanisms of disease processes, to physics-infused learning deriving the governing equations of unknown physics, and so on. 
The SI driven scientific methods can help practitioners produce optimal and novel solutions across domains and use-cases in science and intelligence, and provide an essential step towards the \textit{Nobel Turing Challenge}, creating the machine intelligence engine for scientific discovery \cite{Kitano2021NobelTC}.

\newpage

\bibliographystyle{unsrtnat}
\bibliography{main}

\section*{Acknowledgements}
The authors would like to thank Tom Kalil and Adam Marblestone for their support, Joshua Elliot and Josh Tenenbaum for fruitful discussions, Sam Arbesman for useful reviews. 




\section*{Competing interests}
The authors declare no competing interests.

\section*{Additional information}
\textbf{Correspondence} and requests for materials should be addressed to A.L.


\end{document}